\documentclass[10pt,twocolumn,letterpaper]{article}
\pdfoutput=1

\usepackage[pagenumbers]{cvpr} 

\usepackage[dvipsnames]{xcolor}

\usepackage{booktabs}
\usepackage{bbm}
\usepackage{multirow}
\usepackage{marvosym}
\usepackage{tabularx,ragged2e}
\usepackage{tikz}
\usepackage{caption}
\usepackage{subcaption}
\usepackage{pgfplots}
\usepackage[export]{adjustbox}
\usepackage{environ}
\usepackage{makecell}

\usepackage{pifont}
\newcommand{\cmark}{\ding{51}}%
\newcommand{\xmark}{\ding{55}}%

\usepackage{enumitem}
\setenumerate{noitemsep}
\setlist{nosep} 

\usetikzlibrary{spy,backgrounds,calc}
\tikzset{tight/.style={inner sep=0, outer sep=0}}

\definecolor{cvprblue}{rgb}{0.21,0.49,0.74}
\usepackage[pagebackref,breaklinks,colorlinks,citecolor=cvprblue]{hyperref}

\makeatletter
\newsavebox{\measure@tikzpicture}
\NewEnviron{scaletikzpicturetowidth}[1]{%
  \def\tikz@width{#1}%
  \begin{lrbox}{\measure@tikzpicture}%
  \BODY
  \end{lrbox}%
  \pgfmathparse{#1/\wd\measure@tikzpicture}%
  \BODY
}
\makeatother

\DeclareRobustCommand{\rchi}{{\mathpalette\irchi\relax}}
\newcommand{\irchi}[2]{\raisebox{\depth}{$#1\chi$}} 

\newcommand{\myImgW}{3.3}
\newcommand{\myImgH}{2.4}

\newcommand{\myImgWd}{2.5}
\newcommand{\myImgHd}{2.5}

\newcommand{\myImgWs}{1.65}
\newcommand{\myImgHs}{1.3}

\newcommand{\myImgWsd}{1.35}
\newcommand{\myImgHsd}{1.7}

\def\paperID{202} 
\def\confName{3DV\xspace}
\def\confYear{2024\xspace}

\title{DeepDR: Deep Structure-Aware RGB-D Inpainting for Diminished Reality}

\newcommand*{\affaddr}[1]{#1}
\newcommand*{\affmark}[1][*]{\textsuperscript{#1}}

\author{%
    Christina Gsaxner\affmark[1] ({\tt {gsaxner@tugraz.at}}), Shohei Mori\affmark[1], Dieter Schmalstieg\affmark[1,2], Jan Egger\affmark[1,3], \\ Gerhard Paar\affmark[4], Werner Bailer\affmark[4] and Denis Kalkofen\affmark[1,5] 
    \vspace{0.2cm} \\
    \affaddr{\affmark[1]Graz University of Technology}, \affaddr{\affmark[2]University of Stuttgart}, \affaddr{\affmark[3]University of Duisburg-Essen}, \\
    \affaddr{\affmark[4]Joanneum Research}, \affaddr{\affmark[5]Flinders University}
    \vspace{-0.2cm}
}

\begin{document}

\maketitle

\begin{abstract}
Diminished reality (DR) refers to the removal of real objects from the environment by virtually replacing them with their background. Modern DR frameworks use inpainting to hallucinate unobserved regions. While recent deep learning-based inpainting is promising, the DR use case is complicated by the need to generate coherent structure and 3D geometry (\ie, depth), in particular for advanced applications, such as 3D scene editing. In this paper, we propose DeepDR, a first RGB-D inpainting framework fulfilling all requirements of DR: Plausible image and geometry inpainting with coherent structure, running at real-time frame rates, with minimal temporal artifacts. Our structure-aware generative network allows us to explicitly condition color and depth outputs on the scene semantics, overcoming the difficulty of reconstructing sharp and consistent boundaries in regions with complex backgrounds. Experimental results show that the proposed framework can outperform related work qualitatively and quantitatively.
\end{abstract}

\section{Introduction}
\begin{figure}[t]
  \centering
  \includegraphics[width=\hsize]{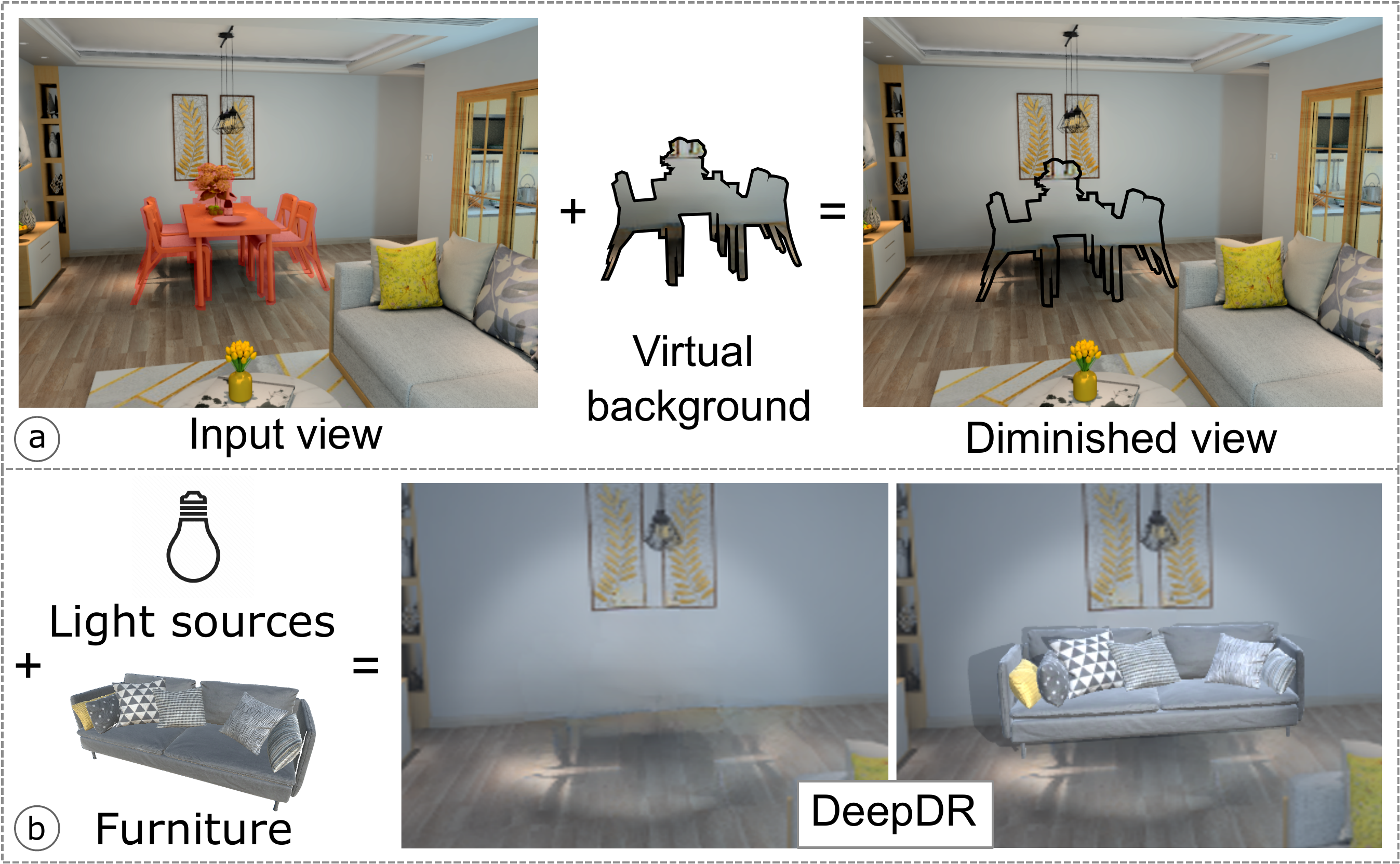}
  \caption{DR aims at replacing objects with their virtual background (a). DeepDR supports structure-aware RGB-D inpainting for DR experiences, enabling 3D scene editing by, \eg, adding light sources and replacing furniture (b).}
  \label{fig:abstract}
\vspace{-0.35cm}
\end{figure}

Diminished reality (DR) seeks to remove real objects from the environment by replacing them with their background~\cite{mori2017survey}, as illustrated in~\cref{fig:abstract}a. While \textit{multi-observational} approaches~\cite{li2013diminished, meerits2015real, mori2015efficient} can utilize existing information about the scene, \textit{inpainting} fabricates unseen background information and is, thus, more flexible.

Inpainting using generative adversarial networks (GANs) is nowadays successfully used in image space, \eg, to remove unwanted items during image and video editing~\cite{zeng2020high,li2022towards}. Contrary to conventional image or video inpainting, DR focuses on modifying a 3D scene rather than solely the image space, \eg, for removing 3D objects that distract from the immersive experience~\cite{korkalo2010light,kawai2016augmented, hashiguchi2018perceived}, or for replacing existing 3D objects with virtual ones in re-design~\cite{siltanen2014complete, sasanuma2016diminishing, zhang2016emptying, kari2021transformr}. In these scenarios, it is important to \textit{consider the underlying 3D geometry} of the scene for realistic rendering of virtual content and interactions with the hallucinated background, \eg, regarding occlusion and lighting (see ~\cref{fig:abstract}b). Thus, image space inpainting is not sufficient for DR applications -- depth information needs to be coherently inpainted as well~\cite{kawai2015diminished,mori2020inpaintfusion}.
Further, DR has strict requirements in \textit{adhering to the structural boundaries of the underlying scene}~\cite{gkitsas2021panodr,pintore2022instant}. This is conflicting with the tendency towards producing blurry results at ambiguous object boundaries and regions with mixed semantics, which is commonly seen in image inpainting CNNs~\cite{song2018spg, nazeri2019edgeconnect, liao2021image}. Lastly, unlike ordinary video inpainting~\cite{kim2019deep,liu2021fuseformer,li2022towards,zeng2020learning}, DR needs the ability to run in real-time, \textit{avoiding dissonance and flickering between consecutive frames}, without using future frame information.

In this paper, we propose DeepDR: The first approach to inpainting RGB-D frames with support for all aforementioned criteria of DR applications (see~\cref{tab:relwork} for a structured comparison to the state-of-the-art). DeepDR~has been designed as an end-to-end GAN, which performs inpainting of color images and their corresponding depth maps simultaneously. To enforce sharp structures with coherent semantics, we explicitly condition our model on the segmentation of the scene. 
To this end, we propose a novel structure-aware RGB-D decoder, which ensures adherence to the underlying structural boundaries. To further limit temporal artifacts over a series of frames, we adopt a simple, yet effective, recurrent strategy based on convolutional long short-term memory (ConvLSTM)~\cite{lai2018learning}. Thus, our framework produces temporally and structurally coherent inpainting. We emphasize inpainting the depth channel to ensure preserving a coherent 3D structure of the scene, which enables a realistic user experience in DR. Compared to related approaches, which rely on completing the various inpainting tasks in a sequential manner, our system processes inputs simultaneously, allowing each sub-task to benefit from the others. This allows to learn a comprehensive scene understanding, leading to a more plausible and consistent inpainting. We evaluate DeepDR~in the context of interior re-design, a quintessential DR application, and we show that it can outperform previous methods qualitatively and quantitatively.
In summary, we make the following contributions:
\setlist[itemize]{leftmargin=*}
\begin{itemize}
    \item We propose the first GAN for inpainting the color and depth channels of a DR system, which is capable of maintaining temporal and structural consistency. 
    
    \item We introduce a novel structure-aware RGB-D decoder that supports generating sharp and plausible structures.
    
    \item We qualitatively and quantitatively evaluate DeepDR for indoor and outdoor DR applications, by applying it to synthetic and real data.
\end{itemize}

\section{Related work}
\label{sec:related_work}

\textbf{Image inpainting.} Data-driven inpainting using deep learning leverages information from large databases. By implicitly or explicitly learning about the semantics of the scene, deep learning can produce high-quality results, spatially consistent with the image content. The seminal work of Context Encoders \cite{pathak2016context} first demonstrated the potential of a generative adversarial network (GAN) for image inpainting. Subsequent methods improve this approach, \eg, using coarse-to-fine nets~\cite{iizuka2017globally}, attention~\cite{yu2018generative,liu2019coherent,yi2020contextual}, iterative refinement~\cite{yang2017high,zeng2020high,li2020recurrent} or feature fusion~\cite{zeng2019learning,liu2020rethinking}. Partial or gated convolutions~\cite{liu2018image,yu2019free} enable the handling of irregular masks without introducing artifacts, an important capability that we utilize in our work. Recently, diffusion-based inpainting delivers visually impressive results~\cite{lugmayr2022repaint,rombach2022high}. However, their inference times of several seconds up to hours prohibit an application for real-time video, such as DR~\cite{lugmayr2022repaint}.

\textbf{RGB-D inpainting.} Depth inpainting literature largely focuses on filling missing depth values in regions \textit{visible} in RGB images, for compensating failures of common depth sensors, \eg,  at transparent, reflective, or distant surfaces~\cite{liu2016robust,xue2017depth,Zhang_2018_CVPR,huang2019indoor,zhang2022indepth}. Depth inpainting of \textit{hidden} structures, \eg,  in diminished parts of a scene, has been considered in only few works so far~\cite{dhamo2019peeking,fujii2020rgb,bevsic2020dynamic,pintore2022instant}. Earlier works~\cite{dhamo2019peeking,fujii2020rgb} explore different fusion strategies of RGB and depth information but do not leverage structural guidance or temporal consistency. DynaFill~\cite{bevsic2020dynamic} relies on a sequential approach, where the color domain is coarsely inpainted and a separate depth completion network obtains geometry. For maintaining temporal consistency, it requires odometry, \ie, camera poses. This has many pitfalls, as each sub-task relies on the results from the previous step, comes with significant computational overhead and is difficult to deploy. Pintore et al.~\cite{pintore2022instant} focus on the arguably simpler task of completely emptying rooms, while we also want to inpaint regions with complex and mixed semantics. Further, they do not deal with frame-to-frame consistency.

\textbf{Structural priors.} An ongoing challenge in inpainting is the reconstruction of sharp boundaries and structures consistent with the surrounding context, especially in regions with mixed semantics, where object boundaries are ambiguous. These structures are particularly important in DR, where interactivity with the scene is desired~\cite{gkitsas2021panodr,pintore2022instant}.
Structural priors, such as edges~\cite{nazeri2019edgeconnect}, contours~\cite{xiong2019foreground} or semantic segmentation~\cite{song2018spg,liao2021image,ardino2021semantic,gkitsas2021panodr} can guide the inpainting of images. Amongst them, PanoDR~\cite{gkitsas2021panodr} also targets an application in DR. However, their framework does not consider temporal coherence and 3D geometry. Sequential frameworks, which first complete the structural image, and feed it to the image generation network, are common. However, recent advances in image-to-image translation show that semantic information at the input of a generator may vanish through multiple downsampling and normalization stages~\cite{park2019semantic}. Hence, simultaneous frameworks for completing structure and texture at the same time have become popular~\cite{liao2021image,ardino2021semantic,gkitsas2021panodr}. Inspired by these successes, we incorporate explicit structural guidance via intermediate semantic segmentation and extend it to the depth domain.
\begin{table}[t]
  \centering
  \caption{Overview of current deep inpainting works for DR.}
  \small
    \begin{tabular}{lcccc}
          & Color & Depth & Structure & Temporal \\
    \toprule
    TransfoMR~\cite{kari2021transformr} & \cmark     & \xmark    & \xmark    & \cmark \\
    DynaFill~\cite{bevsic2020dynamic} & \cmark     & \cmark    & \xmark   & \cmark \\
    PanoDR~\cite{gkitsas2021panodr} & \cmark     & \xmark    & \cmark    & \xmark \\
    Pintore et al.~\cite{pintore2022instant} & \cmark & \cmark & \xmark & \xmark \\
    \midrule
    DeepDR (Ours) & \cmark  & \cmark     & \cmark     & \cmark \\
    \bottomrule
    \end{tabular}%
  \label{tab:relwork}%
\vspace{-0.2cm}
\end{table}%
\begin{figure*}[t]
    \centering
    \includegraphics[width=0.9\linewidth]{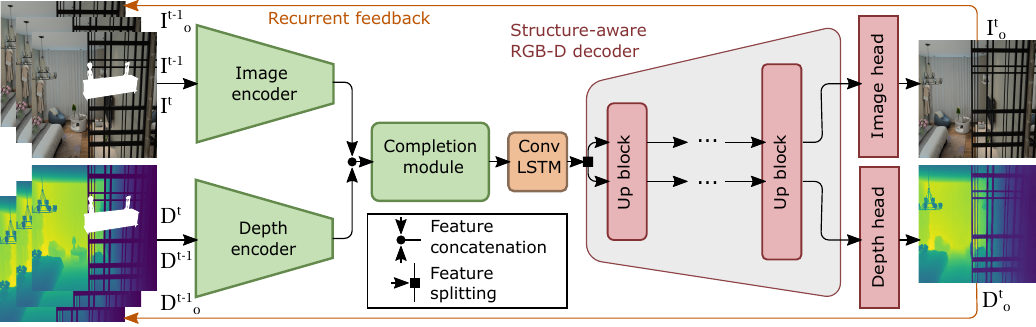}
    \caption{Overview of the proposed network. Image $I$ and depth $D$ inputs are encoded separately, before fusing features on a higher dimension and jointly completing them. Our semantics-aware decoder consists of a series of up blocks in which features are conditioned on semantic information. Finally, the outputs $I_o^t$ and $D_o^t$ are fed as auxiliary inputs to the next time step in a recurrent feedback loop.}
    \label{fig:architecture_overview}
\end{figure*}

\textbf{Temporal consistency.} Video inpainting attempts to extend image inpainting to the temporal domain to ensure frame-to-frame consistency. Several approaches use 3D convolutions~\cite{wang2019video,kim2019deep,chang2019free,chang2019learnable}, attention~\cite{lee2019copy,liu2021fuseformer,zeng2020learning} TransforMR~\cite{kari2021transformr} shows that deployment and real-time performance of some of these methods are feasible on mobile devices for DR. However, depth is not considered, which leads to shadow and occlusion artifacts. Diffusion-based techniques tend to be computationally expensive, requiring preprocessing and fine-tuning, which renders them impractical for real-time applications~\cite{ceylan2023pix2video}. Another direction in video inpainting are optical flow-based methods, which emerge as most promising~\cite{lai2018learning,xu2019deep,gao2020flow,li2022towards}. Albeit flow-based methods show impressive results for tasks such as video editing, most cannot be applied directly to DR. They rely on forward and backward flow, requiring knowledge about past and future frames, which is not available in online scenarios. Furthermore, optical flow is expensive to compute. In our framework, we utilize optical flow only during training and rely on a recurrent network to reduce temporal artifacts. 

\section{Method}

This section outlines the architecture of DeepDR. Our dual-stream encoder (\cref{sec:encoder}) extracts contextual features from masked images masked depth separately at shallow layers, then fuses and jointly completes them. A structure-aware decoder (\cref{sec:decoder}) uses two task-specific feature streams for RGB and depth with shared parameters. It estimates a semantic segmentation map from deep features and uses this map to modulate the RGB and depth feature generation. Thus, it is able to produce high-quality images and depths with a coherent semantic structure, which is persistent over domains and contexts. Finally, to reduce temporal artifacts between consecutive frames, we use a recurrent feedback loop with a ConvLSTM layer (\cref{sec:temp}). An overview of our model is given in \cref{fig:architecture_overview}. In the following, we explain our core components. Further architectural details are given in the supplementary material \cref{sec:architecture_suppl}.
\subsection{Dual-stream encoder and completion module}
\label{sec:encoder}
Recent findings in image inpainting suggest that deep features in a CNN contain the majority of structural information, while shallow layers contain textural information~\cite{liu2020rethinking}. Since RGB and depth inputs are texturally different, but represent the same underlying structure, we encode RGB and depth in two separate but parallel streams (illustrated by the green trapezoid in ~\cref{fig:architecture_overview}). The encoders use a coarse-to-fine architecture, which has proven to be highly effective for inpainting tasks~\cite{yu2018generative,liu2019coherent,yi2020contextual}. After extracting features through $l\in [1, \dots, L]$ fine layers, we place a completion module to fuse them, such that the network can complete RGB and depth inputs simultaneously and thus, coherently. The completion module consists of a series of dilated convolutions~\cite{yu2017dilated} to expand the receptive field of the network and efficiently utilize global information. We use gated convolution layers~\cite{yu2018generative} throughout our encoder and completion module, which dynamically learn to select appropriate features from masked and unmasked regions.
\subsection{Structure-aware RGB-D decoder}
\label{sec:decoder}
\begin{figure*}[ht]%
    \centering
    \includegraphics[width=\linewidth]{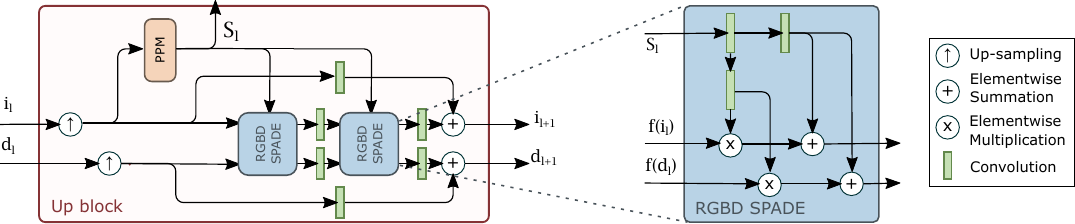}
    \caption{Our up blocks contain a residual architecture. From the up-sampled image feature $i_l$, a semantic segmentation map $S_l$ is predicted using a pyramid pooling module (PPM). This segmentation is fed into the RGB-D SPADE, together with $i_l$ and $d_l$. Within the RGB-D SPADE, the semantic map is embedded to feature space and convolved to obtain a learned, spatial scaling and bias map, with which inputs are transformed. Thus, the output features are consistently modulated by semantic segmentation.}
    \label{fig:upblock}
\vspace{-0.2cm}
\end{figure*}%
Our structure-aware RGB-D decoder is inspired by the spatially-adaptive normalization (SPADE) principle~\cite{park2019semantic}. SPADE aims to overcome the problem of vanishing semantics, where sequential convolution, non-linearity, and normalization operations in a traditional CNN ``wash away'' structural information. It conditions generated features directly on semantic priors, by modulating them in normalization layers using a learned transformation. The same principle can be applied to other image-to-image translation tasks, such as inpainting. Our approach further exploits the fact that the RGB and depth inputs share the same underlying semantics -- thus, we extend SPADE for RGB-D inpainting. Our decoder consists of a series of $L$ up-sampling blocks based on residual learning (ResNet)~\cite{he2016deep}, which consist of two RGB-D SPADE layers with intermediate convolutions and a skip connection (see \cref{fig:upblock}a). Each up block receives an upsampled RGB and depth feature, $i_l$ and $d_l$, from the previous layer. From the RGB feature, we explicitly model the underlying scene semantics by predicting a segmentation map on the current feature scale, $S_l$, using a pyramid pooling module~\cite{zhao2017pyramid} (see supplementary \cref{sec:seg_suppl} for examples). This map, together with $up(i_l)$ and $up(d_l)$, are forwarded to the RGB-D SPADE (\cref{fig:upblock}b). Within the RGB-D SPADE, the segmentation map is embedded into feature space, and convolved to obtain learned, spatial modulation parameters. The RGB and depth features are transformed by these parameters,  conditioning them on the semantic information. The parameters of RGB-D SPADE layers are shared between the RGB and depth streams, ensuring consistent semantic structures in both of them. 
\subsection{Maintaining temporal consistency}
\label{sec:temp}
As an alternative to expensive video inpainting techniques, we use a simple, yet effective method based on a recurrent network and ConvLSTM~\cite{shi2015convolutional}, originally proposed for blind video temporal consistency~\cite{lai2018learning}, and adapt it for RGB-D inpainting. Compared to video inpainting, which usually uses past and future frames,
it allows our network to process frames in a sequential, online manner, \eg, from $t=1$ to $T$. At every time step $t$, our network additionally receives the previous input image $I^{t-1}$ and depth $D^{t-1}$, as well as their corresponding outputs, $I^{t-1}_o$ and $D^{t-1}_o$, as auxiliary information. A ConvLSTM layer at the end of our completion module captures spatio-temporal correlations between consecutive frames in the feature space. While we use the optical flow between frames during training (see \cref{sec:objectives} for details), our method does not require flow at inference time. Thus, it is very efficient. Furthermore, we can process inputs of arbitrary length -- be it single frames (in which case we set $I^{t-1} = I^{t},\, D^{t-1} = D^{t}$) or long video sequences. 

\subsection{Training objectives}
\label{sec:objectives}

\begin{figure}[ht]
    \centering
    \includegraphics[width=\linewidth]{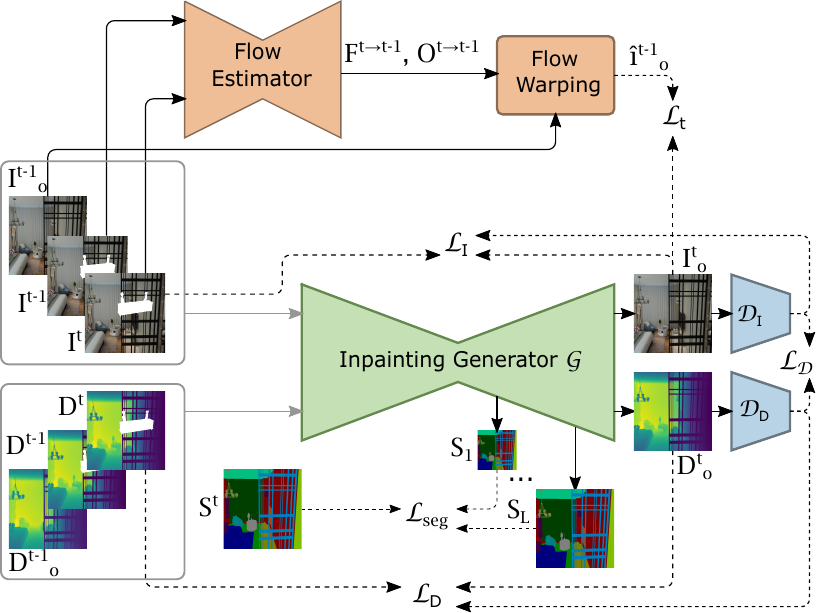}
    \caption{Illustration of training approach. We use appropriate loss terms and adversarial learning to promote accuracy and realism of inpainted RGB and depth ($\mathcal{L}_{I}$, $\mathcal{L}_{D}$). Via flow estimation and warping, we compute a temporal loss $\mathcal{L}_{t}$ to ensure temporal consistency. Finally, we enforce our model to learn structural information by additionally providing semantic supervision via $\mathcal{L}_{seg}$.}
    \label{fig:training_overview}
\end{figure}

Our generator $\mathcal{G}$ is trained with a combined loss function, which contains terms for supervising image inpainting, depth inpainting, semantic segmentation and temporal coherence (see~\cref{fig:training_overview}):
\begin{equation}
    \mathcal{L}_{G} = \mathcal{L}_{I} + \mathcal{L}_{D} + \mathcal{L}_{seg} + \mathcal{L}_{temp}.
\end{equation}%

\textbf{Adversarial learning.}
On top of our generator, we use two global PatchGAN discriminators~\cite{isola2017image}, $\mathcal{D}_I$ and $\mathcal{D}_D$, to distinguish between real and inpainted RGB and depth patches. Thus, our network is trained in an adversarial fashion. We use Hinge loss~\cite{lim2017geometric} to compute our losses for training the discriminator, $\mathcal{L}_{\mathcal{D,I}}$ and $\mathcal{L}_{\mathcal{D,D}}$, as well as adversarial generator loss terms $\mathcal{L}_{adv,I}^G$ and $\mathcal{L}_{adv,D}^G$. 

\textbf{Image inpainting.}
We use the $\ell_1$-reconstruction loss $\mathcal{L}_{rec,I}$ between synthesized pixels and the ground truth to ensure pixel-level reconstruction for image inpainting. Further, we use the perceptual loss $\mathcal{L}_{per}$~\cite{johnson2016perceptual} and style loss $\mathcal{L}_{sty}$~\cite{gatys2016image} to encourage the network to produce RGB images perceptually similar to the ground truth. These data-driven losses enforce similarity in the feature space. Perceptual loss penalizes differences in features directly, while style loss minimizes the difference between feature distributions, de-localizing the feature information. Thus, image inpainting is supervised by the objective
\begin{equation}
    \mathcal{L}_{I} = \lambda_{rec} \mathcal{L}_{rec,I} + \lambda_{per} \mathcal{L}_{per} + \lambda_{sty} \mathcal{L}_{sty} +  \mathcal{L}_{adv,I}^G.
\end{equation}%

\textbf{Depth inpainting.}
For depth inpainting, we again use the $\ell_1$-reconstruction loss $\mathcal{L}_{rec,D}$ to penalize individual pixel errors. However, this loss does not take the local pixel neighborhood into account, which can lead to blurry edges and discontinuous surfaces in reconstructed depth images. Hence, to encourage smooth depth predictions with sharp steps, we use a gradient-based loss term 
\begin{equation}
    \mathcal{L}_{grad} = ||\nabla D - \nabla D_o||_1,
\end{equation}%
where $\nabla$ is the Sobel operator. Thus, depth loss is
\begin{equation}
    \mathcal{L}_{D} = \lambda_{rec} \mathcal{L}_{rec,D} + \lambda_{grad} \mathcal{L}_{grad} + \mathcal{L}_{adv,D}^G.
\end{equation}%

\textbf{Semantic segmentation.}
By predicting intermediate semantic segmentations in our structure-aware decoder, we ensure that our model explicitly learns semantic information from inputs. For supervising this prediction, we compute the cross-entropy loss for each intermediate segmentation map $S_{l}$, upsampled to the original input resolution, with the ground truth segmentation $S$, 
\begin{equation}
    \mathcal{L}_{seg} = -\lambda_{seg} \frac{1}{L} \sum_{l\in L} \sum_{j\in S} S^j \log (up(S_{l}^{j})),
\end{equation}
where $j$ denotes the classes of the segmentation map.

\textbf{Temporal coherence.}
Similar to Lai \etal \cite{lai2018learning}, we enforce temporal coherence by training our model with short-term ($st$)  and long-term ($lt$) temporal losses between current and previous, and current and first frame, respectively:
\begin{align}
 \begin{split}
    \mathcal{L}_{st} = \sum_{t=2}^{T} O^{t\rightarrow t-1} ||I^t_o - \hat{I}^{t-1}_o||_1, \\ \mathcal{L}_{lt} = \sum_{t=3}^{T} O^{t \rightarrow 1} ||I^t_o - \hat{I}^{1}_o||_1.     
 \end{split}
\end{align}
Here, $\hat{I}^{x}_o$ is the output image at time $x$, warped to the current time point using the optical flow $F^{t \rightarrow x}$ between $I^t$ and $I^{x}$, and $O^{t \rightarrow x}$ is the corresponding occlusion mask (see~\cref{fig:training_overview}). In our framework, during training, $F$ and $O$ are computed using MaskFlownet~\cite{zhao2020maskflownet}. The temporal loss term is
\begin{equation}
    \mathcal{L}_{temp} = \lambda_{t} (\mathcal{L}_{lt} + \mathcal{L}_{st}).
\end{equation}%
A full description of objectives and training details are found in the supplementary material \cref{sec:training_suppl} and \cref{sec:training_suppl_2}.

\section{Experiments and results}

\subsection{Datasets}

While several benchmarks exist for RGB inpainting~\cite{zhou2017places,karras2017progressive}, there are few datasets suitable for RGB-D object removal and DR (see supplementary \cref{sec:datasets_suppl}). To the best of our knowledge, DynaFill~\cite{bevsic2020dynamic} is the sole dataset that offers ground truth by presenting scenes both with and without individual objects that need to be removed. Same as in the original paper, we extract masks from dynamic objects and use the default training and validation split. Since DynaFIll only covers outdoor driving scenarios with limited variability, we additionally evaluate our method on InteriorNet~\cite{li2018interiornet}, were, similar to other works~\cite{kim2019deep,liu2021fuseformer,li2022towards}, we simulate the object removal task by inpainting random object-like masks during training and testing (see supplementary \cref{fig:dr:testset_interiornet}, \cref{fig:dr:testset_scannet}). Specifically, those masks are generated from instance segmentations belonging to non-background classes (\ie, excluding walls, ceilings, floors, windows, and doors). We split the 618 layouts in InteriorNet into 494 for training and 62 for testing and validation. To show the generalizability of our model and demonstrate its performance in-the-wild, we further use 100 layouts from ScanNet~\cite{dai2017scannet} for testing the models trained on InteriorNet.

\subsection{Comparison with other methods}

As already mentioned, only few works about RGB-D inpainting of hidden structures are known to us~\cite{dhamo2019peeking,fujii2020rgb,bevsic2020dynamic,pintore2022instant}. Only DynaFill~\cite{bevsic2020dynamic} is accessible, although training and testing code are not provided. Therefore, we re-compute results and metrics on their dataset using the publicly available demo model. Sequential frameworks, where RGB information is completed first, and missing depth information is filled based on the reconstructed image using depth completion, are an alternative for DR~\cite{kari2021transformr,pintore2022instant}. Hence, we build our baselines on top of recent RGB inpainting methods, and use state-of-the-art depth completion networks, InDepth~\cite{zhang2022indepth}, DM-LRN~\cite{senushkin2021decoder} and NLSPN~\cite{park2020non}, to fill missing depth regions from the RGB inpainting. Hereafter, we use the best-performing depth completion method on each dataset for our comparison, which is InDepth for InteriorNet, and NLSPN for ScanNet and DynaFill. A detailed comparison is provided in the supplementary \cref{sec:depth_compl_suppl}. Based on performance and code availability, we compare to 
DeepFillV2~\cite{yu2019free}, PanoDR~\cite{gkitsas2021panodr} and E2FGVI~\cite{li2022towards}, which represent standard, structure-guided, and video inpainting, respectively. For a fair comparison, we re-train the models on our datasets using their publicly available training code.

\subsection{Quantitative results}
To quantitatively assess the performance of our approach, we use the pixel-level metrics peak signal-to-noise ratio (PSNR) and mean absolute error (MAE) for image, and root mean squared error (RMSE) in meters for depth inpainting. However, these metrics only measure pixel-wise concordance and tend to favor blurry over perceptually similar images, which is problematic for DR. Measures computed on deep features better mirror human perception~\cite{zhang2018unreasonable} and are, thus, considered more meaningful for our evaluation. We use learned perceptual image patch similarity (LPIPS)~\cite{zhang2018unreasonable} and Fr\'echet inception distance (FID)~\cite{heusel2017gans} for images, and video FID (VFID)~\cite{wang2018video} for sequences. We further compare efficiency by measuring inference time, multiply-add operations (MADs) and total parameters.

It can be seen from~\cref{tab:comparison_interior},~\cref{tab:comparison_dyna} and~\cref{tab:comparison_scan} that DeepDR outperforms all related methods in the feature-based inpainting metrics, on indoor (InteriorNet), outdoor (DynaFill), as well as real, unseen (ScanNet) data. As mentioned, we consider these metrics to be most significant for DR. Considering depth RMSE, it is evident that our joint framework outperforms both sequential methods, consisting of image and depth-from-image inpainting, as well as DynaFill by a large margin. In pixel-based RGB metrics, DeepDR comes second, after E2FGVI or DynaFill. The lead of E2FGVI is larger on ScanNet -- we attribute that to its tendency to produce overly smooth results, which matches the blurry images recurrent in ScanNet. DeepDR achieves leading results in video-based metrics as well, surpassing E2FGVI on InteriorNet and coming second on ScanNet. On DynaFill data, E2FGVI and DynaFill outperform DeepDR in VFID, but the increased temporal smoothness comes at the cost of increased blurriness, as shown by the lower FID and LPIPS, and the qualitative results. Contrary to E2FGVI, our method works for single images or very short sequences. No expensive flow computation is required at inference, making it almost one order of magnitude faster (see~\cref{tab:comparison_efficiency}), which is a critical factor for DR applications, where real-time frame rates are desired. Only DeepFillV2, which is the least powerful method in our tests, is faster than our model. DynaFill assumes availability of accurate camera poses and intrinsics, which may be difficult to obtain in real-world scenarios. 

\begin{table}[htb]
  \centering
  \caption{Quantitative comparison of inpainting models trained on InteriorNet~\cite{li2018interiornet}. For baselines, we use InDepth~\cite{zhang2022indepth} to fill missing depth. 
  }
  \small
  \resizebox{\columnwidth}{!}{
    \begin{tabular}{l |*{4}{c}|c|*{1}{c}}
    \toprule
     & \multicolumn{4}{c|}{RGB} & \multicolumn{1}{c|}{Depth} & \multicolumn{1}{c}{Video}\\
    \cmidrule{2-7} 
     Model & LPIPS $\downarrow$ & FID $\downarrow$ & PSNR $\uparrow$ & MAE $\downarrow$ & RMSE $\downarrow$ & VFID $\downarrow$ \\
    \midrule
    DeepFillV2~\cite{yu2019free} & 0.0150 & 0.448 & 41.6 & 0.0312 & 0.572 & 0.0446 \\
    PanoDR~\cite{gkitsas2021panodr} & \underline{0.0128} & 0.606 & 41.0 & 0.0331 & 0.564 & 0.0360 \\
    E2FGVI~\cite{li2022towards} & 0.0131 & \underline{0.363} & \textbf{43.2} & \textbf{0.0255} & \underline{0.563} & \underline{0.0326}\\
    \midrule
    DeepDR~(Ours)  & \textbf{0.0104} & \textbf{0.218} & \underline{41.9} & \underline{0.0311} & \textbf{0.278} & \textbf{0.0257}\\
    \bottomrule
    \end{tabular}}
    \label{tab:comparison_interior}%
\end{table}

\begin{table}[htb]
  \centering
  \caption{Quantitative comparison of inpainting models trained on DynaFill~\cite{bevsic2020dynamic}. For DeepFillV2~\cite{yu2019free}, PanoDR~\cite{gkitsas2021panodr} and E2FGVI~\cite{li2022towards}, we use NLSPN~\cite{park2020non} to fill missing depth. 
  }
  \resizebox{\columnwidth}{!}{
    \begin{tabular}{l |*{4}{c}|c|*{1}{c}}
    \toprule
     & \multicolumn{4}{c|}{RGB} & \multicolumn{1}{c|}{Depth} & \multicolumn{1}{c}{Video}\\
    \cmidrule{2-7} 
     Model & LPIPS $\downarrow$ & FID $\downarrow$ & PSNR $\uparrow$ & MAE $\downarrow$ & RMSE $\downarrow$ & VFID $\downarrow$ \\
    \midrule
    DeepFillV2~\cite{yu2019free} & 0.0238 & 4.122 & 34.2 & \underline{0.0062} & 7.92 & 1.185\\
    PanoDR~\cite{gkitsas2021panodr} & 0.0250 & 5.579 & 31.8 & 0.0119 &  8.12 & 1.822 \\
    E2FGVI~\cite{li2022towards} & \underline{0.0169}  & 2.826 & \underline{35.2} & \textbf{0.0054} & 7.83 & \underline{0.777} \\
    DynaFill~\cite{bevsic2020dynamic} & 0.0197 & \underline{2.665} & \textbf{38.8} & 0.0107 & \underline{7.78} & \textbf{0.636} \\
    \midrule
    DeepDR~(Ours) & \textbf{0.0168} & \textbf{2.415} & 34.2 & \underline{0.0062} & \textbf{4.51} & 0.788\\
    \bottomrule
    \end{tabular}}
  \label{tab:comparison_dyna}%
\end{table}%

\begin{table}[htb]
  \centering
  \caption{Generalizability experiment on ScanNet~\cite{dai2017scannet} of inpainting models trained on InteriorNet~\cite{li2018interiornet}. For baselines, we use NLSPN~\cite{park2020non} to fill missing depth. 
  }
  \small
  \resizebox{\linewidth}{!}{
    \begin{tabular}{l |*{4}{c}|c|*{1}{c}}
    \toprule
     & \multicolumn{4}{c|}{RGB} & \multicolumn{1}{c|}{Depth} & \multicolumn{1}{c}{Video} \\
    \cmidrule{2-7} 
    Model & LPIPS $\downarrow$ & FID $\downarrow$ & PSNR $\uparrow$ & MAE $\downarrow$ & RMSE $\downarrow$ & VFID $\downarrow$ \\
    \midrule
    DeepFillV2~\cite{yu2019free} & 0.0208 & 0.693 & 40.1 & 0.0400 & \underline{0.508} & 0.873 \\
    PanoDR~\cite{gkitsas2021panodr} & 0.0119 & 0.348 & 41.5 & 0.0304 & 0.536 & 0.358 \\
    E2FGVI~\cite{li2022towards} & \underline{0.0110} & \underline{0.295} & \textbf{46.7} & \textbf{0.0176} & 0.512 & \textbf{0.206} \\
    \midrule
    DeepDR & \textbf{0.0108} & \textbf{0.292} & \underline{42.4} & \underline{0.0280} & \textbf{0.484} & \underline{0.218} \\
    \bottomrule
    \end{tabular}}
  \label{tab:comparison_scan}%
\vspace{-0.25cm}
\end{table}%

\begin{table}[htb]
  \centering
  \caption{Efficiency of DeepDR on an NVIDIA GeForce GTX 1080 Ti GPU in comparison to the baselines.
  }
  \footnotesize
    \begin{tabular}{l |*{3}{c}}
    \toprule
    Model & Time $\downarrow$ (ms) & MADs $\downarrow$ & Params $\downarrow$ \\
    \midrule
    DeepFillV2~\cite{yu2019free} & \textbf{3.73} & \textbf{25.3 G} & \textbf{4.1 M}  \\
    PanoDR~\cite{gkitsas2021panodr} & 7.07 & 189.6 G & 78.8 M\\
    E2FGVI~\cite{li2022towards} & 40.0 & 309.1 G & 41.8 M \\
    DynaFill~\cite{bevsic2020dynamic}* & 14.3 & \underline{78.6 G} & \underline{22.1 M} \\
    \midrule
    DeepDR & \underline{4.43} & 184.3 G & 69.9 M \\
    \bottomrule
    \multicolumn{4}{l}{*Measurements do not include camera pose computation.}
    \end{tabular}
  \label{tab:comparison_efficiency}%
\vspace{-0.5cm}
\end{table}%

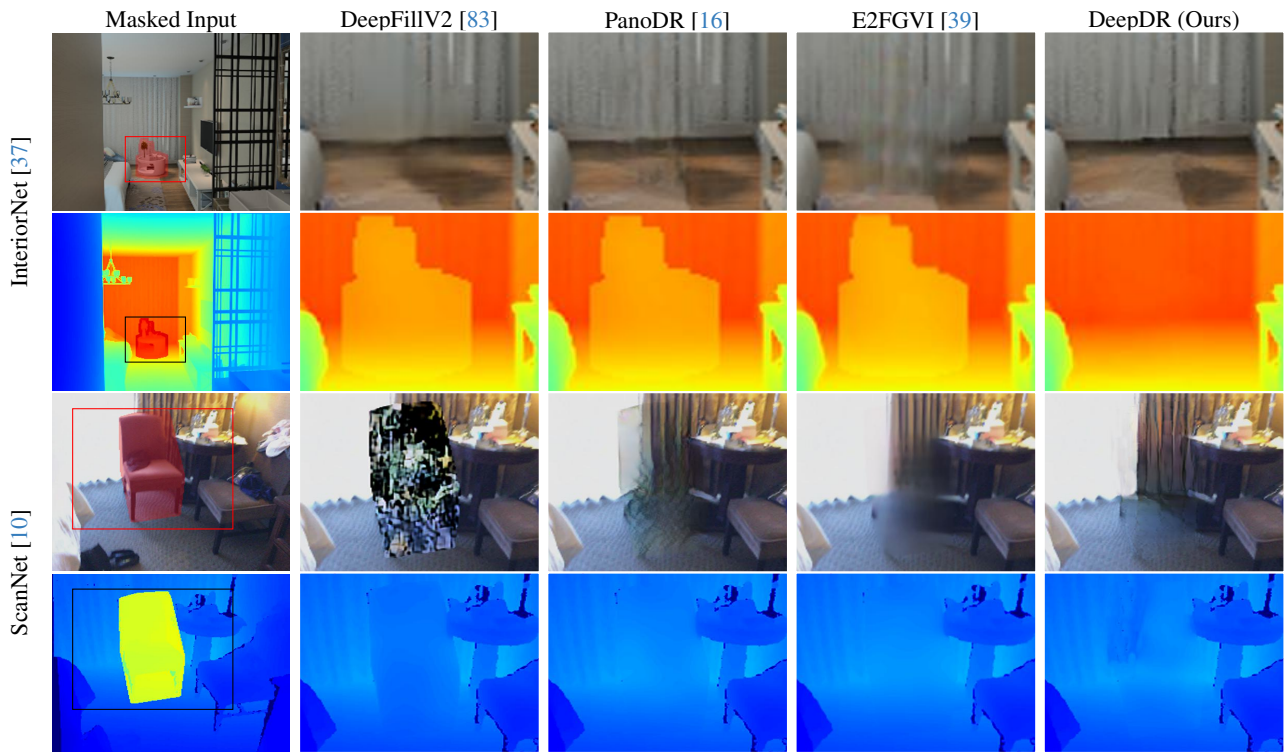
\begin{figure*}[ht]
    \noindent
    \begin{scaletikzpicturetowidth}{\textwidth}
    \begin{tikzpicture}[spy using outlines={magnification=4, rectangle, width=3.2cm, height=2.4cm, white}, remember picture]
        \draw (0,\myImgH / 1.8) node {\small{Masked Input}};
        \draw (\myImgW,\myImgH / 1.8) node {\small{DeepFillV2~\cite{yu2019free}}};
        \draw (2*\myImgW,\myImgH / 1.8) node {\small{PanoDR~\cite{gkitsas2021panodr}}};
        \draw (3*\myImgW,\myImgH / 1.8) node {\small{E2FGVI~\cite{li2022towards}}};
        \draw (4*\myImgW,\myImgH / 1.8) node {\small{DeepDR~(Ours)}};

        \draw (-2, -\myImgH / 2) node[rotate=90] {\small{InteriorNet~\cite{li2018interiornet}}};
        \node[tight] (n1) at (0,0) {\includegraphics[width=0.18\textwidth]{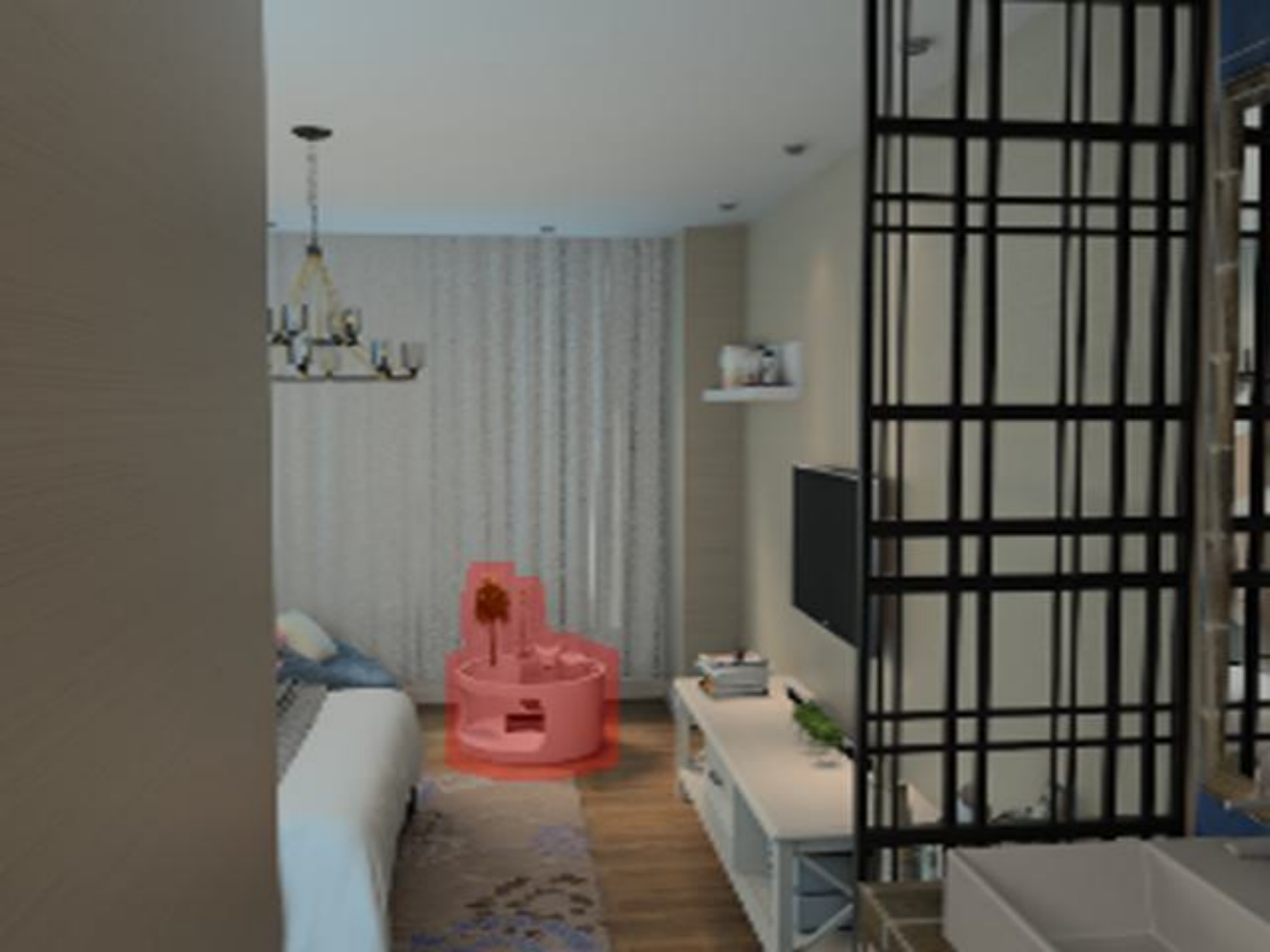}};
        \node[tight] (n2) at (0,-\myImgH)
        {\includegraphics[width=.18\textwidth,valign=m]{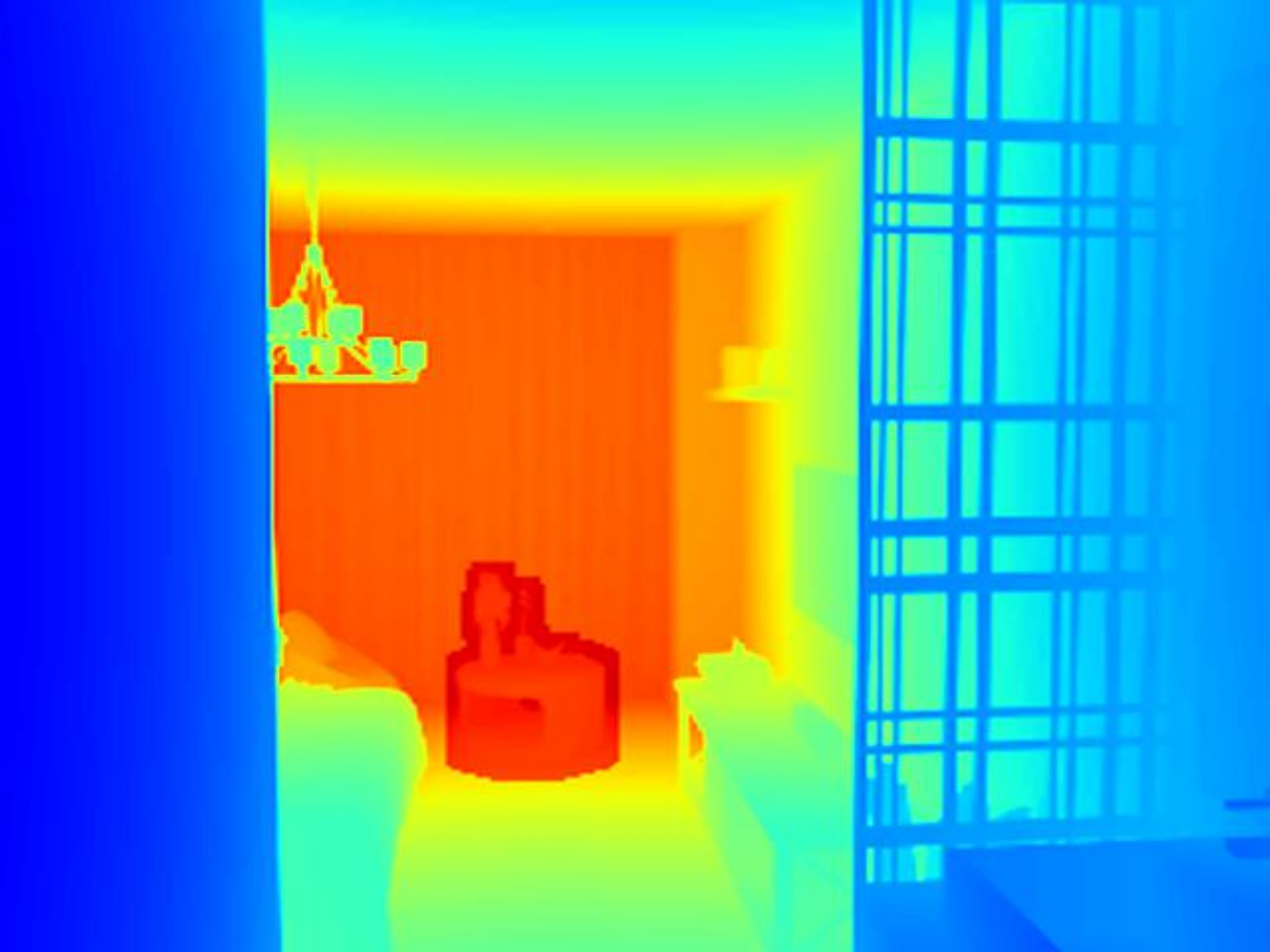}};
    
        \node[tight] (n3) at (\myImgW, 0) {\includegraphics[width=0.18\textwidth]{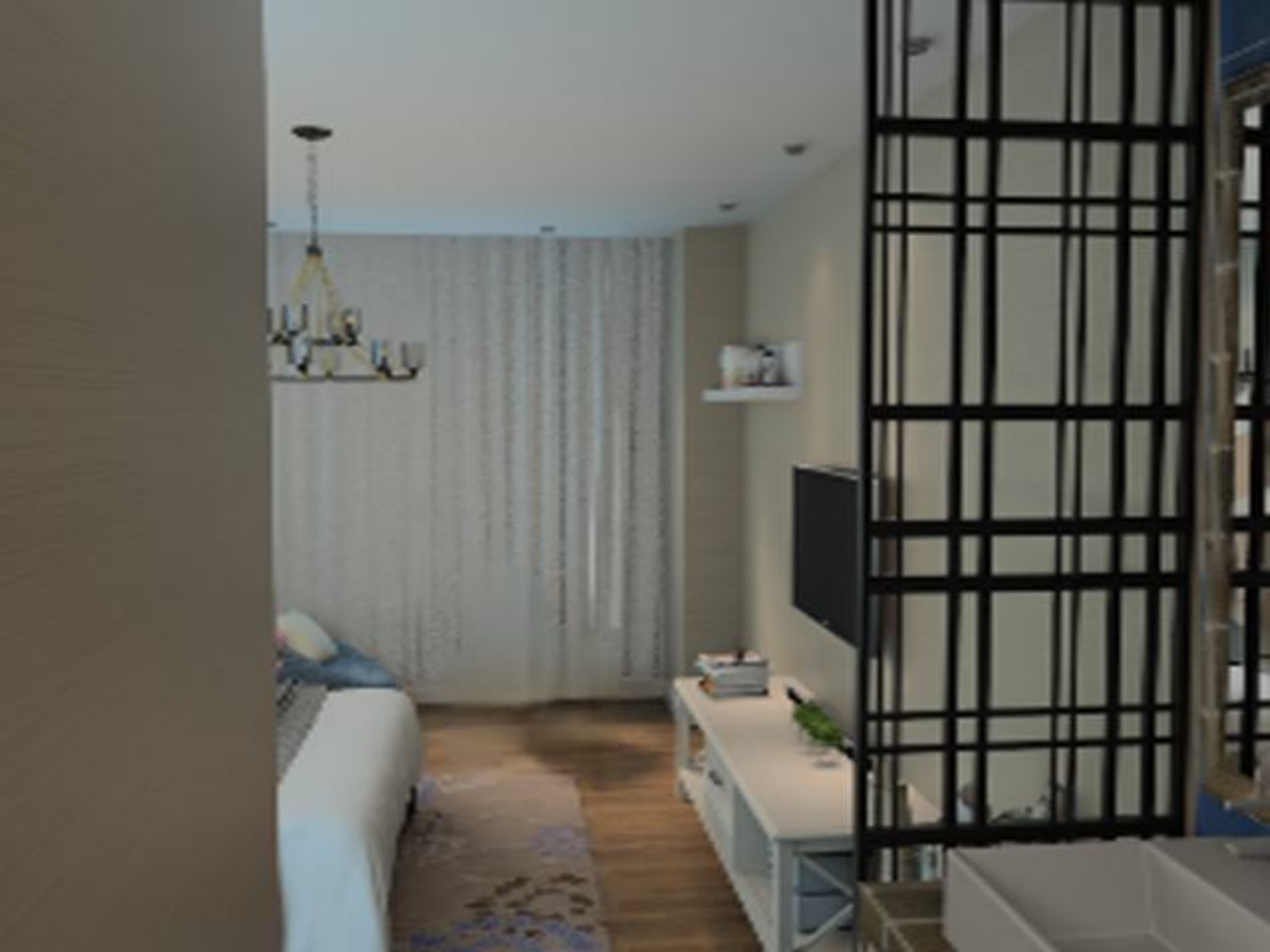}};
        \spy on (\myImgW-0.21,-0.50) in node at (\myImgW, 0);
        \node[tight] (n4) at (\myImgW, -\myImgH) {\includegraphics[width=.18\textwidth,valign=m]{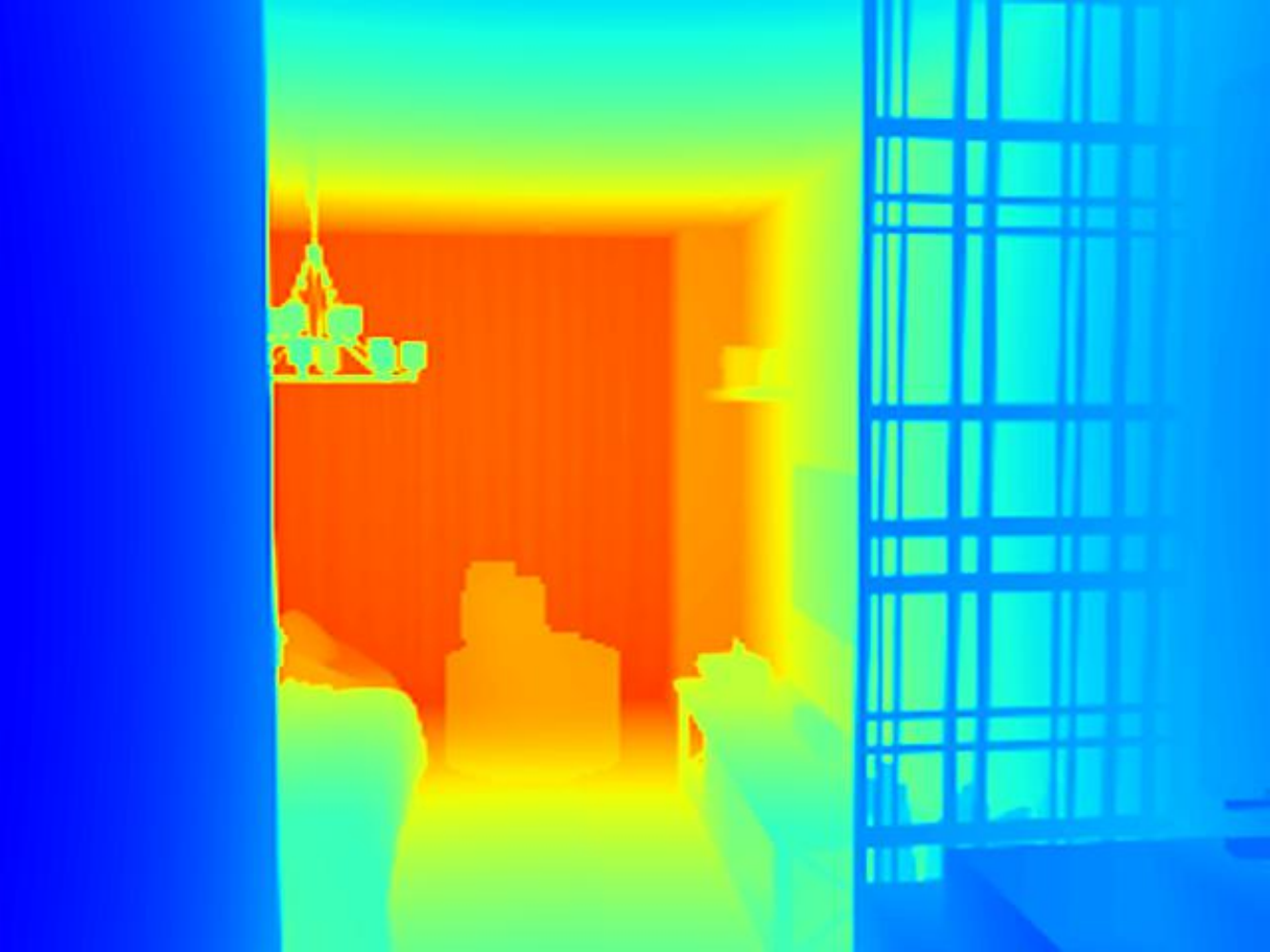}};
        \spy on (\myImgW-0.21, -\myImgH-0.50) in node at (\myImgW, -\myImgH);
        
        \node[tight] (n5) at (2*\myImgW,0) {\includegraphics[width=0.18\textwidth]{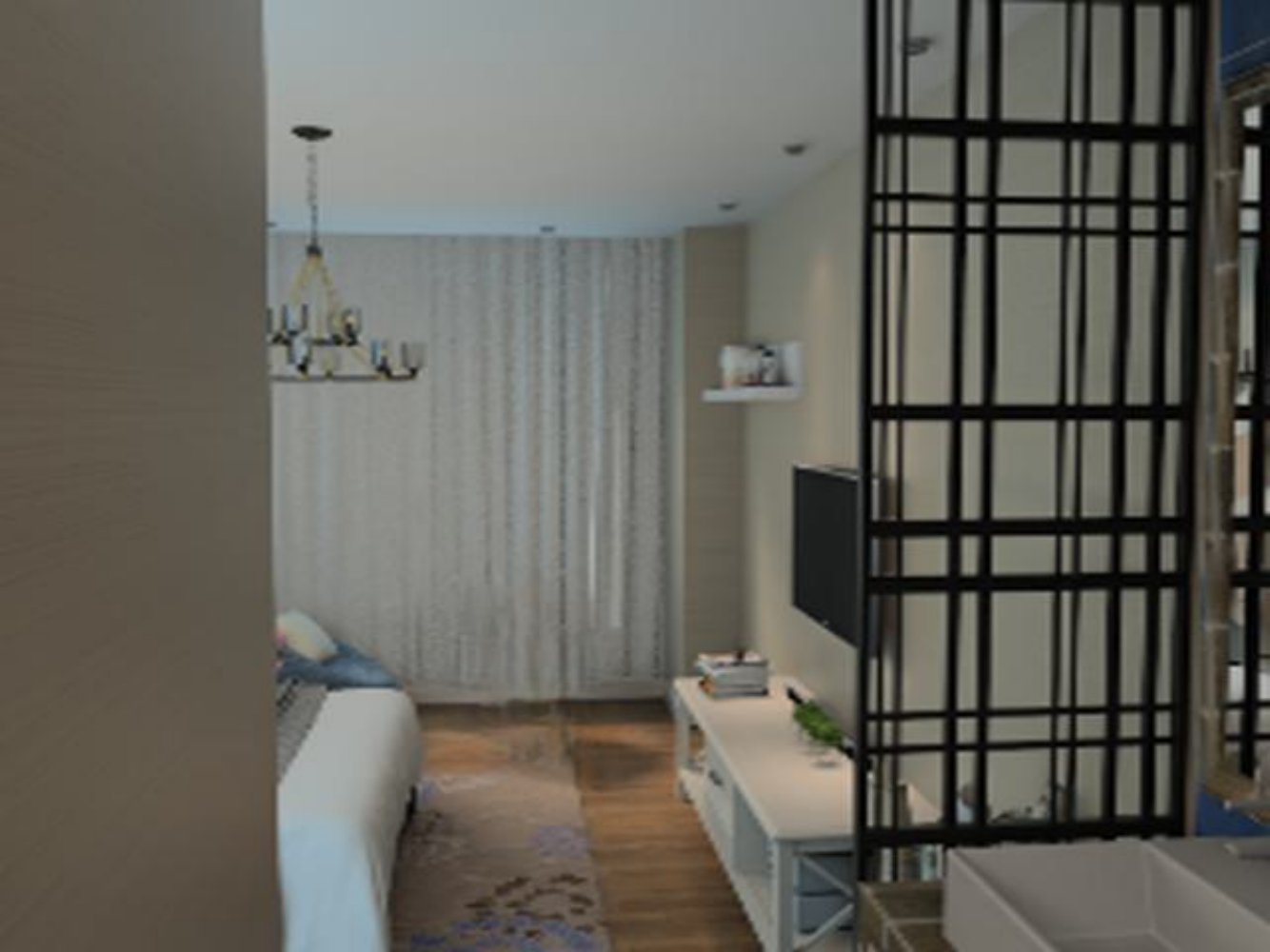}};
        \spy on (2*\myImgW-0.21,-0.50) in node at (2*\myImgW, 0);
        \node[tight] (n6) at (2*\myImgW, -\myImgH) {\includegraphics[width=.18\textwidth,valign=m]{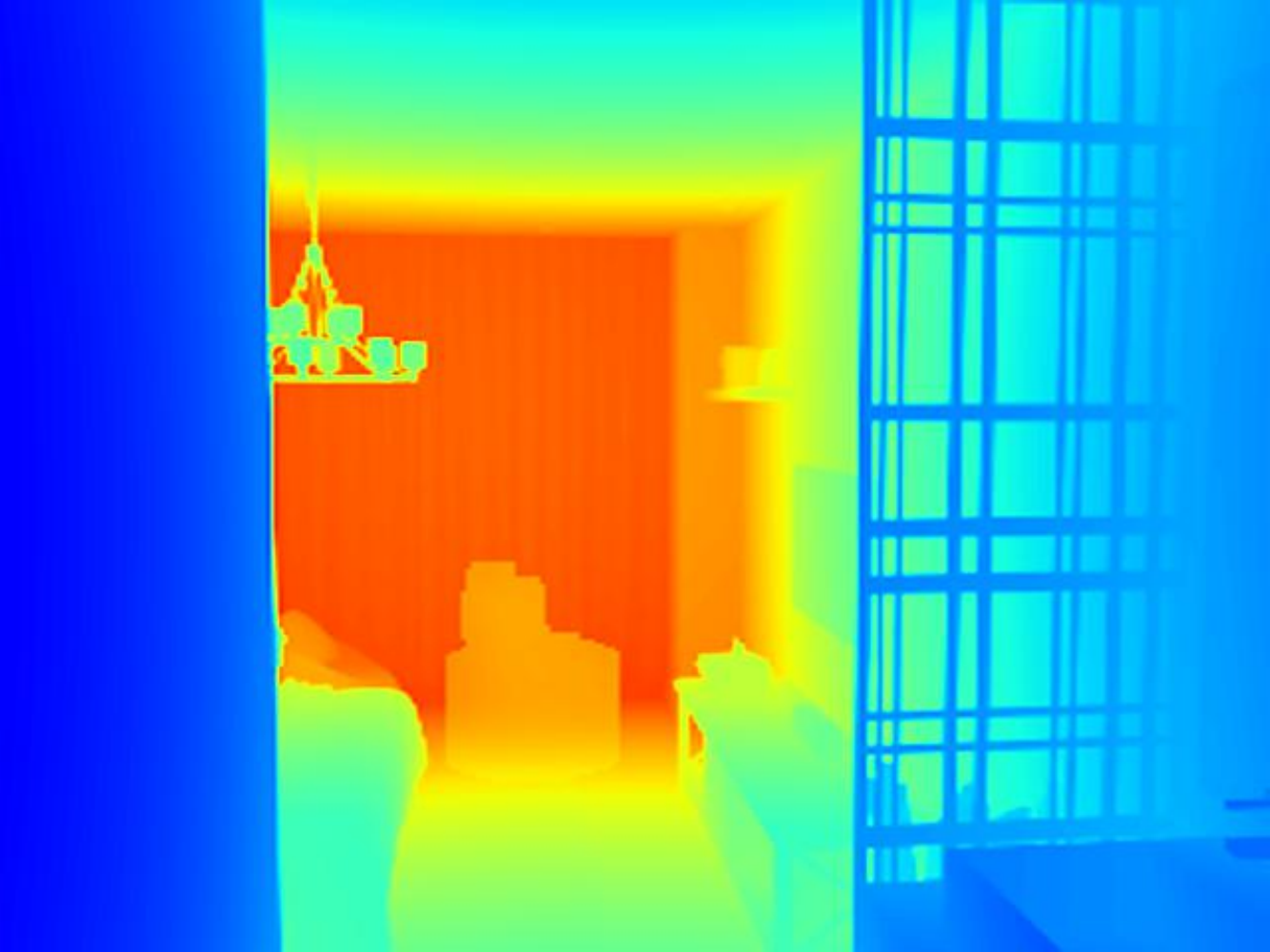}};
        \spy on (2*\myImgW-0.21, -\myImgH-0.50) in node at (2*\myImgW, -\myImgH);
        
        \node[tight] (n7) at (3*\myImgW,0) {\includegraphics[width=0.18\textwidth]{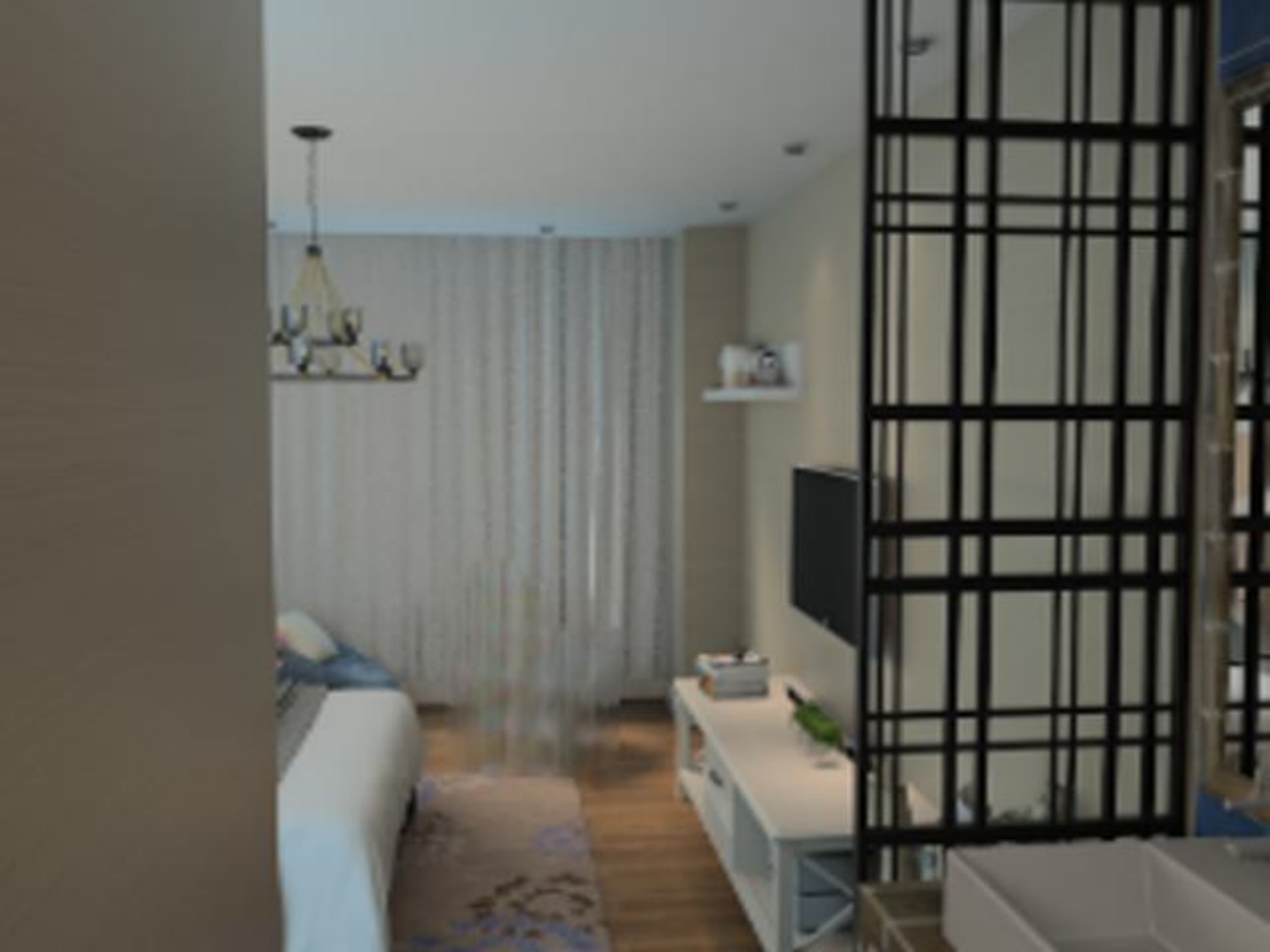}};
        \spy on (3*\myImgW-0.21,-0.50) in node at (3*\myImgW, 0);
        \node[tight] (n8) at (3*\myImgW, -\myImgH) {\includegraphics[width=.18\textwidth,valign=m]{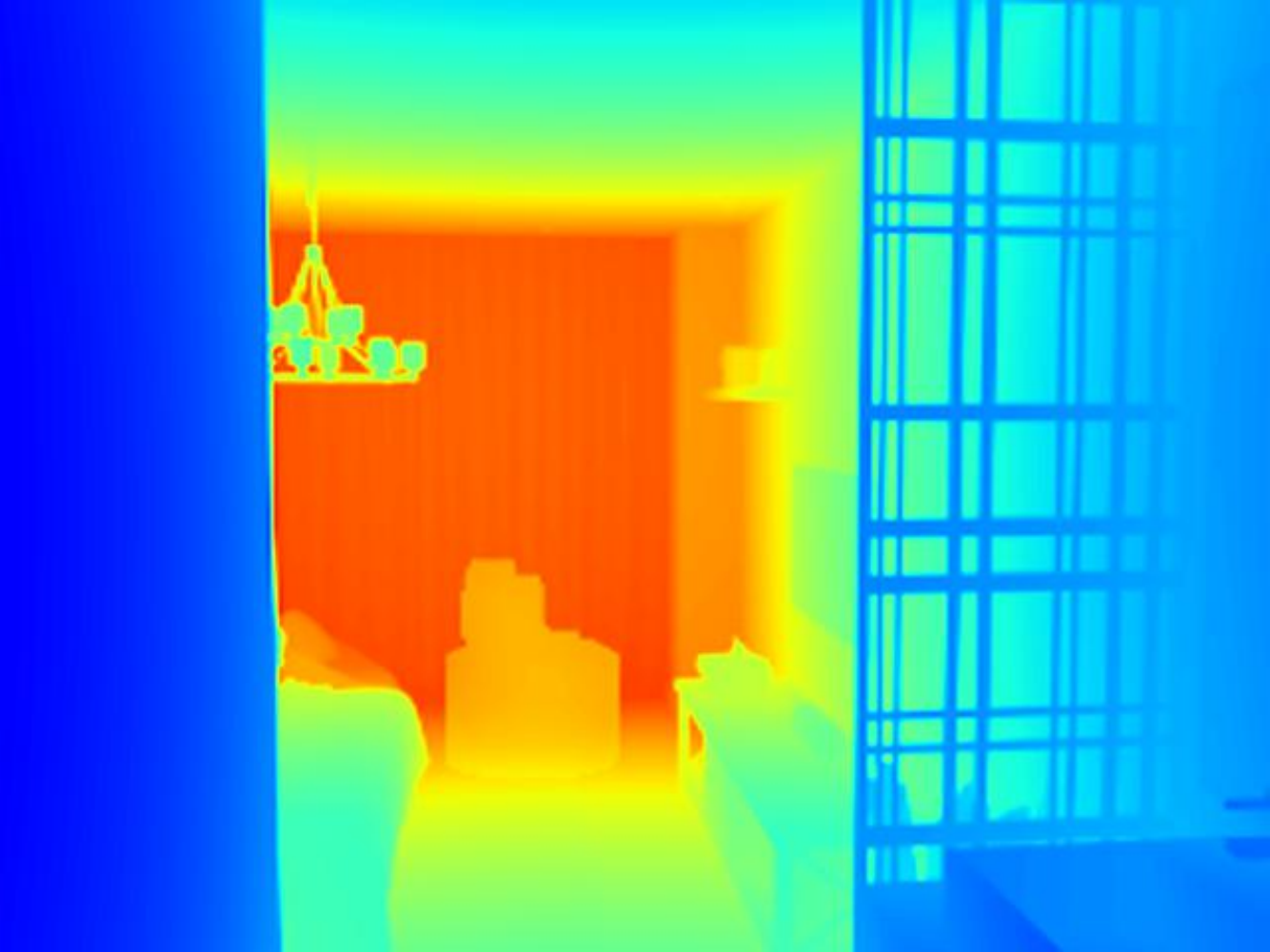}};
        \spy on (3*\myImgW-0.21, -\myImgH-0.50) in node at (3*\myImgW, -\myImgH);
        
        \node[tight] (n9) at (4*\myImgW,0) {\includegraphics[width=0.18\textwidth]{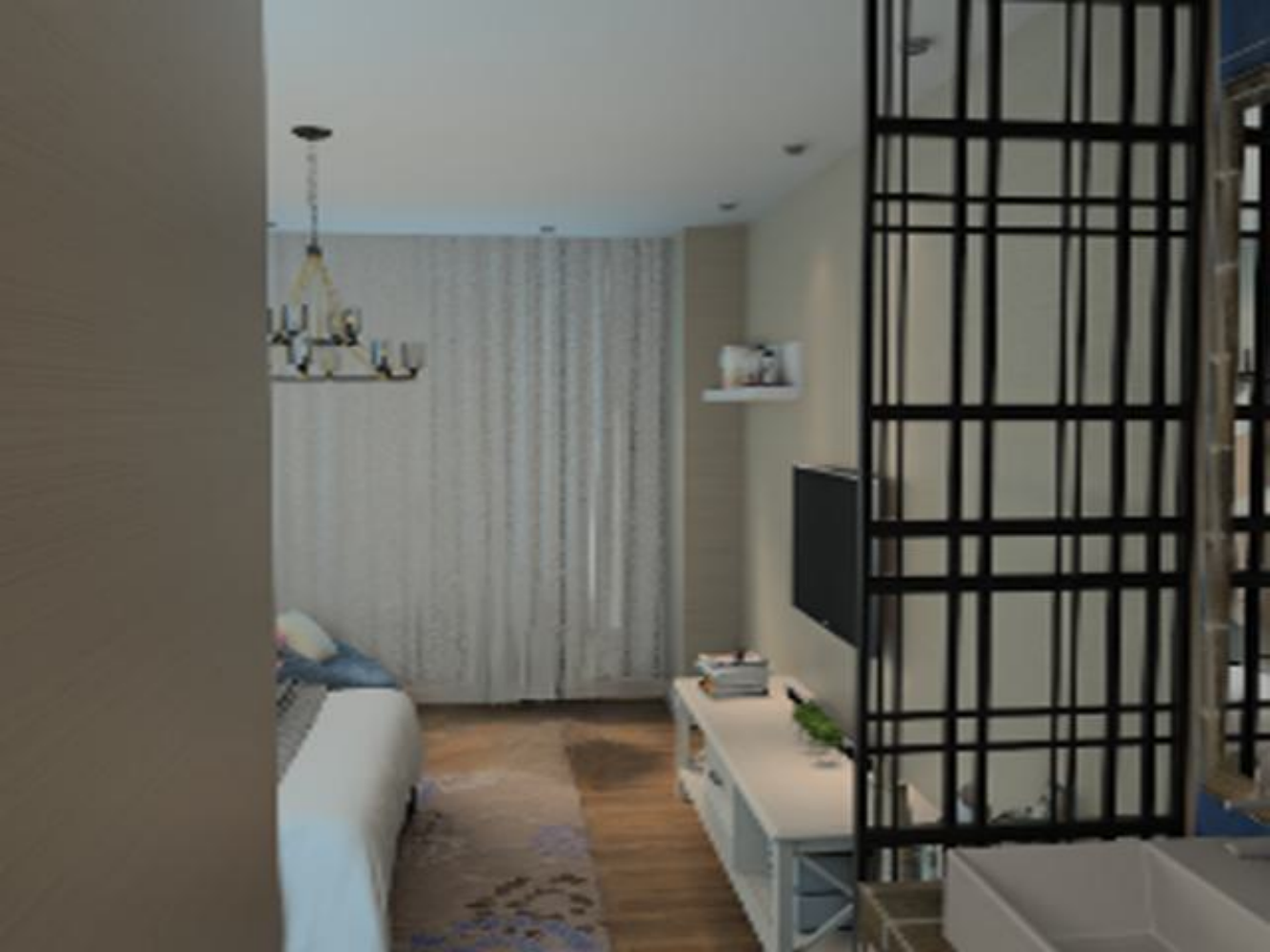}};
        \spy on (4*\myImgW-0.21,-0.50) in node at (4*\myImgW, 0);
        \node[tight] (n10) at (4*\myImgW,-\myImgH) {\includegraphics[width=.18\textwidth,valign=m]{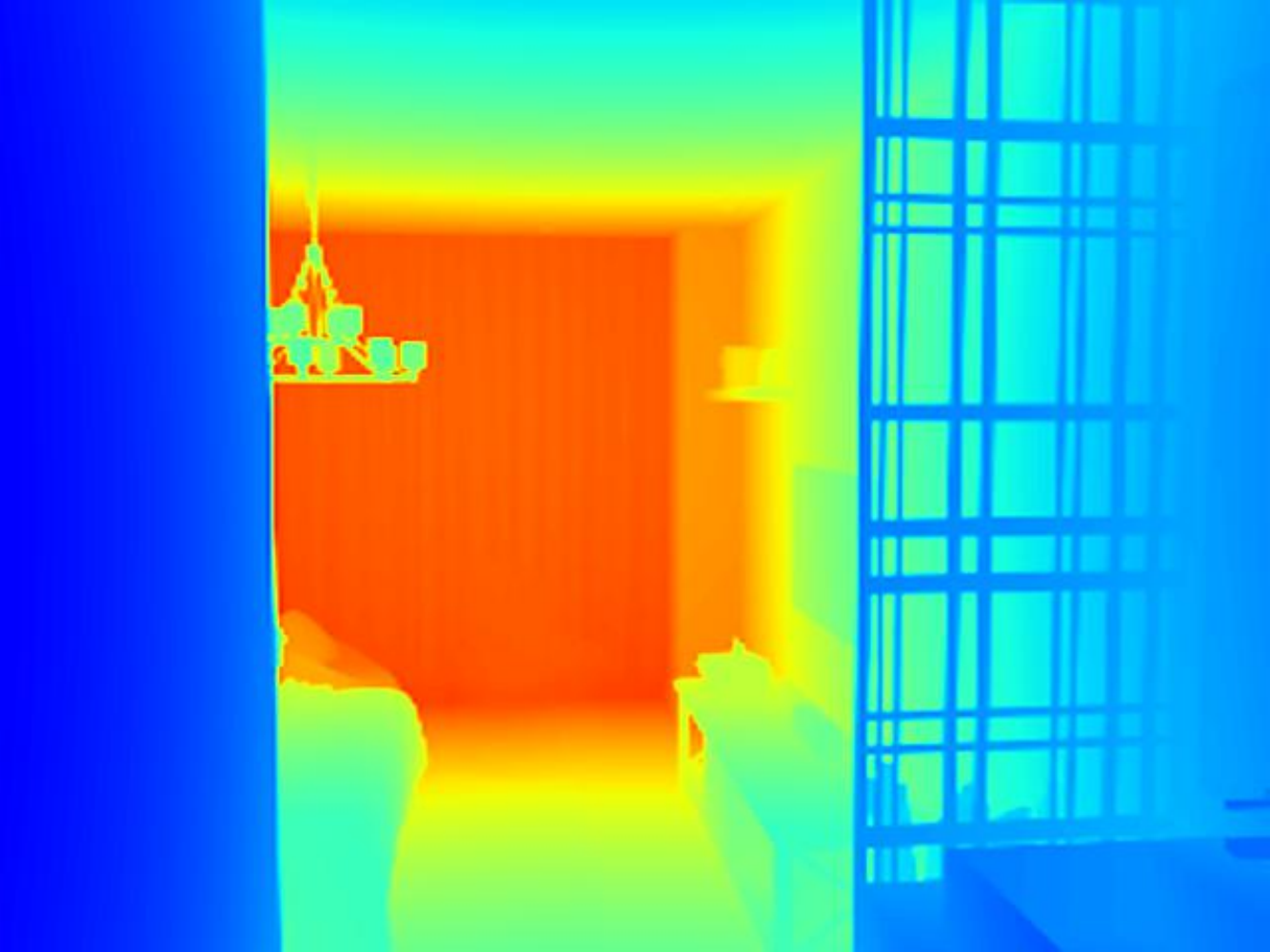}};
        \spy on (4*\myImgW-0.21, -\myImgH-0.50) in node at (4*\myImgW, -\myImgH);
        
        \draw[draw=red] (-0.21-0.4, -0.50-0.3) rectangle ++(0.8, 0.6);
        \draw[draw=black] (-0.21-0.4, -\myImgH-0.50-0.3) rectangle ++(0.8, 0.6);
        
        \draw (-2, -3*\myImgH + \myImgH / 2) node[rotate=90] {\small{ScanNet~\cite{dai2017scannet}}};
        
        \node[tight] (n1) at (0, -2*\myImgH) {\includegraphics[width=0.18\textwidth]{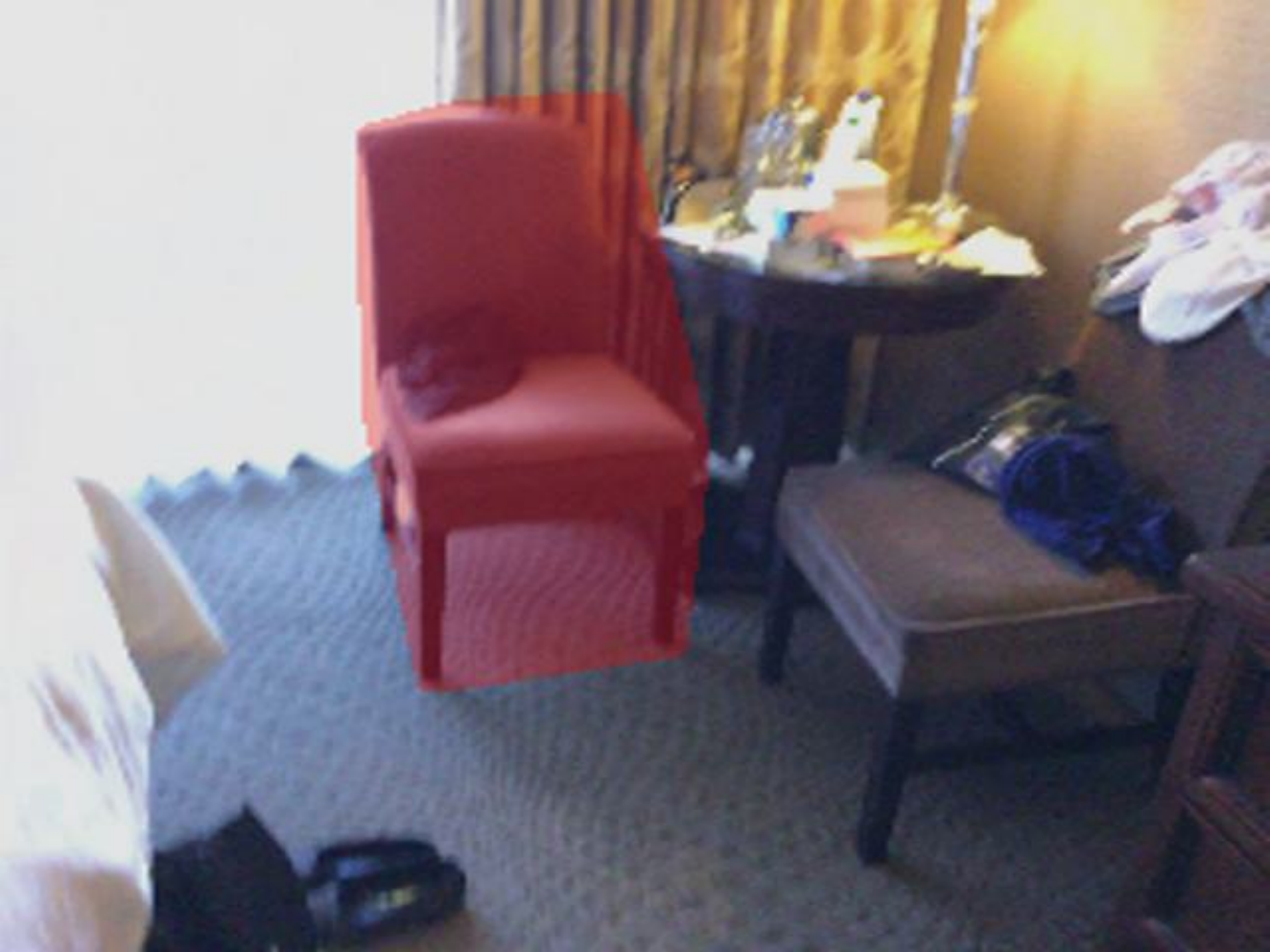}};
        \node[tight] (n2) at (0,-3*\myImgH)
        {\includegraphics[width=.18\textwidth,valign=m]{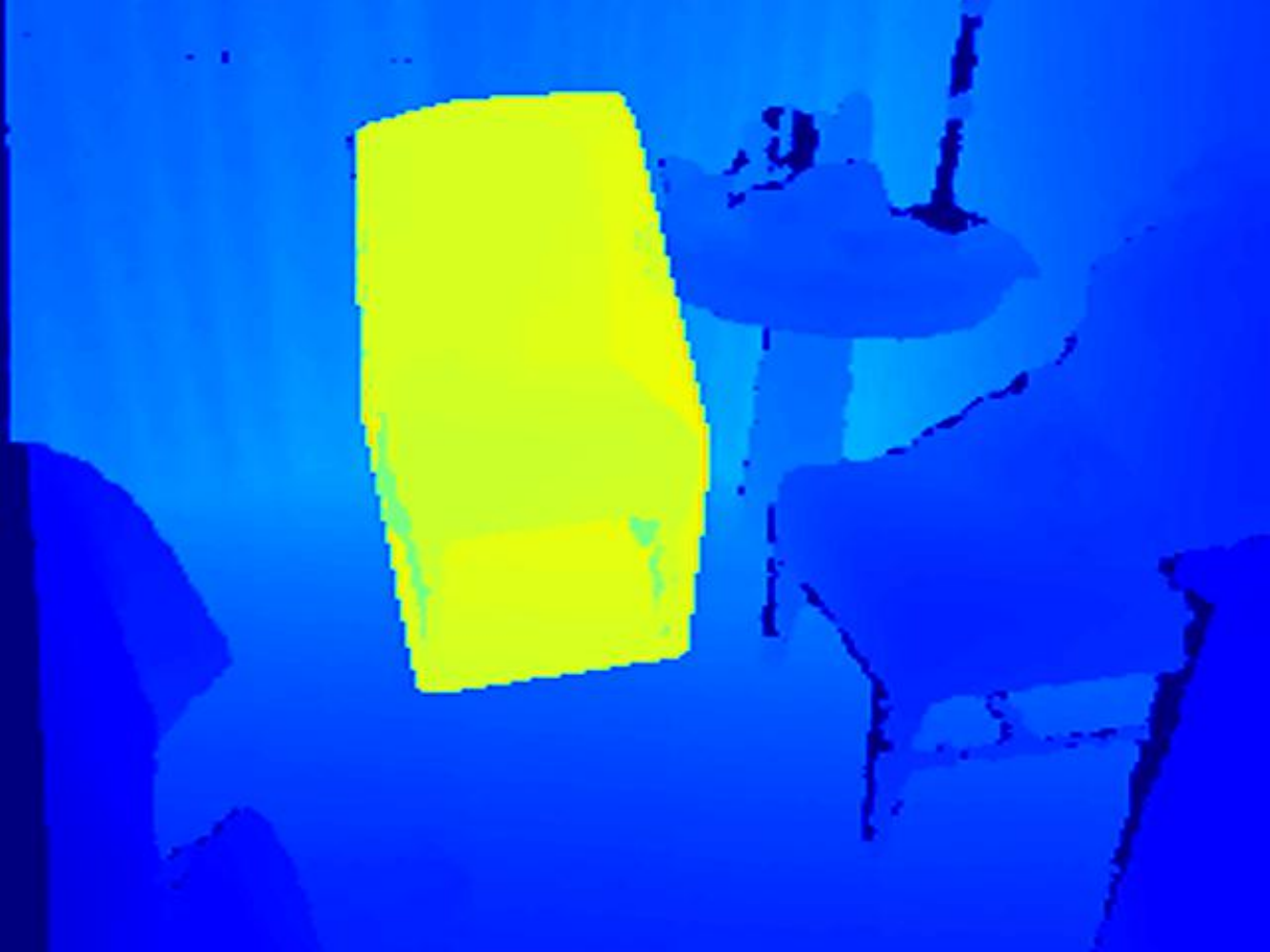}};
    
        \node[tight] (n3) at (\myImgW, -2*\myImgH) {\includegraphics[width=0.18\textwidth]{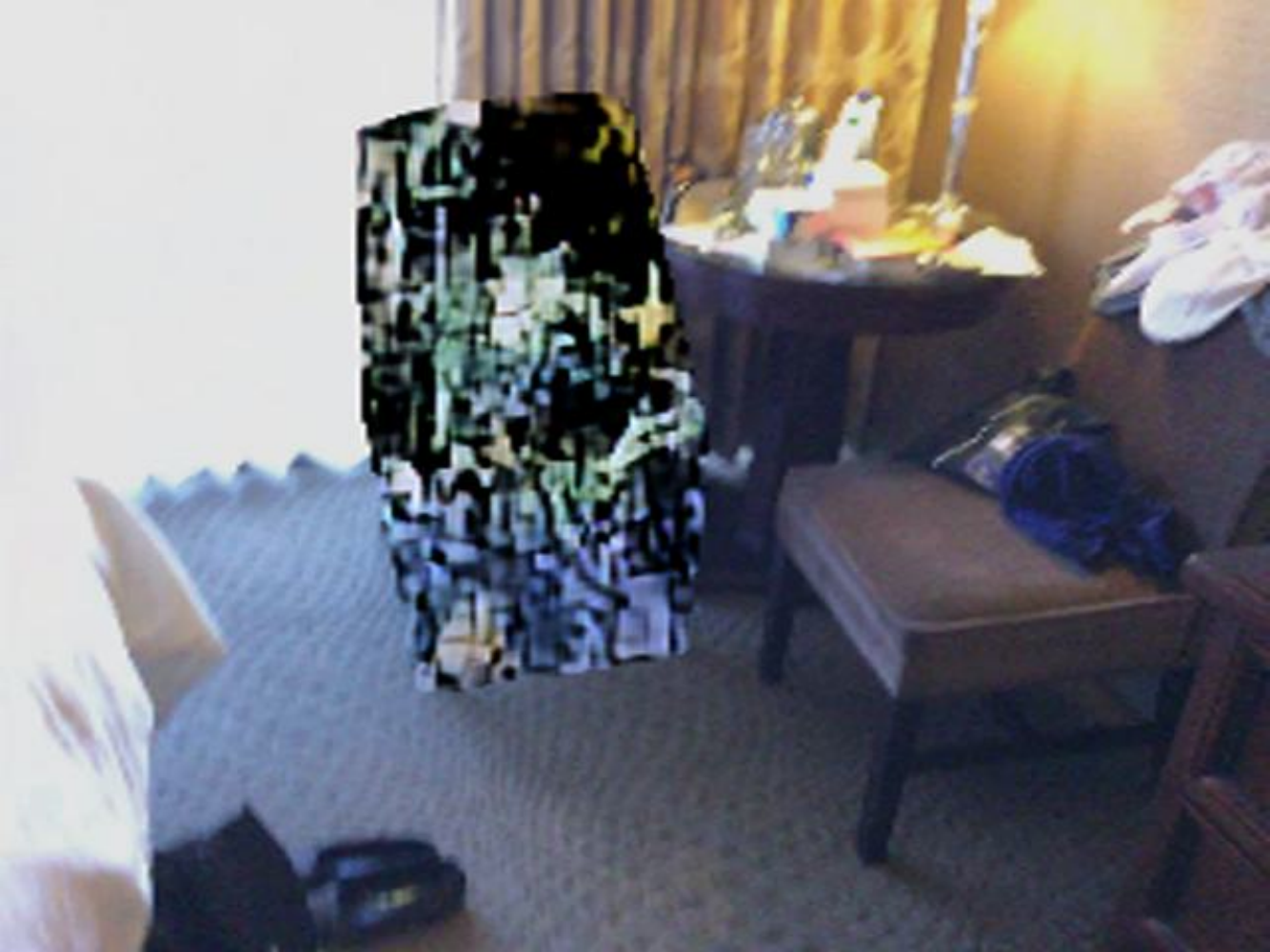}};
        \spy [magnification=1.4] on (\myImgW-0.21, -2*\myImgH+0.18) in node at (\myImgW, -2*\myImgH);
        \node[tight] (n4) at (\myImgW, -3*\myImgH) {\includegraphics[width=.18\textwidth,valign=m]{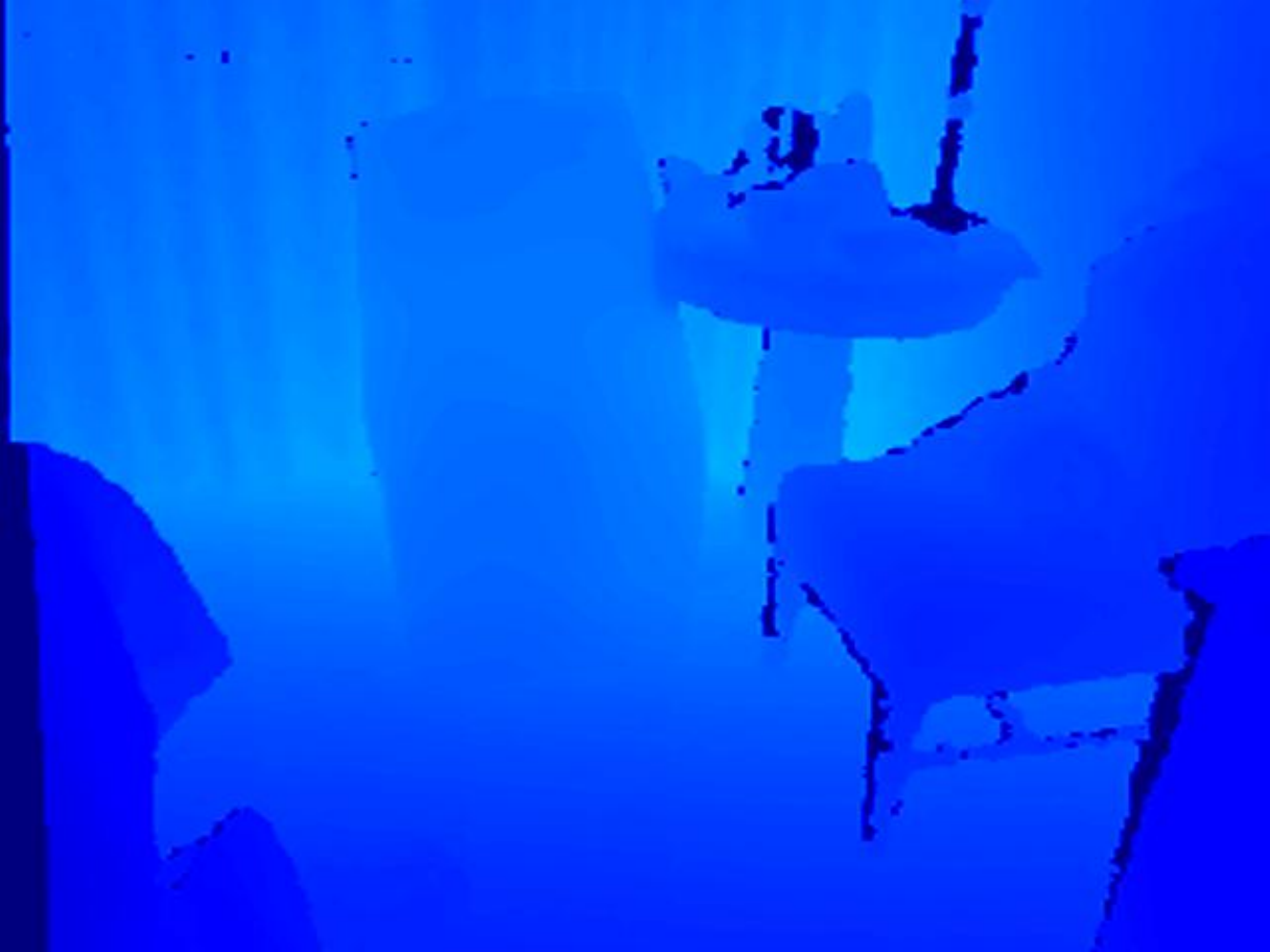}};
        \spy [magnification=1.4] on (\myImgW-0.21, -3*\myImgH+0.18) in node at (\myImgW, -3*\myImgH);
        
        \node[tight] (n5) at (2*\myImgW, -2*\myImgH) {\includegraphics[width=0.18\textwidth]{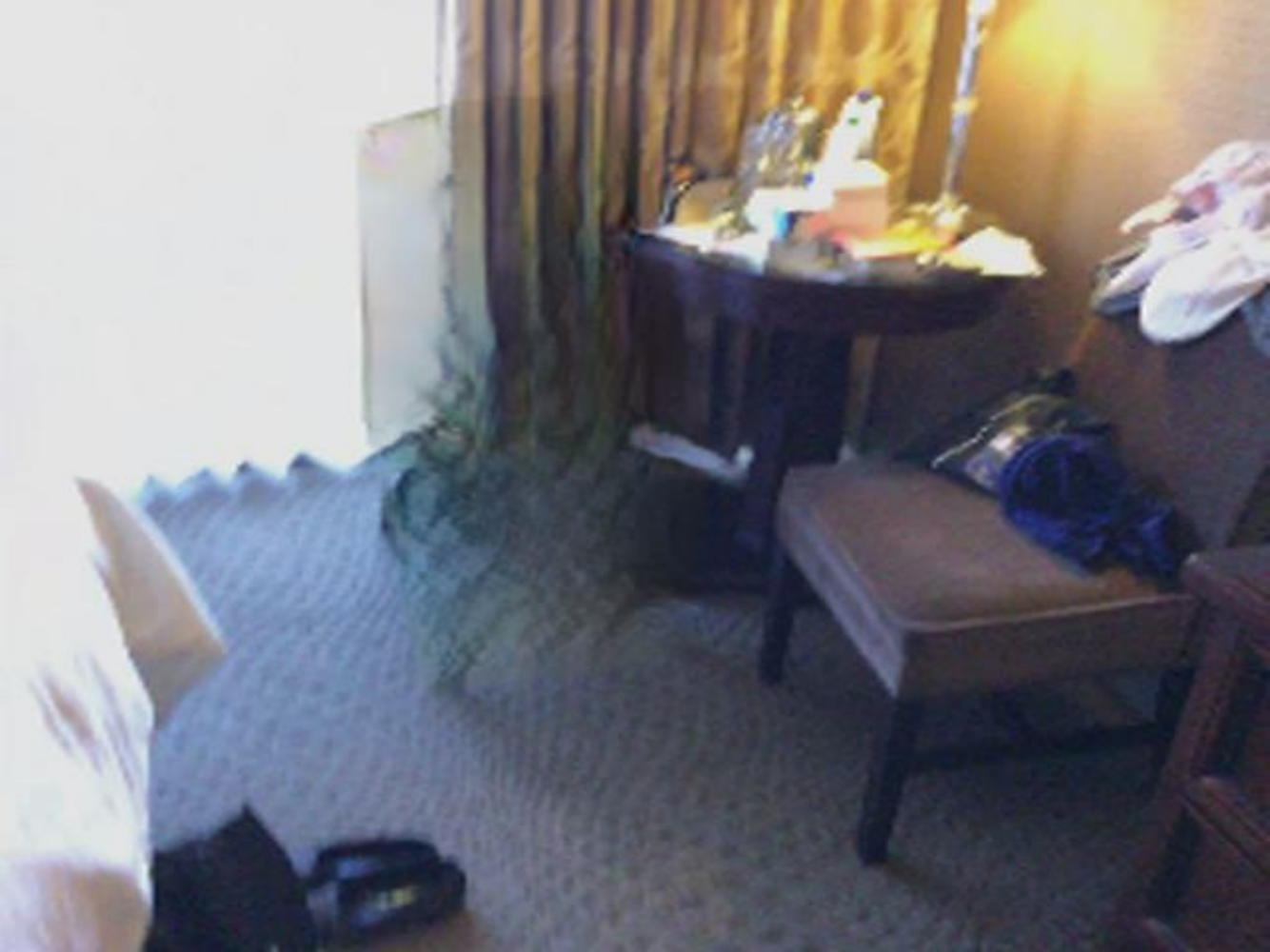}};
        \spy [magnification=1.4] on (2*\myImgW-0.21, -2*\myImgH+0.18) in node at (2*\myImgW, -2*\myImgH);
        \node[tight] (n6) at (2*\myImgW, -3*\myImgH) {\includegraphics[width=.18\textwidth,valign=m]{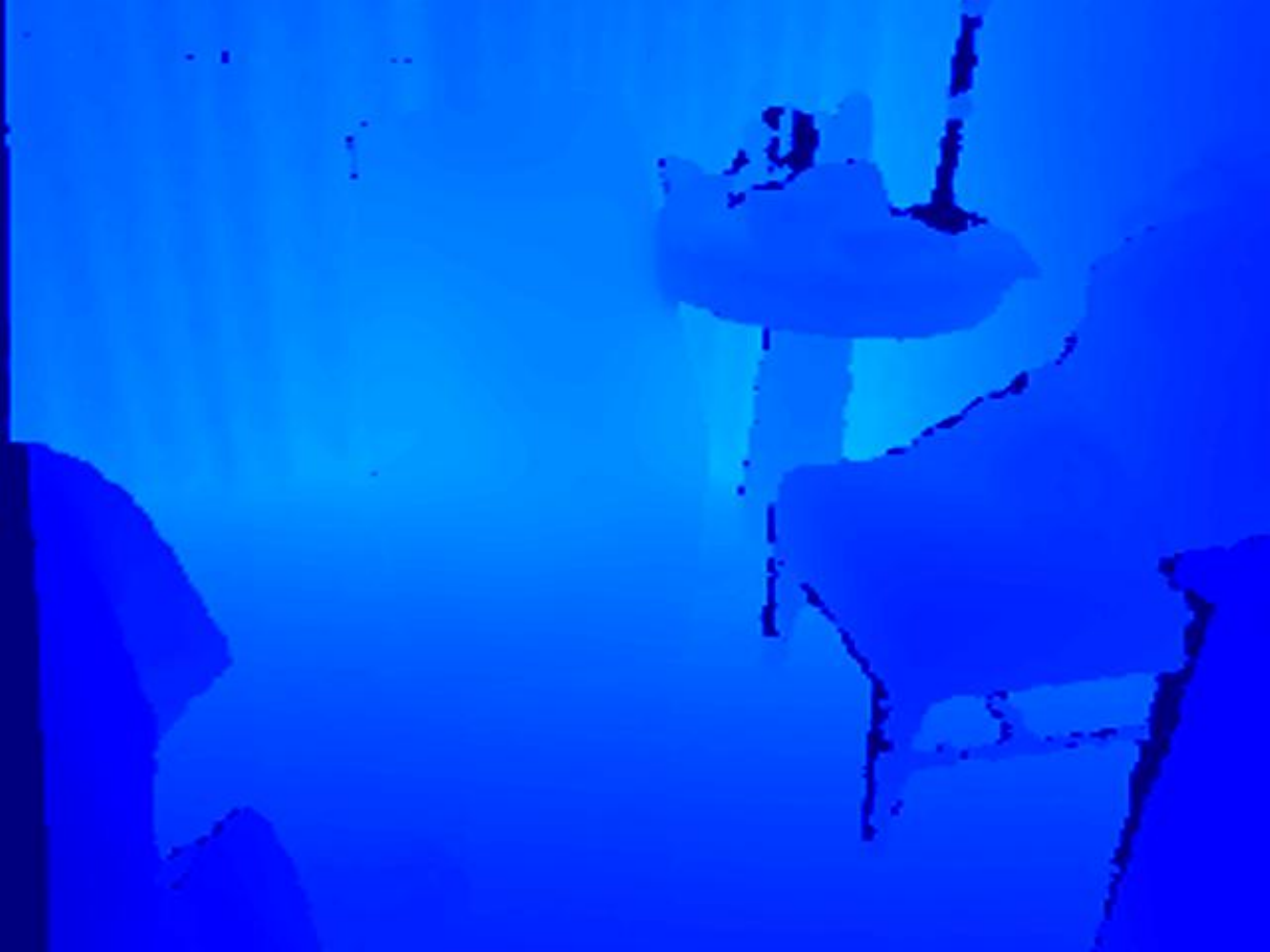}};
        \spy [magnification=1.4] on (2*\myImgW-0.21, -3*\myImgH+0.18) in node at (2*\myImgW, -3*\myImgH);
        
        \node[tight] (n7) at (3*\myImgW, -2*\myImgH) {\includegraphics[width=0.18\textwidth]{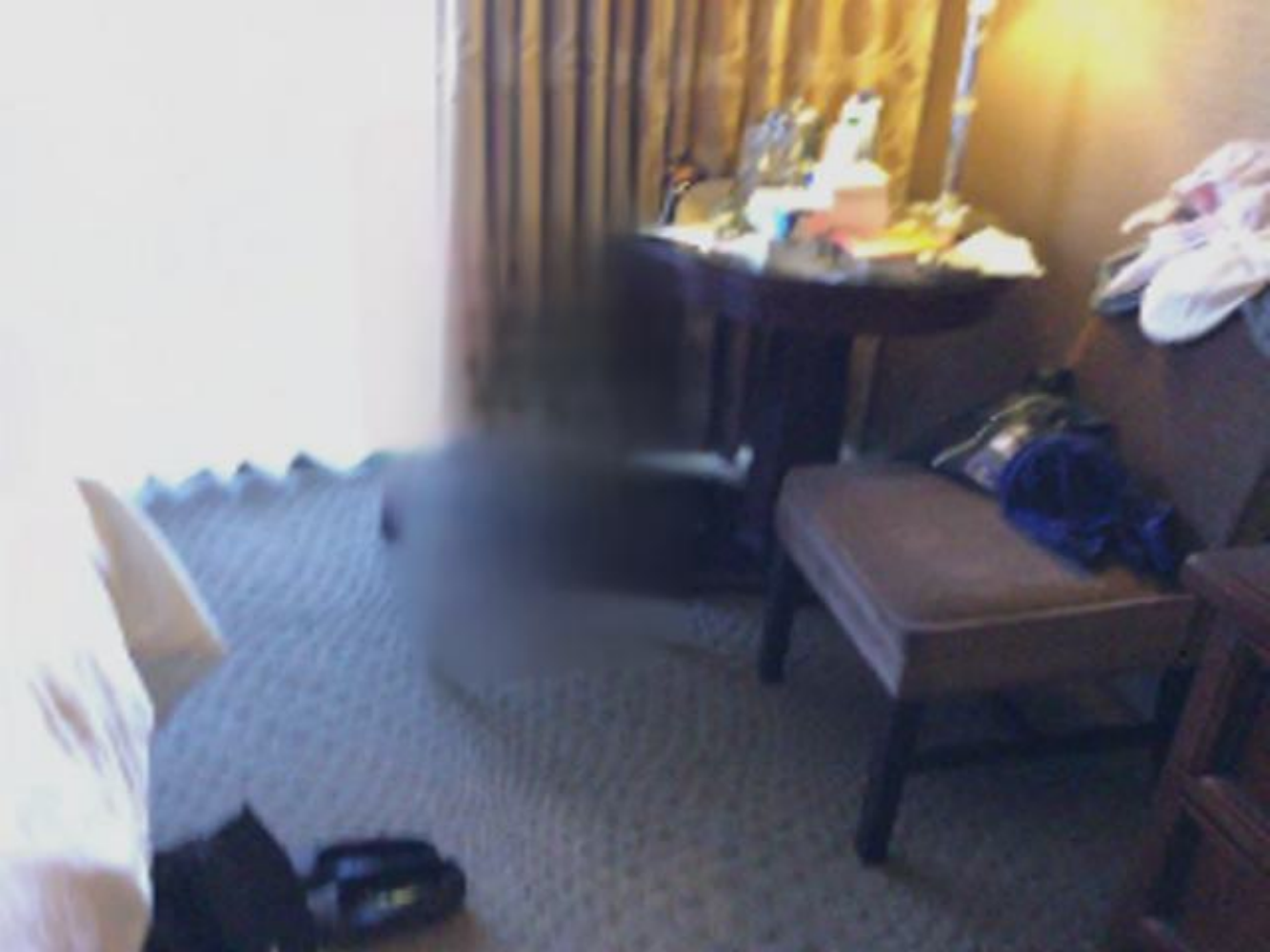}};
        \spy [magnification=1.4] on (3*\myImgW-0.21, -2*\myImgH+0.18) in node at (3*\myImgW, -2*\myImgH);
        \node[tight] (n8) at (3*\myImgW, -3*\myImgH) {\includegraphics[width=.18\textwidth,valign=m]{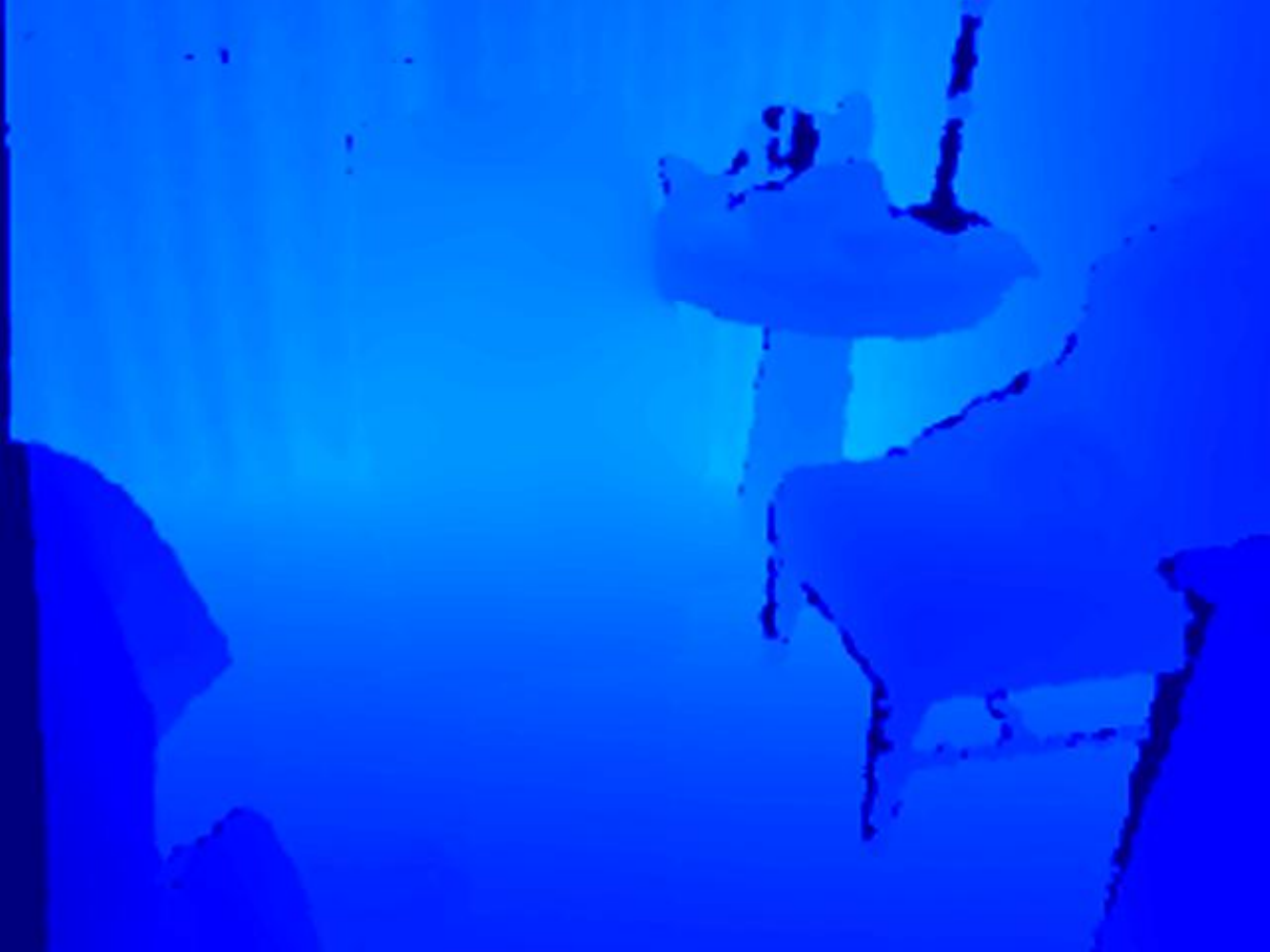}};
        \spy [magnification=1.4] on (3*\myImgW-0.21, -3*\myImgH+0.18) in node at (3*\myImgW, -3*\myImgH);
        
        \node[tight] (n9) at (4*\myImgW, -2*\myImgH) {\includegraphics[width=0.18\textwidth]{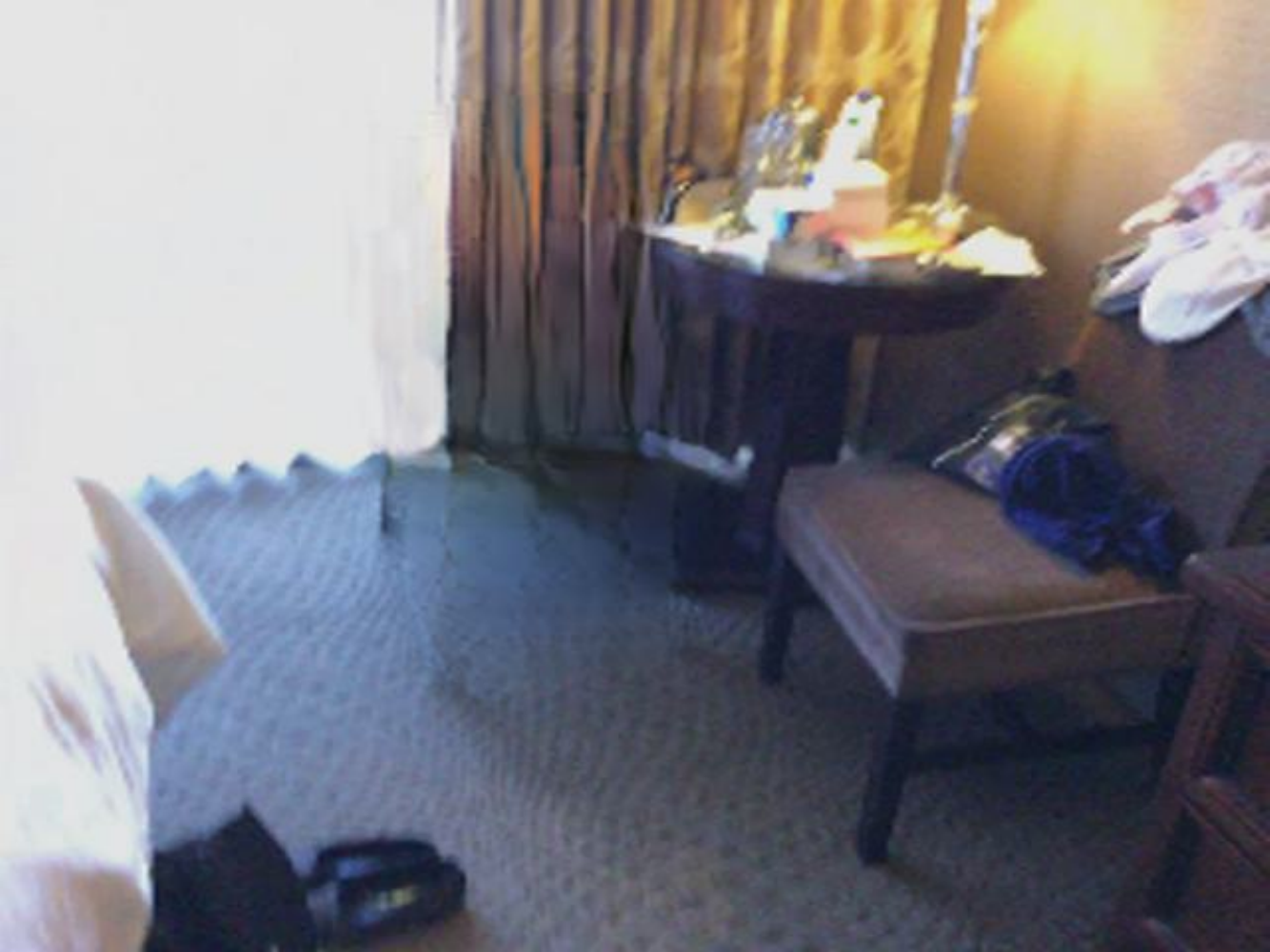}};
        \spy [magnification=1.4] on (4*\myImgW-0.21, -2*\myImgH+0.18) in node at (4*\myImgW, -2*\myImgH);
        \node[tight] (n10) at (4*\myImgW,-3*\myImgH) {\includegraphics[width=.18\textwidth,valign=m]{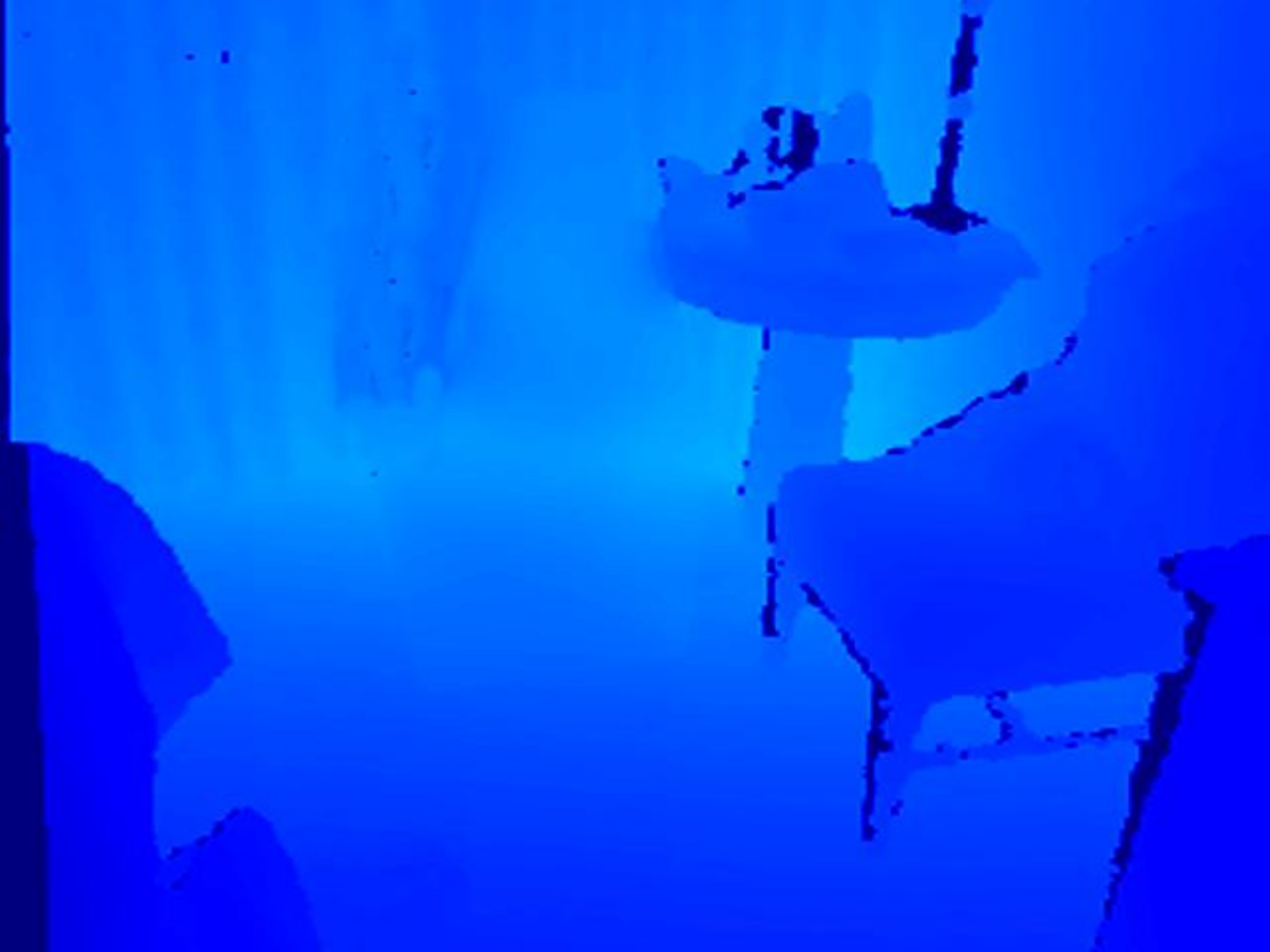}};
        \spy [magnification=1.4] on (4*\myImgW-0.21, -3*\myImgH+0.18) in node at (4*\myImgW, -3*\myImgH);
        
        \draw[draw=red] (-0.21-1.1, -2*\myImgH+0.18-0.8) rectangle ++(2.13, 1.6);
        \draw[draw=black] (-0.21-1.1, -3*\myImgH+0.18-0.8) rectangle ++(2.13, 1.6);
    \end{tikzpicture}
    \end{scaletikzpicturetowidth}
    \caption{Qualitative comparison for diminishing objects from InteriorNet~\cite{li2018interiornet} (synthetic) and ScanNet~\cite{dai2017scannet} (real) with models trained on InteriorNet. Result images are zoomed to the red and black outlines in the first column. }
    \label{fig:comparison_interior_scan}
\end{figure*}%

\def\offx1{0.0}
\def\offy1{0.1}
\def\mag1{3.5}
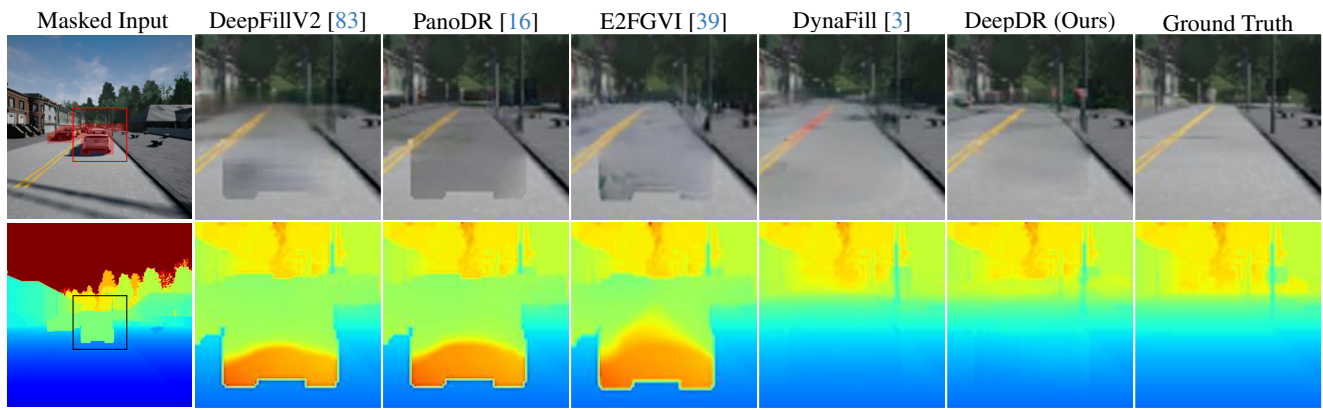
\begin{figure*}
    \centering
    \noindent
    \begin{tikzpicture}[spy using outlines={rectangle, width=\myImgWd cm, height=\myImgHd cm, white}, remember picture]
        \draw (0*\myImgWd, \myImgHd / 1.8) node {\small{ Masked Input}};
        \draw (1*\myImgWd, \myImgHd / 1.8) node {\small{DeepFillV2~\cite{yu2019free}}};
        \draw (2*\myImgWd, \myImgHd / 1.8) node {\small{PanoDR~\cite{gkitsas2021panodr}}};
        \draw (3*\myImgWd, \myImgHd / 1.8) node {\small{E2FGVI~\cite{li2022towards}}};
        \draw (4*\myImgWd, \myImgHd / 1.8) node {\small{DynaFill~\cite{bevsic2020dynamic}}};
        \draw (5*\myImgWd, \myImgHd / 1.8) node {\small{DeepDR (Ours)}};
        \draw (6*\myImgWd, \myImgHd / 1.8) node {\small{Ground Truth}};
        
        \node[tight] (n1) at (0, -0*\myImgHd) {\includegraphics[width=0.14\linewidth]{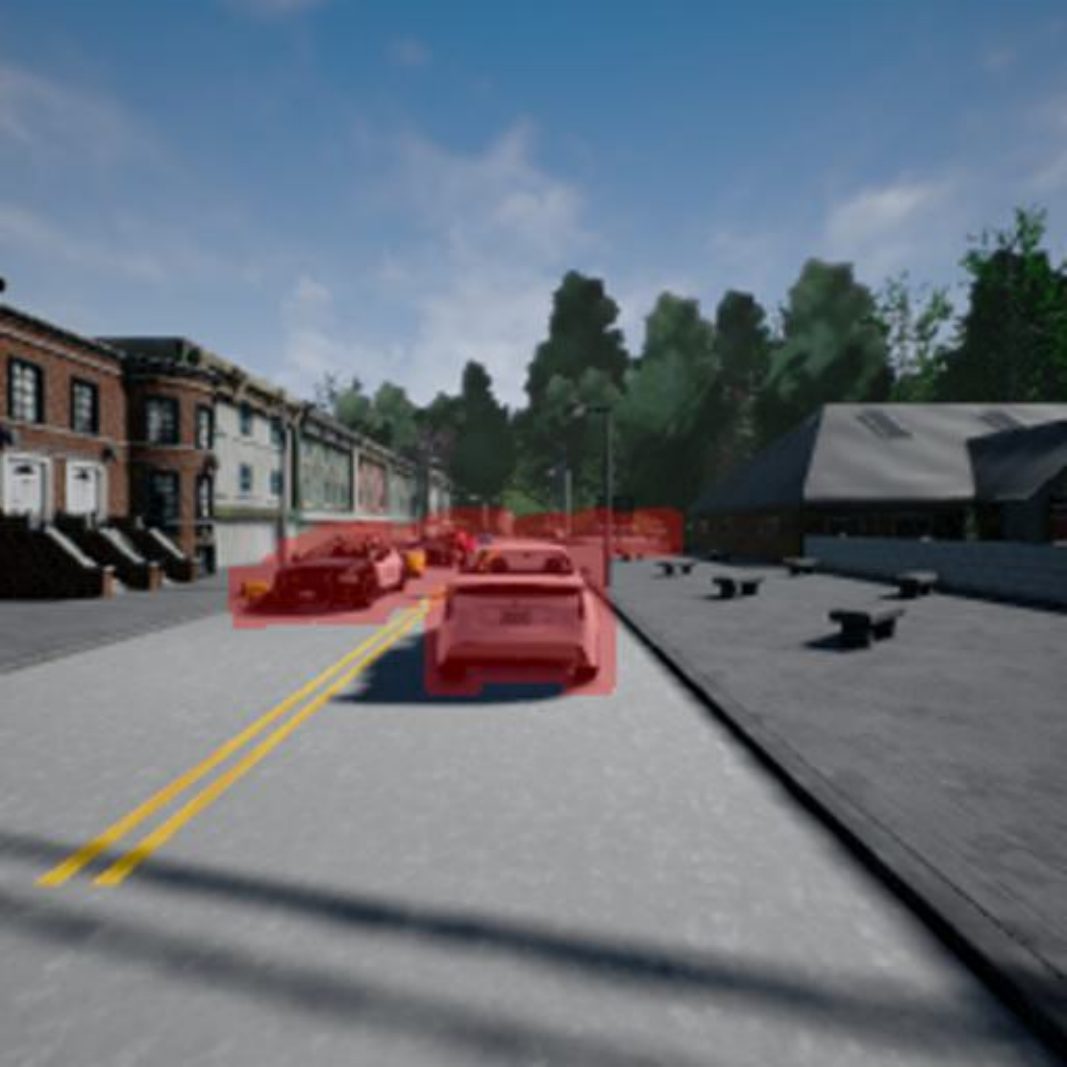}};
        \node[tight] (n4) at (0, -1*\myImgHd)
        {\includegraphics[width=0.14\linewidth,valign=m]{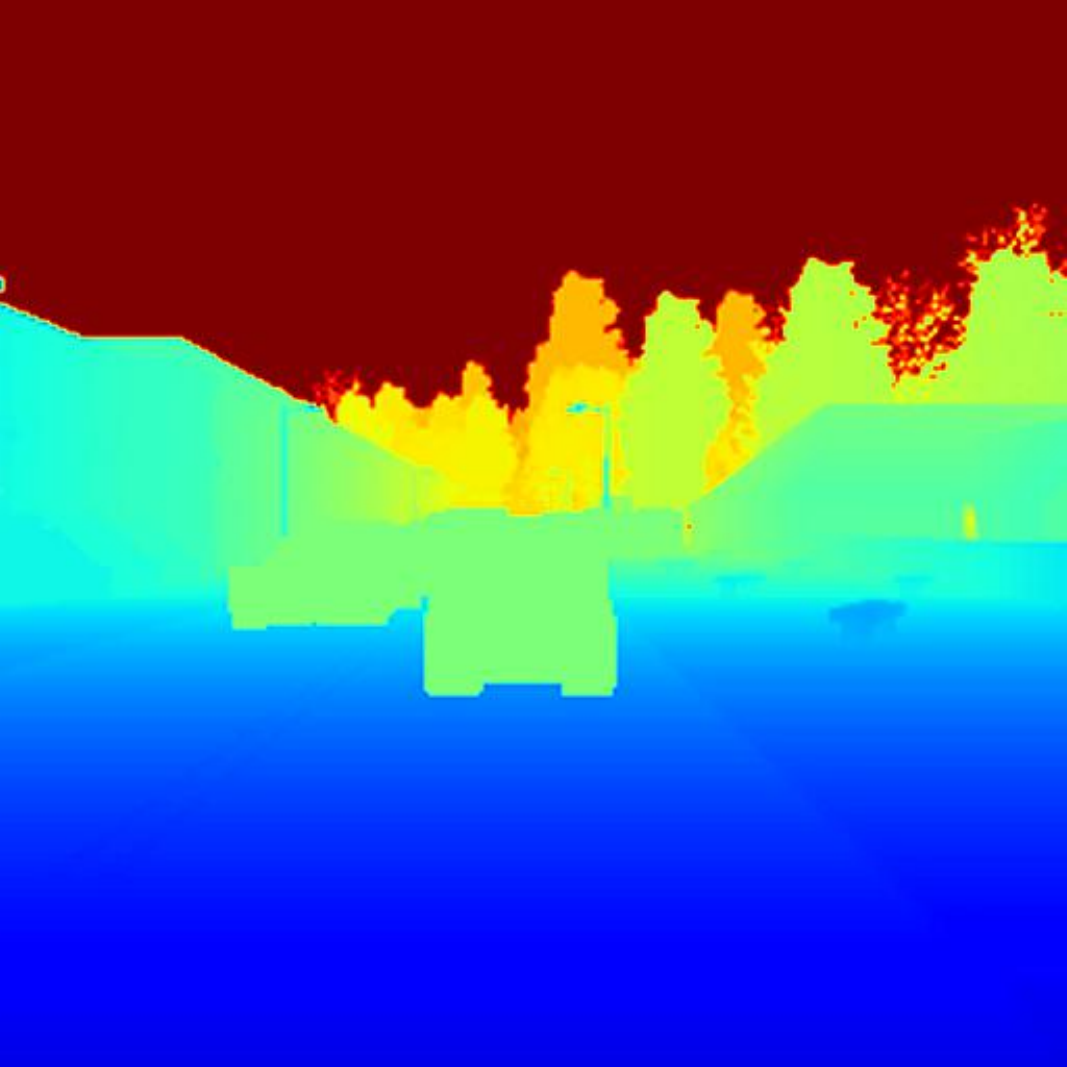}};
        
        \node[tight] (n3) at (\myImgWd, -0*\myImgHd) {
        \includegraphics[width=0.14\linewidth]{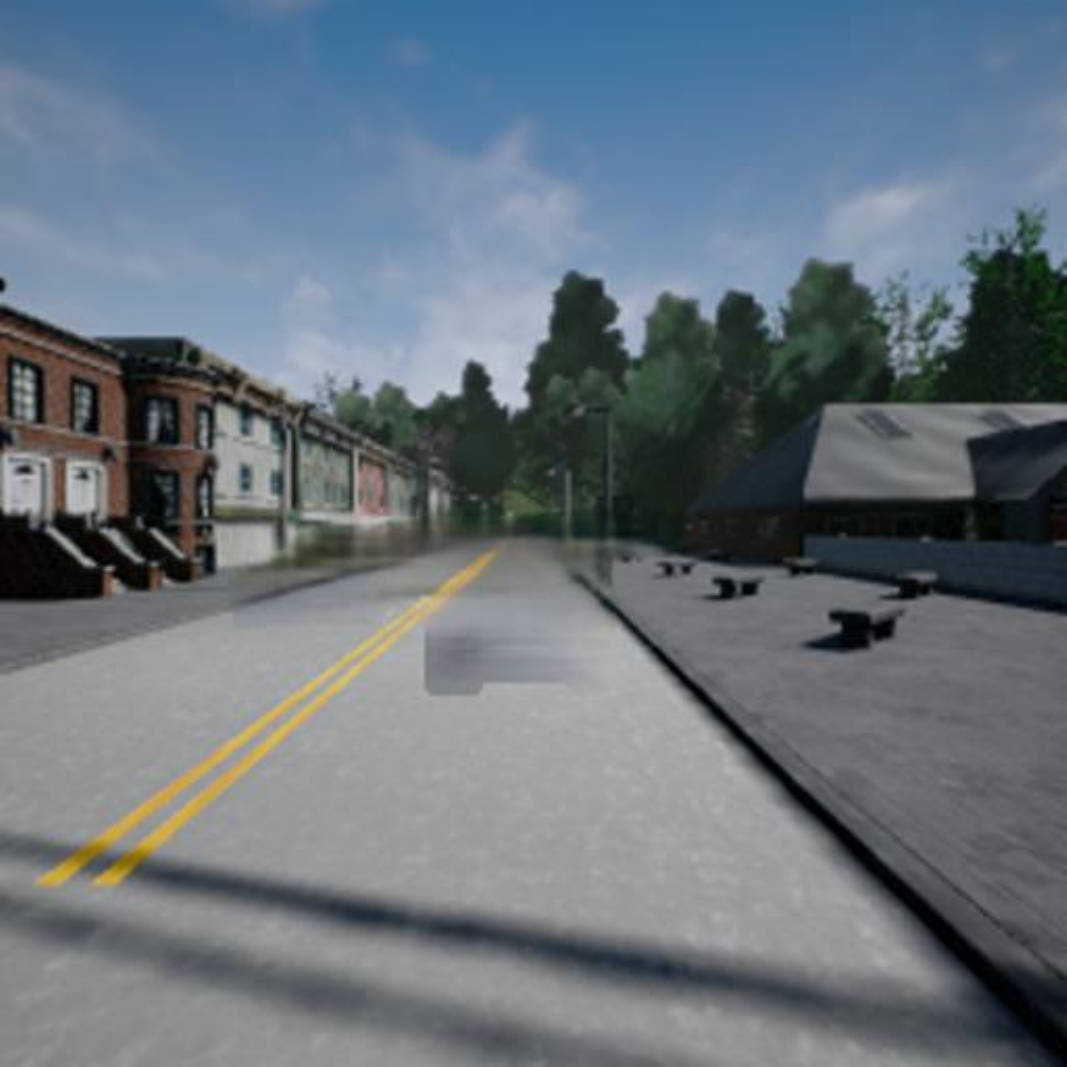}
        };
        \spy[magnification=\mag1] on (\myImgWd-\offx1, -0*\myImgHd-\offy1) in node at (1*\myImgWd, -0*\myImgHd);
        \node[tight] (n4) at (\myImgWd, -1*\myImgHd){
        \includegraphics[width=0.14\linewidth,valign=m]{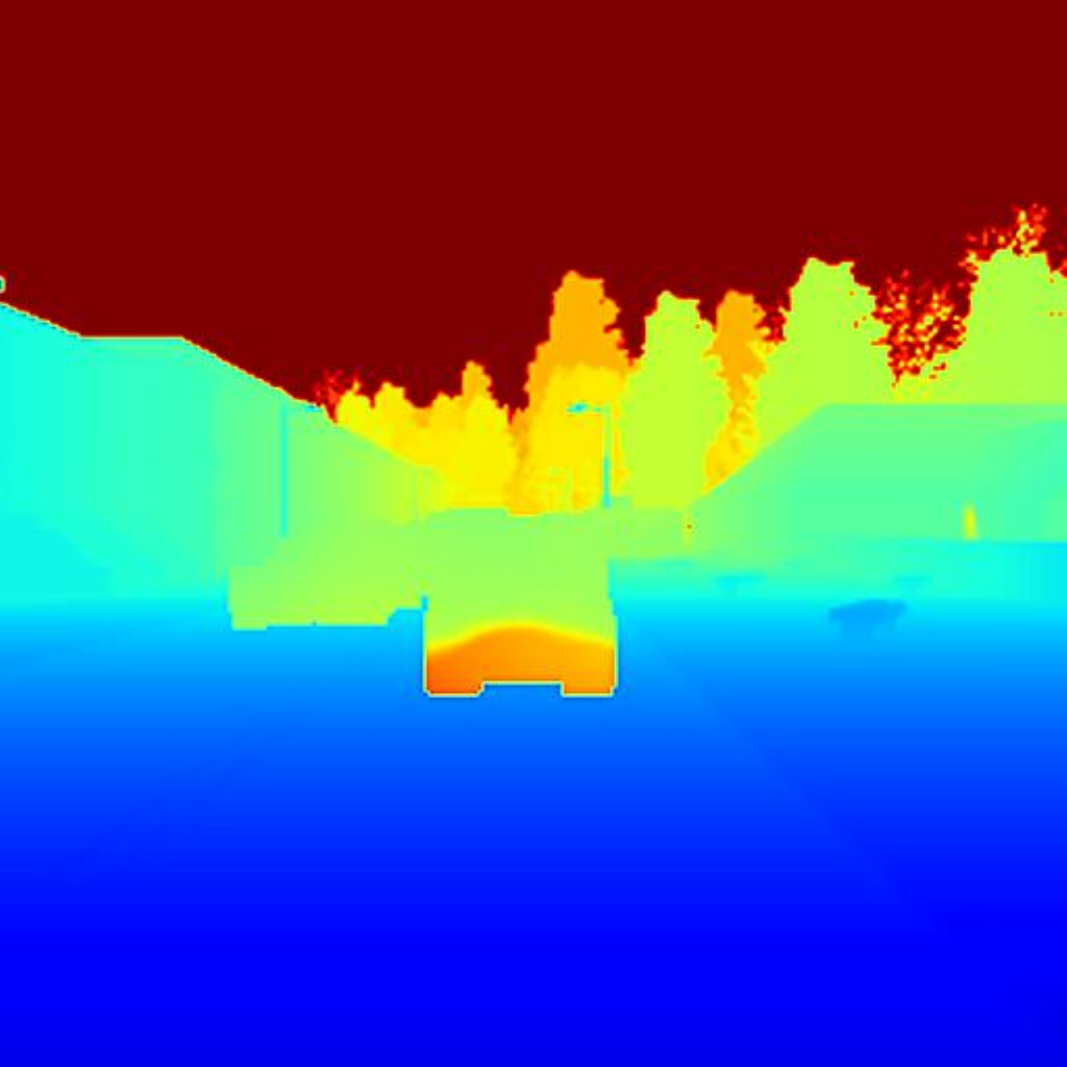}
        };
        \spy[magnification=\mag1] on (\myImgWd-\offx1, -1*\myImgHd-\offy1) in node at (1*\myImgWd, -1*\myImgHd);
        
        \node[tight] (n5) at (2*\myImgWd, -0*\myImgHd) {
        \includegraphics[width=0.14\linewidth]{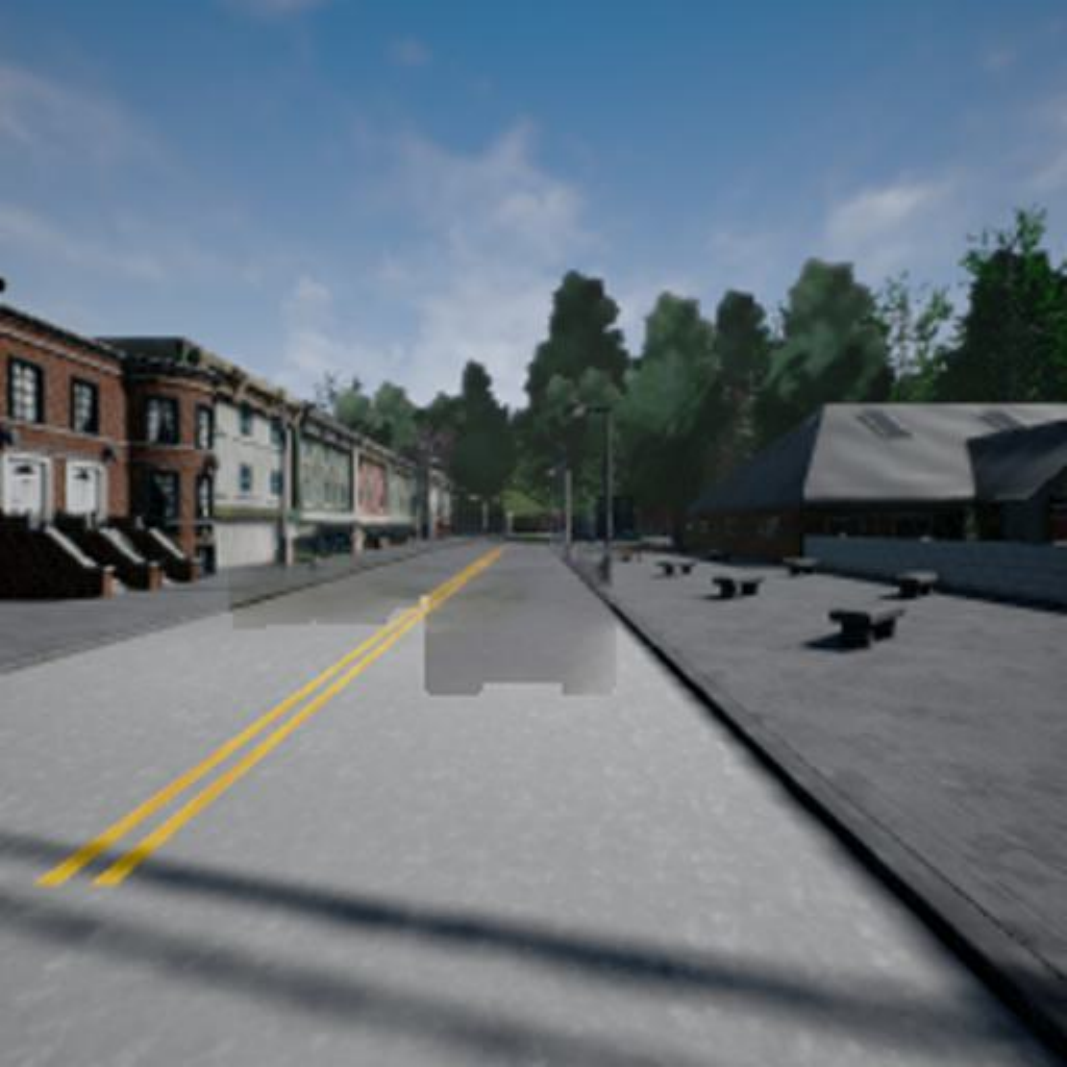}
        };
        \spy[magnification=\mag1] on (2*\myImgWd-\offx1, -0*\myImgHd-\offy1) in node at (2*\myImgWd, -0*\myImgHd);
        \node[tight] (n6) at (2*\myImgWd, -1*\myImgHd) {
        \includegraphics[width=0.14\linewidth,valign=m]{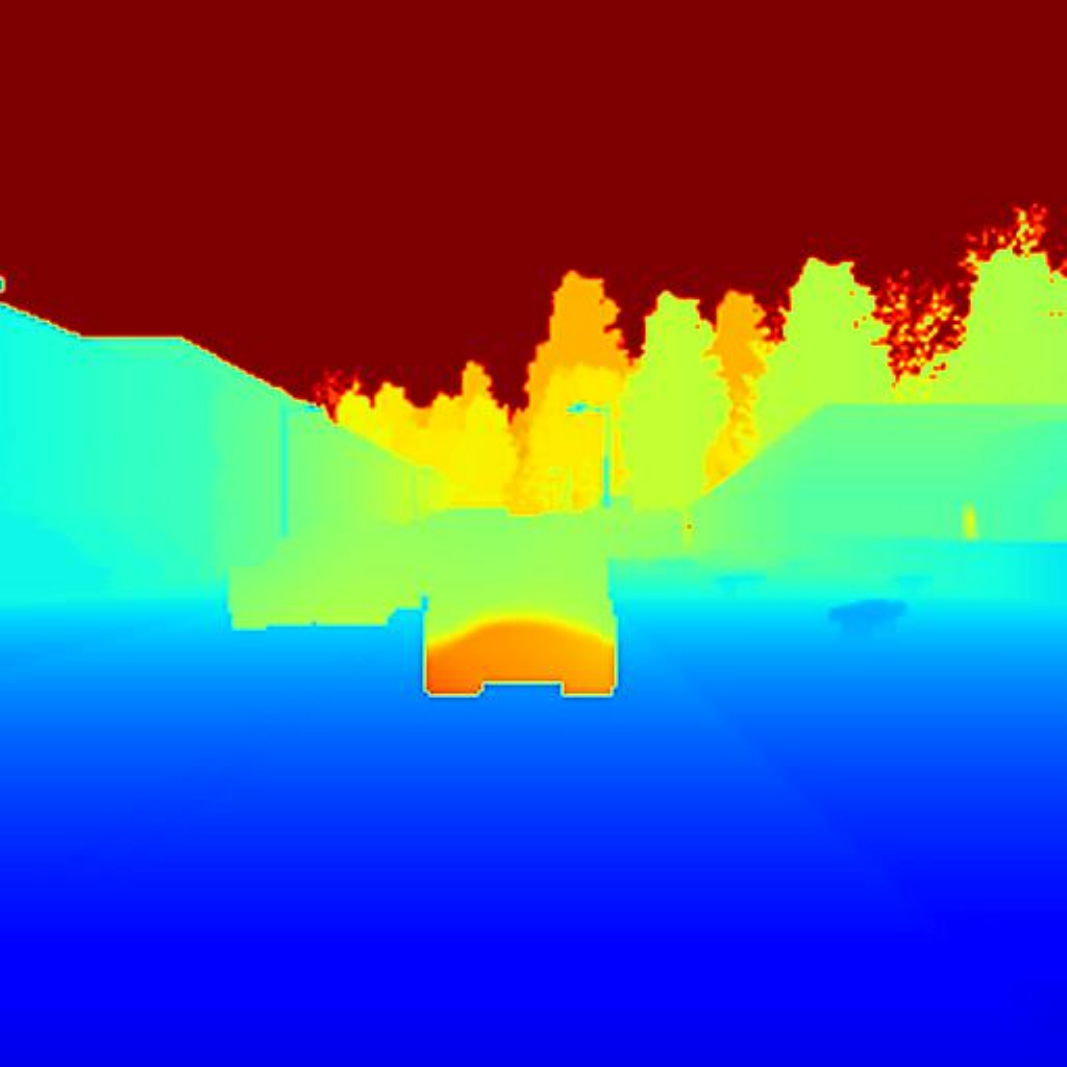}
        };
        \spy[magnification=\mag1] on (2*\myImgWd-\offx1, -1*\myImgHd-\offy1) in node at (2*\myImgWd, -1*\myImgHd);

        \node[tight] (n7) at (3*\myImgWd, -0*\myImgHd) {
        \includegraphics[width=0.14\linewidth]{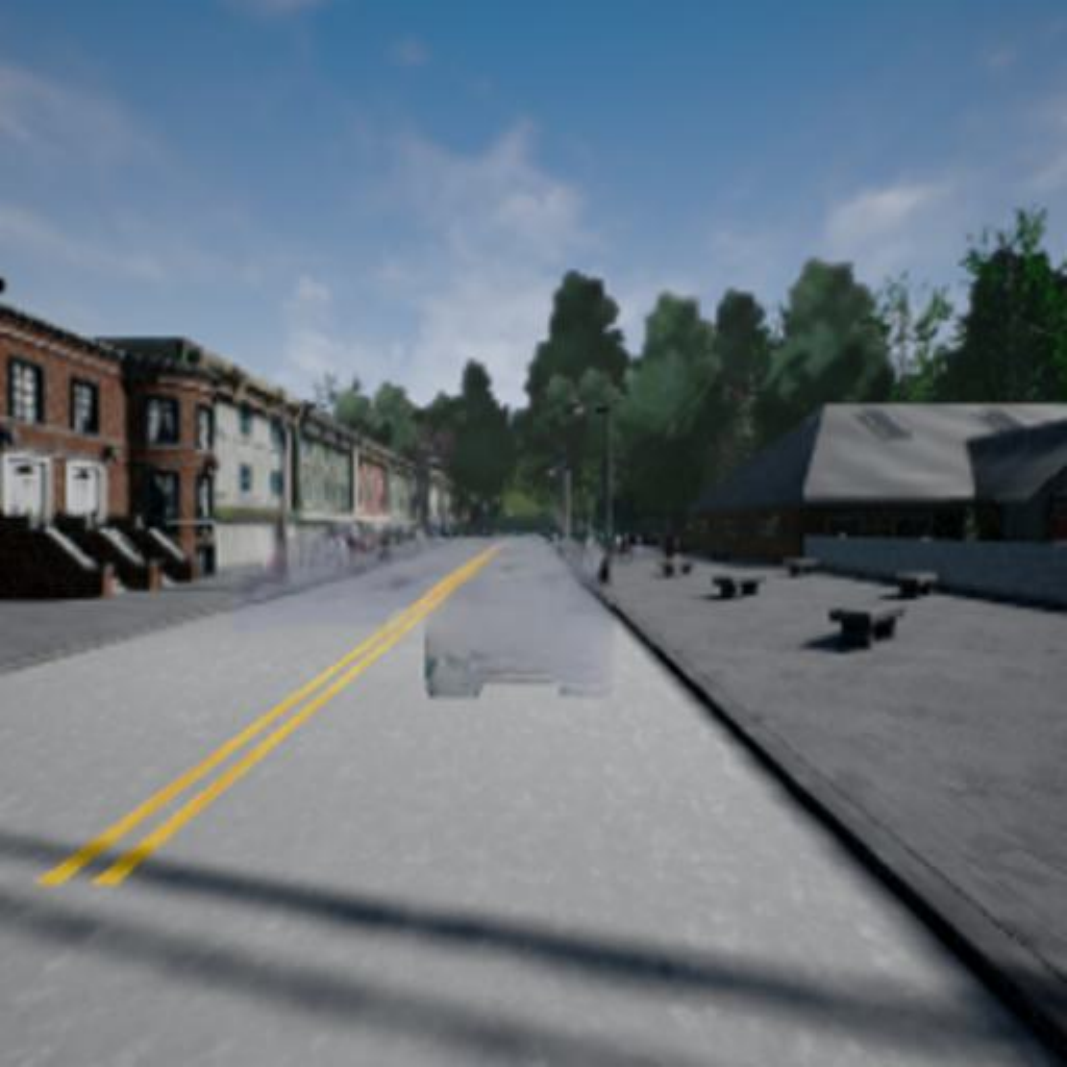}
        };
        \spy[magnification=\mag1] on (3*\myImgWd-\offx1, -0*\myImgHd-\offy1) in node at (3*\myImgWd, -0*\myImgHd);
        \node[tight] (n8) at (3*\myImgWd, -1*\myImgHd) {
        \includegraphics[width=0.14\linewidth,valign=m]{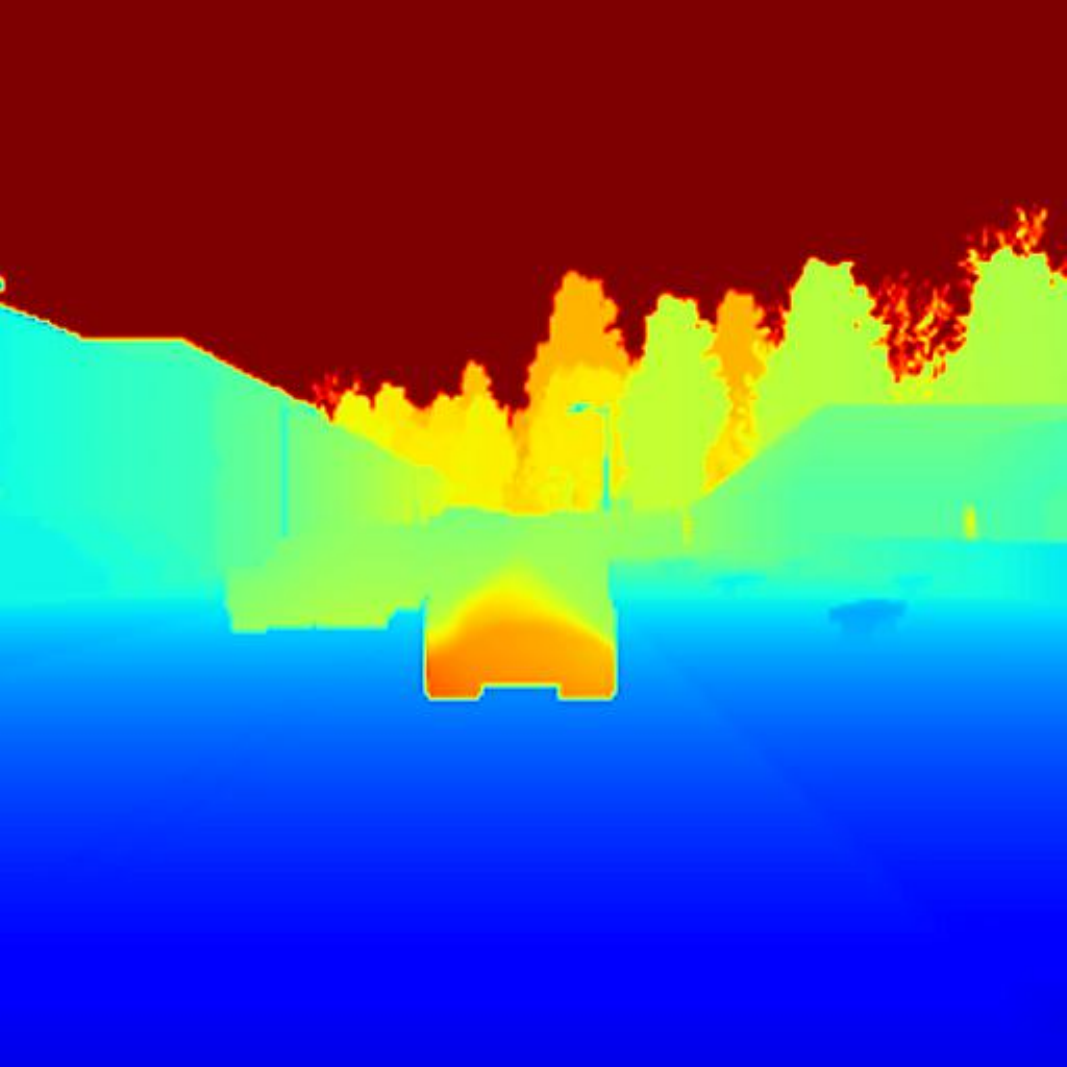}
        };
        \spy[magnification=\mag1] on (3*\myImgWd-\offx1, -1*\myImgHd-\offy1) in node at (3*\myImgWd, -1*\myImgHd);

        \node[tight] (n7) at (4*\myImgWd, -0*\myImgHd) {
        \includegraphics[width=0.14\linewidth]{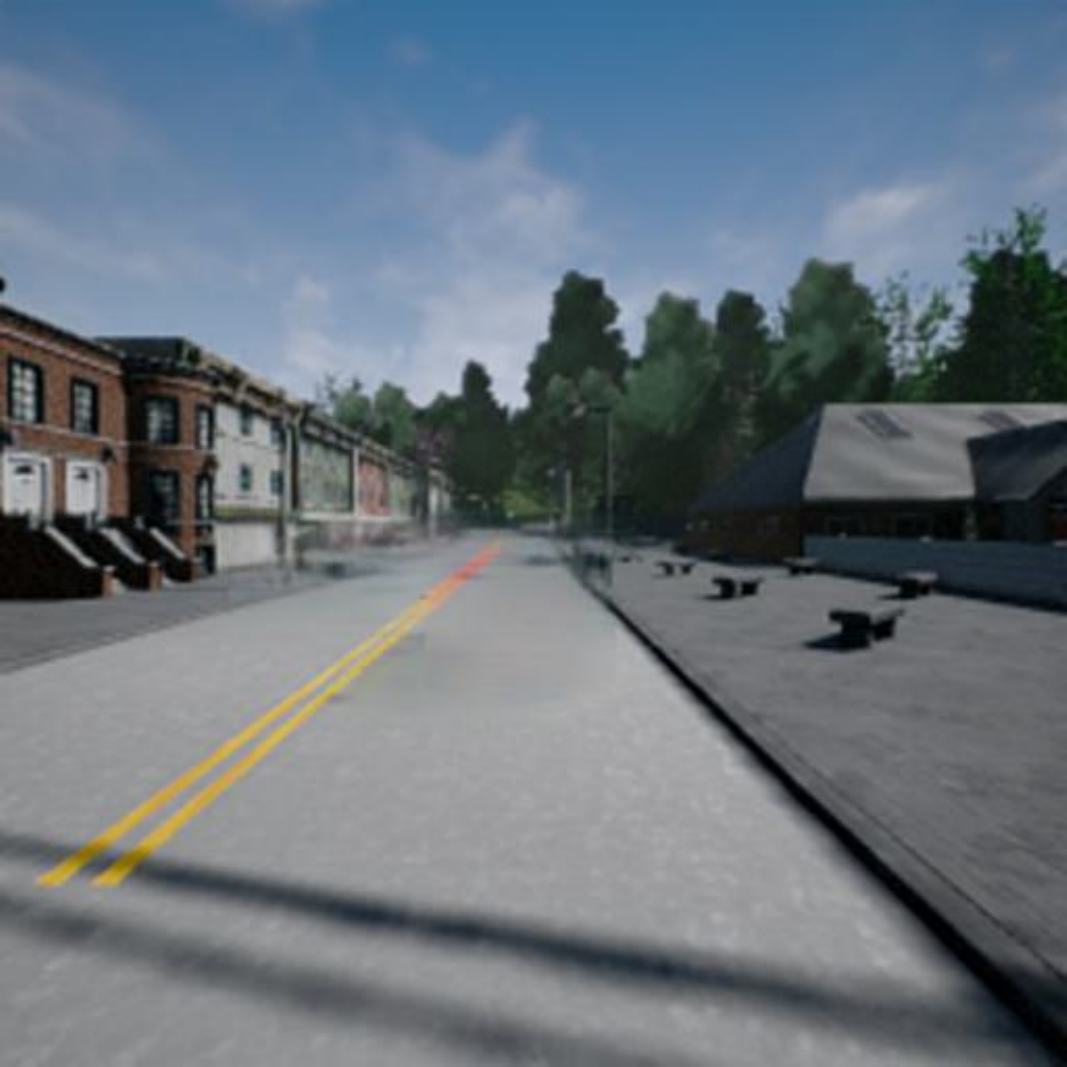}
        };
        \spy[magnification=\mag1] on (4*\myImgWd-\offx1, -0*\myImgHd-\offy1) in node at (4*\myImgWd, -0*\myImgHd);
        \node[tight] (n8) at (4*\myImgWd, -1*\myImgHd) {
        \includegraphics[width=0.14\linewidth,valign=m]{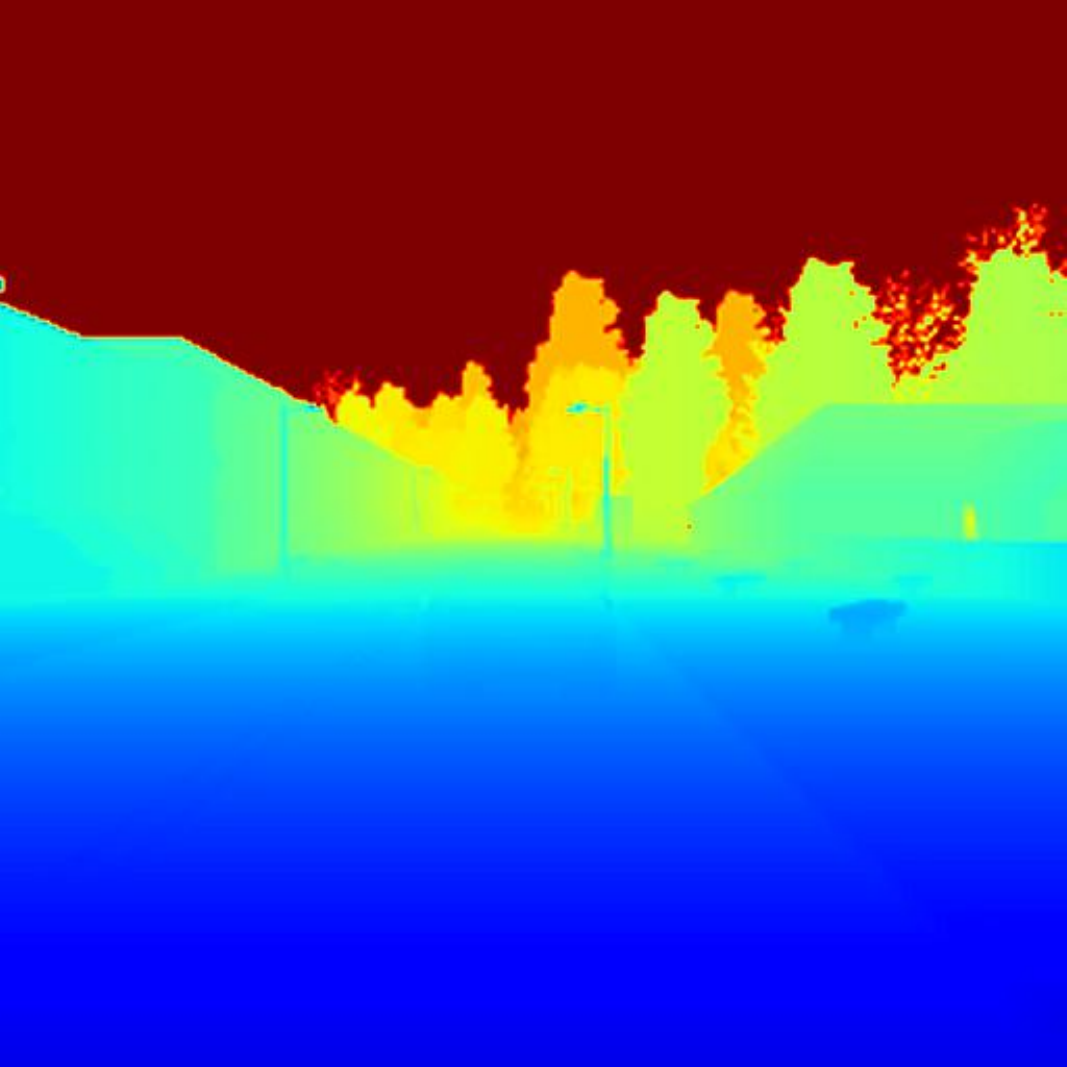}
        };
        \spy[magnification=\mag1] on (4*\myImgWd-\offx1, -1*\myImgHd-\offy1) in node at (4*\myImgWd, -1*\myImgHd);
        
        \node[tight] (n9) at (5*\myImgWd, -0*\myImgHd) {\includegraphics[width=0.14\linewidth]{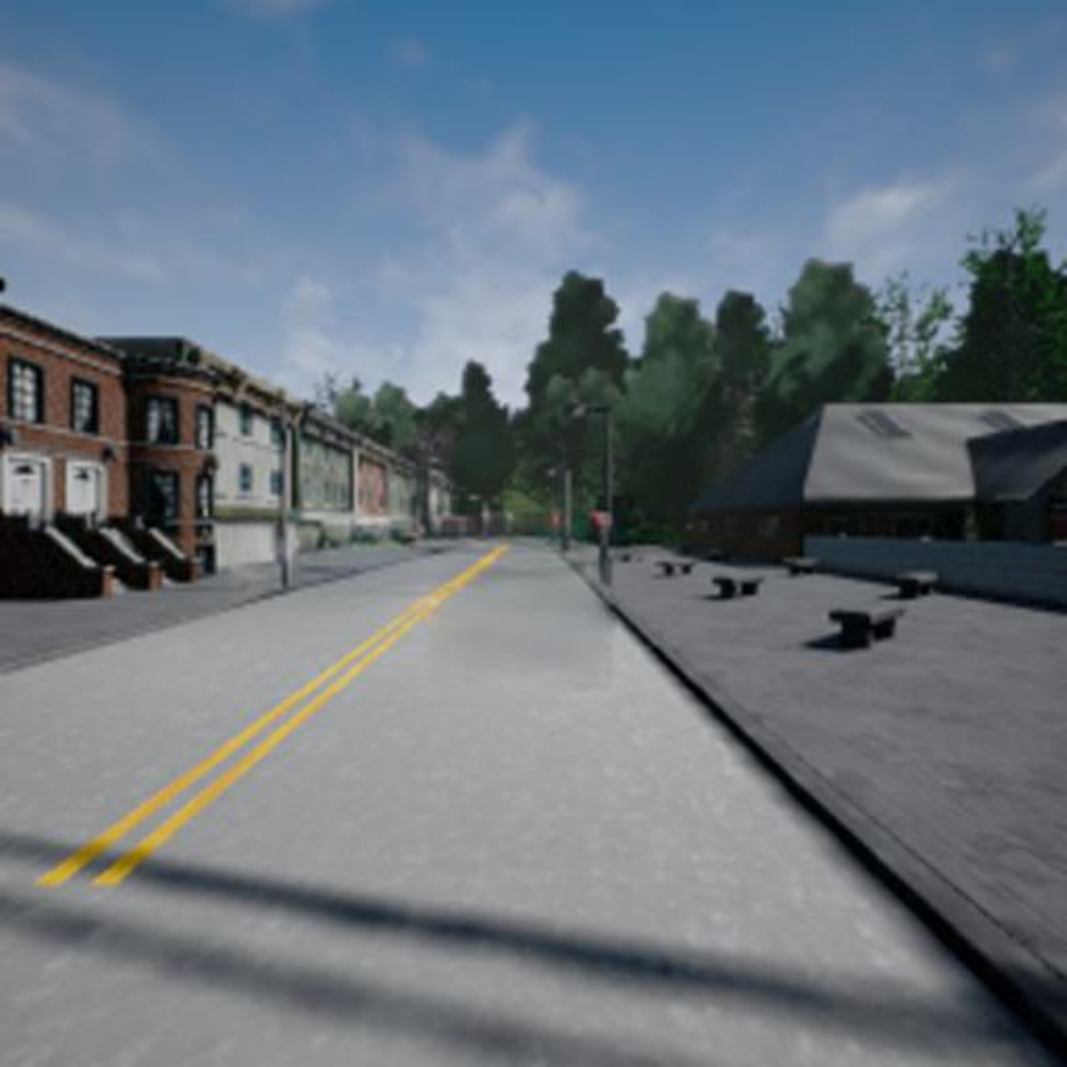}};
        \spy[magnification=\mag1] on (5*\myImgWd-\offx1, -0*\myImgHd-\offy1) in node at (5*\myImgWd, -0*\myImgHd);
        \node[tight] (n10) at (5*\myImgWd, -1*\myImgHd) {\includegraphics[width=0.14\linewidth,valign=m]{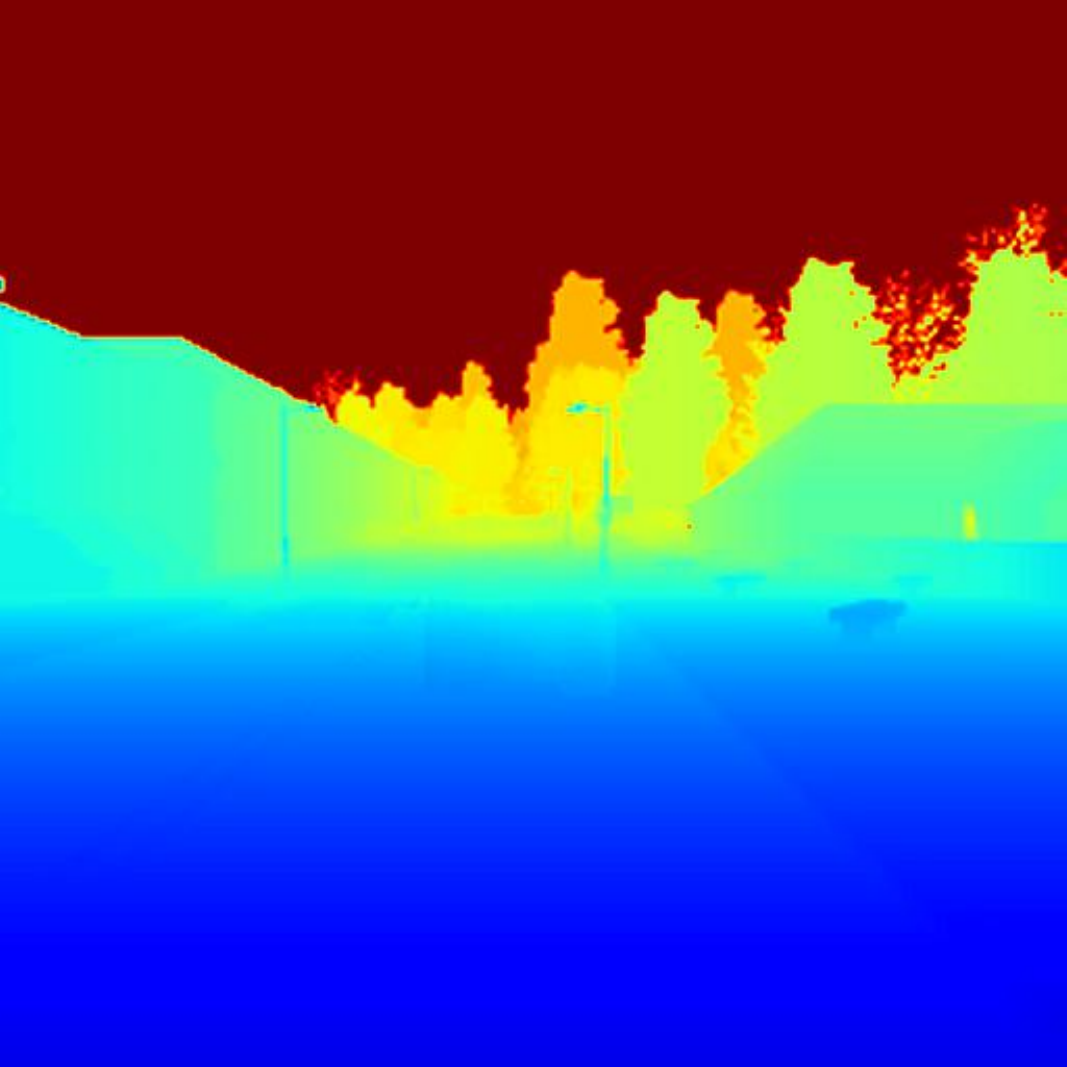}};
        \spy[magnification=\mag1] on (5*\myImgWd-\offx1, -1*\myImgHd-\offy1) in node at (5*\myImgWd, -1*\myImgHd);
        
        \node[tight] (n9) at (6*\myImgWd, -0*\myImgHd) {\includegraphics[width=0.14\linewidth]{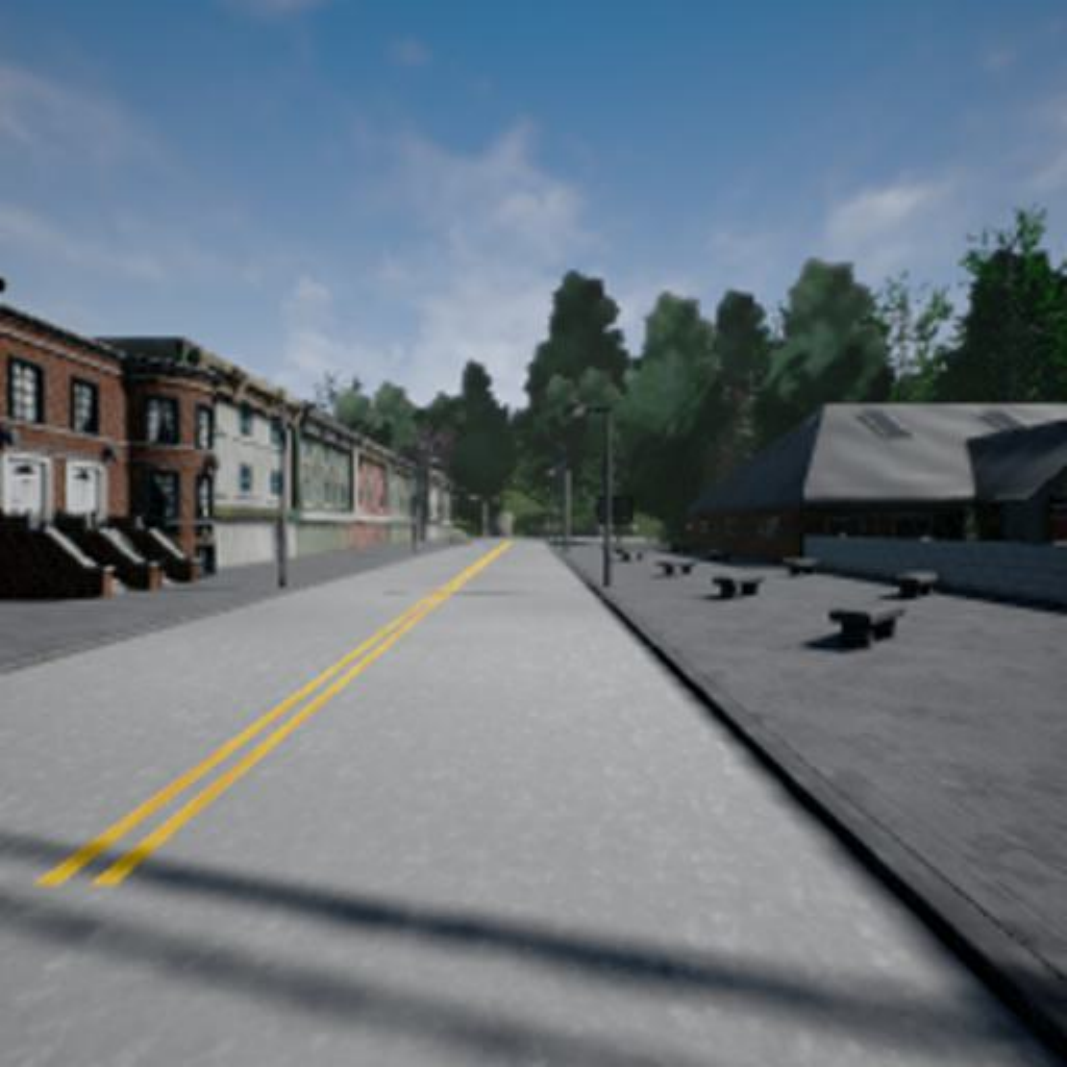}};
        \spy[magnification=\mag1] on (6*\myImgWd-\offx1, -0*\myImgHd-\offy1) in node at (6*\myImgWd, -0*\myImgHd);
        \node[tight] (n10) at (6*\myImgWd, -1*\myImgHd) {\includegraphics[width=0.14\linewidth,valign=m]{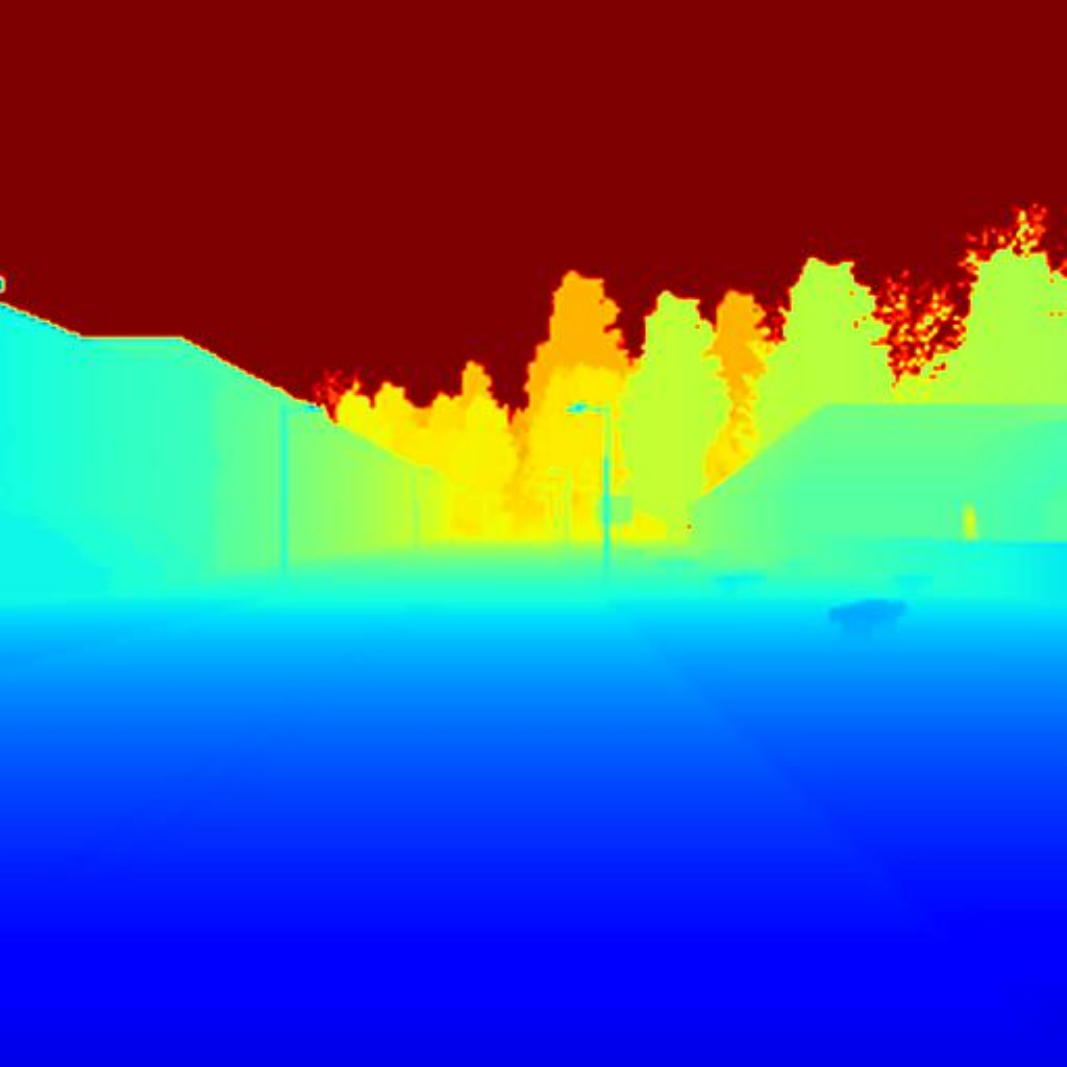}};
        \spy[magnification=\mag1] on (6*\myImgWd-\offx1, -1*\myImgHd-\offy1) in node at (6*\myImgWd, -1*\myImgHd);
        
        \draw[draw=red] (-\offx1-\myImgWd/\mag1/2, -0*\myImgHd-\offy1-\myImgHd/\mag1/2) 
        rectangle ++(\myImgWd / \mag1, \myImgHd / \mag1);
        
        \draw[draw=black] (-\offx1-\myImgWd/\mag1/2, -1*\myImgHd-\offy1-\myImgHd/\mag1/2) 
        rectangle ++(\myImgWd / \mag1, \myImgHd / \mag1);
    \end{tikzpicture}
    \caption{Qualitative comparison for diminishing objects from DynaFill~\cite{bevsic2020dynamic}.}
    \label{fig:comparison_dyna}
    \vspace{-0.25cm}
\end{figure*}
\subsection{Qualitative results}
For qualitative analysis, we compare the performance of our method to the baselines for our DR use case. Thus, we diminish objects existing in the scene to show how well the models can hallucinate realistic background textures and structures coherent with the scene semantics. Note that, for this use case, no ground truth data exists on InteriorNet and ScanNet (\cref{fig:comparison_interior_scan}). Results on DynaFill, which provides ground truth, are shown in \cref{fig:comparison_dyna}. More qualitative examples are given in the supplementary.
DeepDR is able to produce high-quality RGB textures while preserving the structure of the scene. \cref{fig:comparison_interior_scan} shows that although it was trained on purely synthetic data, it can generalize well to real-world examples from ScanNet. Its abilities are particularly evident for complex and textured backgrounds, where other methods tend to produce artifacts or overly smooth results. The benefits of our explicit structural guidance using RGB-D SPADE are also evident: While the baseline methods have difficulties in reconstructing clean borders and sharp edges, our method can recreate them well. Further, it is evident that the baseline depth completion fails at filling complex depth regions with sharp edges (\eg,  between floors and walls), in particular for structures far away from the camera. This observation reveals that sequential approaches suffer from the loss of detail and sharp features in inpainted images. 

\subsection{User study}
We conducted a repeated measures within-subjects user study to demonstrate that our framework can surpass existing works in the task of object removal for DR, and enables advanced 3D scene editing. 
We used 12 scenes from our InteriorNet testing dataset, in each of which we diminished one object in 50-200 consecutive frames. To illustrate the importance of coherent color, structure, and geometry inpainting, we reconstructed a textured 3D mesh from each inpainted RGB-D pair and augmented the reconstructed scene with additional light sources and furniture objects, as shown in~\cref{fig:abstract} and the supplementary \cref{fig:dr:comparison_editing}.
The sequences were presented to the participants in random order, side-by-side with the original input sequence, in which the object of interest was highlighted. The participants were asked to rate each item on a 7-point scale from 1 (``very poor'') to 7 (``very well''). 

\begin{figure}[htb]
    \centering
     \includegraphics[width=\columnwidth,trim=0cm 0.30cm 0cm 0.25cm,clip]{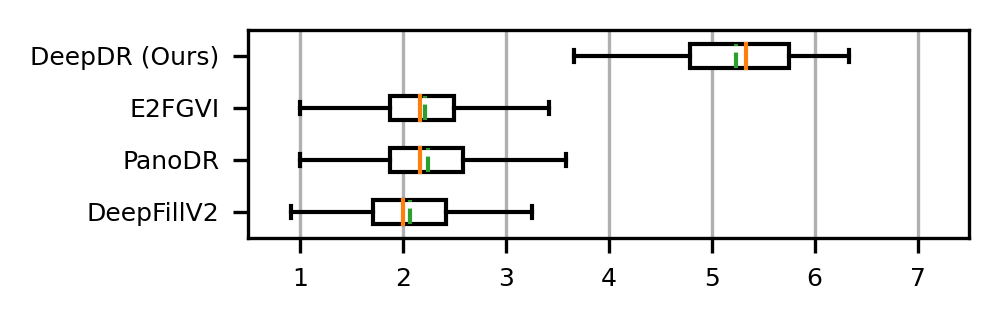}
\vspace{-0.55cm}
    \caption{Average user rating over all participants and samples from 1 (``very poor'') to 7 (``very well'').}
    \label{fig:user_study}
\vspace{-0.4cm}
\end{figure}%

\textbf{Results.} 64 participants (15 female, age 34.0$\pm$9.4) completed the study. The average experience of users with AR/DR and inpainting was 2.4$\pm$1.4 and 2.3$\pm$1.2 on a 5-point scale from ``none'' to ``expert'', respectively. The user ratings are shown in~\cref{fig:user_study}. It is evident that users rate the plausibility and realism of images and geometry inpainted with our method higher than those of others. A Friedman test revealed a significant ($\rchi^2(3)=133.3$, $W=0.7$, $p < 0.001$) difference in inpainting methods. A post-hoc Wilcoxon signed-rank test indicates that the median rating of our method (5.3) is substantially higher than that of DeepFillV2 (2.0, $p < 0.001$), PanoDR (2.2, $p < 0.001$), and E2GFVI (2.2, $p < 0.001$). No significant differences were found between other methods. We conclude that users prefer our DR results in terms of realism and plausibility.

\begin{table}[htb]
  \centering
  \caption{Ablation studies of our model on InteriorNet~\cite{li2018interiornet}.}
  \resizebox{\columnwidth}{!}{
    \begin{tabular}{l |*{4}{c}|c|*{1}{c}}
    \toprule
     & \multicolumn{4}{c|}{RGB} & \multicolumn{1}{c|}{Depth} & \multicolumn{1}{c}{Video} \\
    \cmidrule{2-7} 
    Model & LPIPS $\downarrow$ & FID $\downarrow$ & PSNR $\uparrow$ & MAE $\downarrow$ & RMSE $\downarrow$ & VFID $\downarrow$ \\
    \midrule
    no temporal & 0.0160 & 0.567 & 40.0 & 0.0336 & 0.358 & 0.0487 \\
    no RGBD SPADE & 0.0143 & 0.435 & 40.1 & 0.0408 & 0.374 & 0.0363 \\
    joint encoder & 0.0121 & 0.333 & 40.9 & 0.0340 & 0.306 & 0.0322 \\
    \midrule
    DeepDR (Full) & \textbf{0.0104} & \textbf{0.218} & \textbf{41.9} & \textbf{0.0311} & \textbf{0.278} & \textbf{0.0257} \\
    \bottomrule
    \end{tabular}
    }
  \label{tab:ablation}%
\end{table}%

\subsection{Ablation study}
To demonstrate the effectiveness of the core components of our model, we perform three ablation studies on InteriorNet, see~\cref{tab:ablation}. 
First, we remove temporal coherence from our model by omitting the auxiliary inputs $I^{t-1}$, $D^{t-1}$, $I^{t-1}_o$ and $D^{t-1}_o$, remove the ConvLSTM layer from our architecture and train without temporal loss  $\mathcal{L}_{t}$. As expected, this removal leads to lower perceptual similarity of videos (VFID). Generally, a deteriorated performance is observed, which suggests that our final model is effective in leveraging information from previous frames to fill missing regions.   
Second, we evaluate a model without structural guidance, by replacing the RGB-D SPADE layers in our up blocks with standard transposed convolutions. No intermediate segmentations are available, therefore segmentation loss $\mathcal{L}_{seg}$ is omitted. Evidently, the additional supervision via segmentation is beneficial for our model all along the line. The cost of RGB-D SPADE is an almost doubled inference time (see supplementary~\cref{tab:ablation_suppl}), which could be a limitation for real-time applications on less capable hardware.
Third, we replace our separate encoders with a joint coarse-to-fine encoder. It is apparent that our separate encoder is more effective in extracting appropriate features. 
\subsection{Limitations}

\def\offy2{-0.2}
\def\offx2{0.0}
\def\mag2{2.5}

\begin{figure}[t]
\vspace{-0.2cm}
    \centering
    \noindent
    \begin{tikzpicture}[spy using outlines={rectangle, width=1.6cm, height=1.2cm, white}, remember picture]
        \draw (0,\myImgHs / 1.8) node {\footnotesize{Input}};
        \node[tight] (n1) at (0,0) {\includegraphics[width=0.19\columnwidth]{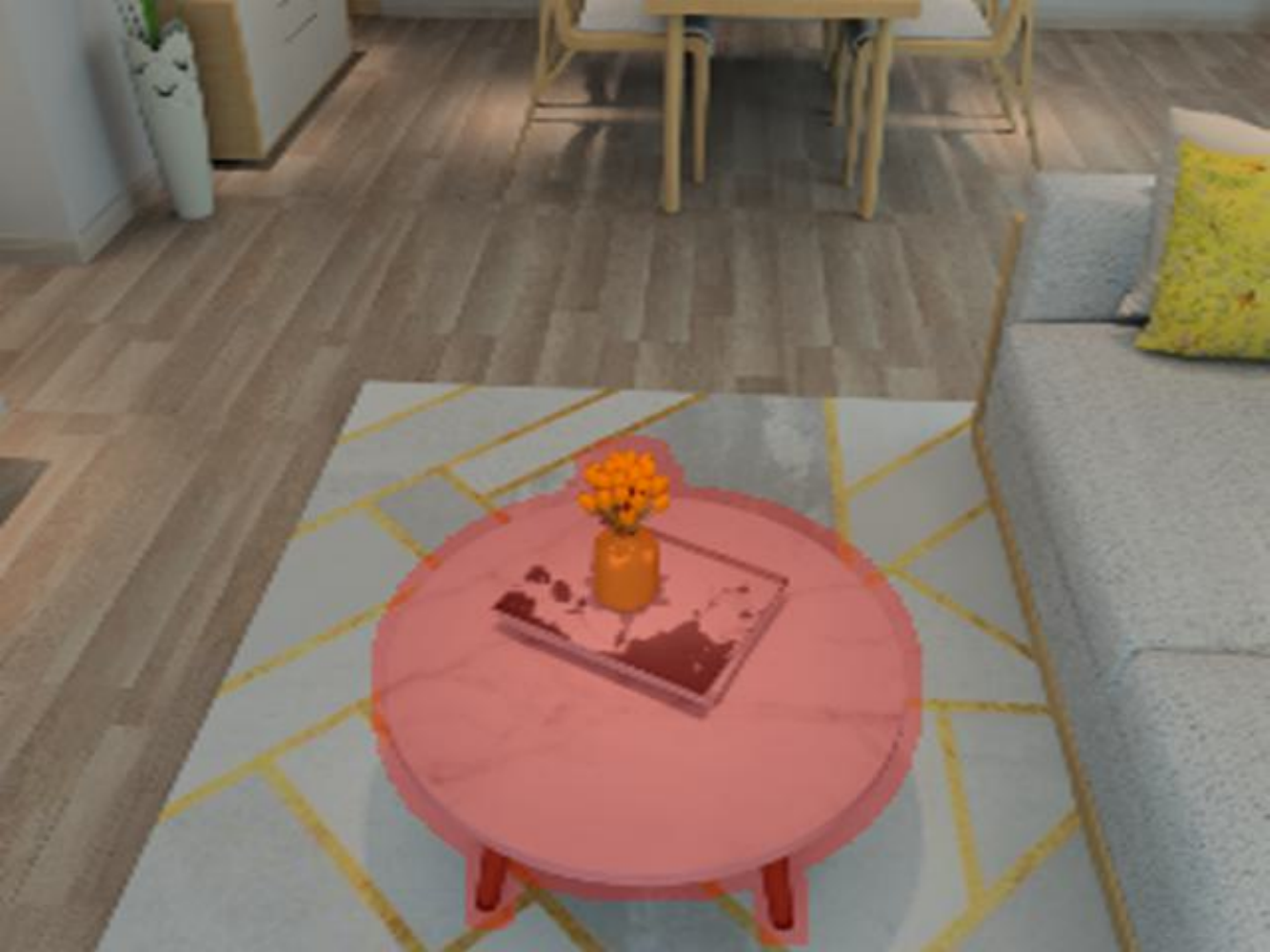}};
        \draw[draw=red] (0.0-0.4, -0.27-0.3) rectangle ++(0.85, 0.6);
        
        \draw (\myImgWs,\myImgHs / 1.8) node {\footnotesize{DeepFillV2}};
        \node[tight] (n3) at (\myImgWs, 0) {\includegraphics[width=0.19\columnwidth]{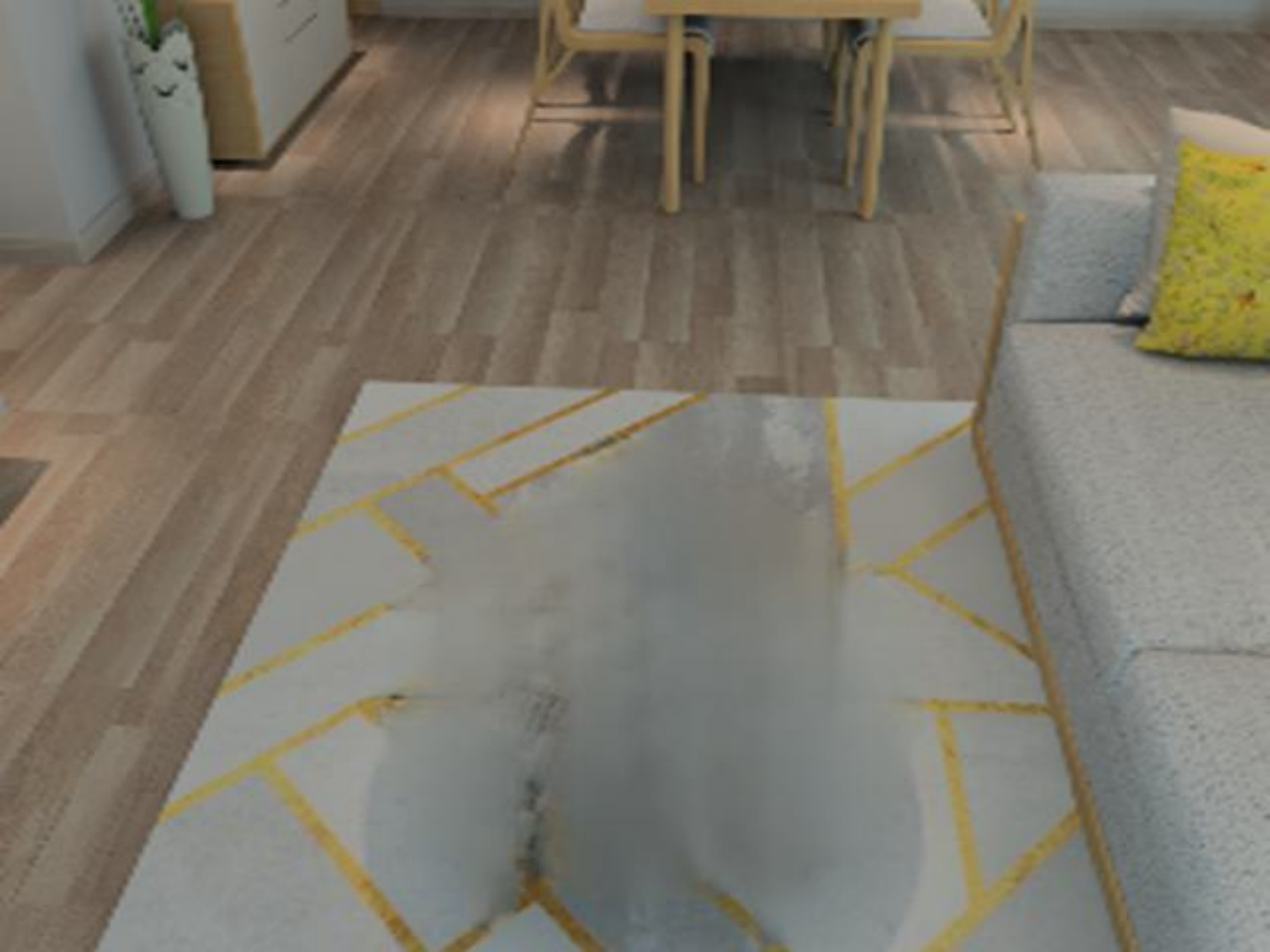}};
        \spy [magnification=1.8] on (\myImgWs, -0.27) in node at (\myImgWs, 0);
        
        \draw (2*\myImgWs,\myImgHs / 1.8) node {\footnotesize{PanoDR}};
        \node[tight] (n5) at (2*\myImgWs,0) {\includegraphics[width=0.19\columnwidth]{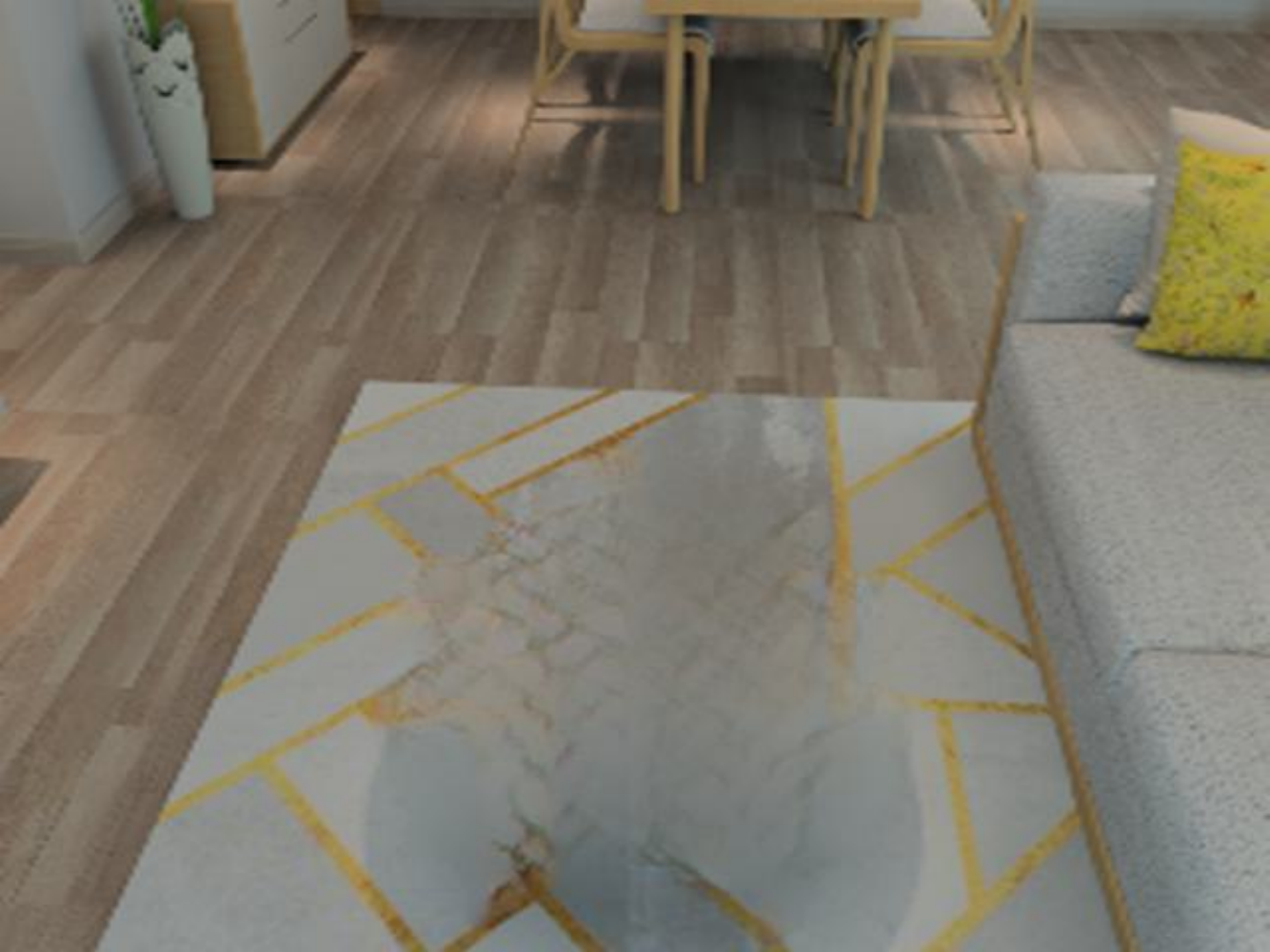}};
        \spy [magnification=1.8] on (2*\myImgWs, -0.27) in node at (2*\myImgWs, 0);
        
        \draw (3*\myImgWs,\myImgHs / 1.8) node {\footnotesize{E2FGVI}};
        \node[tight] (n7) at (3*\myImgWs,0) {\includegraphics[width=0.19\columnwidth]{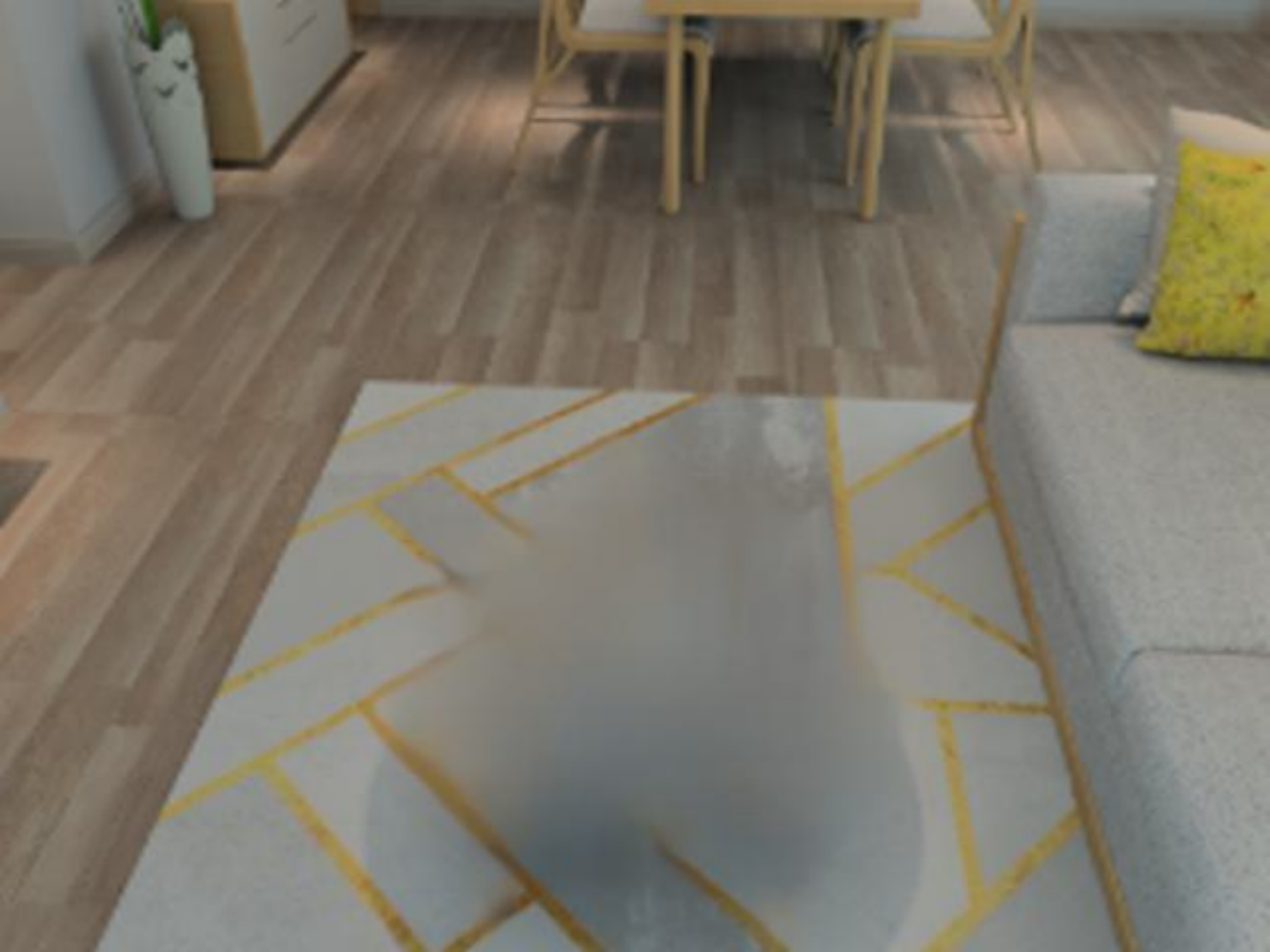}};
        \spy [magnification=1.8] on (3*\myImgWs, -0.27) in node at (3*\myImgWs, 0);
        
        \draw (4*\myImgWs,\myImgHs / 1.8) node {\footnotesize{DeepDR}};
        \node[tight] (n9) at (4*\myImgWs,0) {\includegraphics[width=0.19\columnwidth]{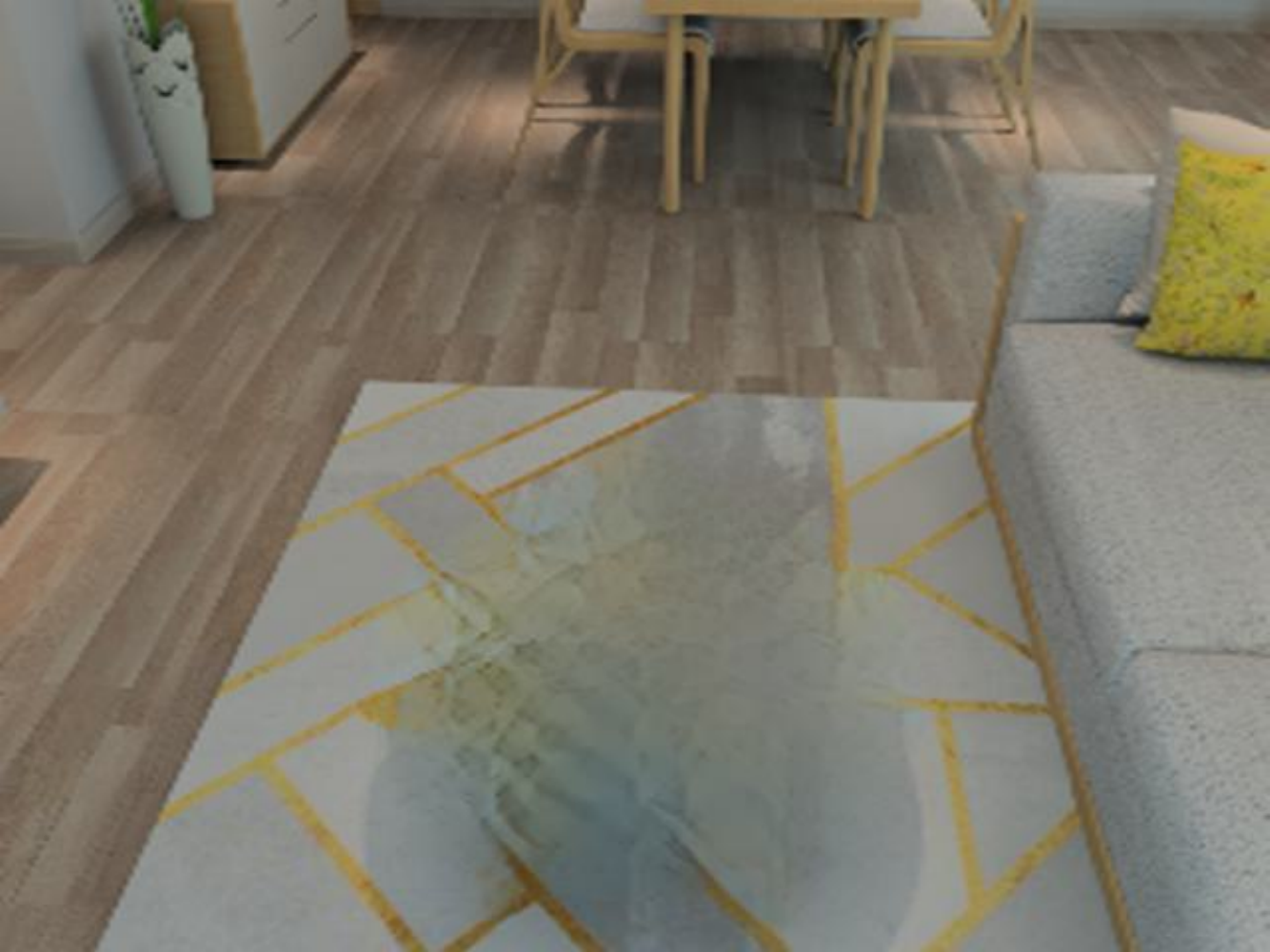}};
        \spy [magnification=1.8] on (4*\myImgWs, -0.27) in node at (4*\myImgWs, 0);
        
        \draw (-0.15,-\myImgHsd +0.8) node {\footnotesize{Input}};
        \node[tight] (n1) at (-0.15,-\myImgHsd) {\includegraphics[width=0.155\columnwidth]{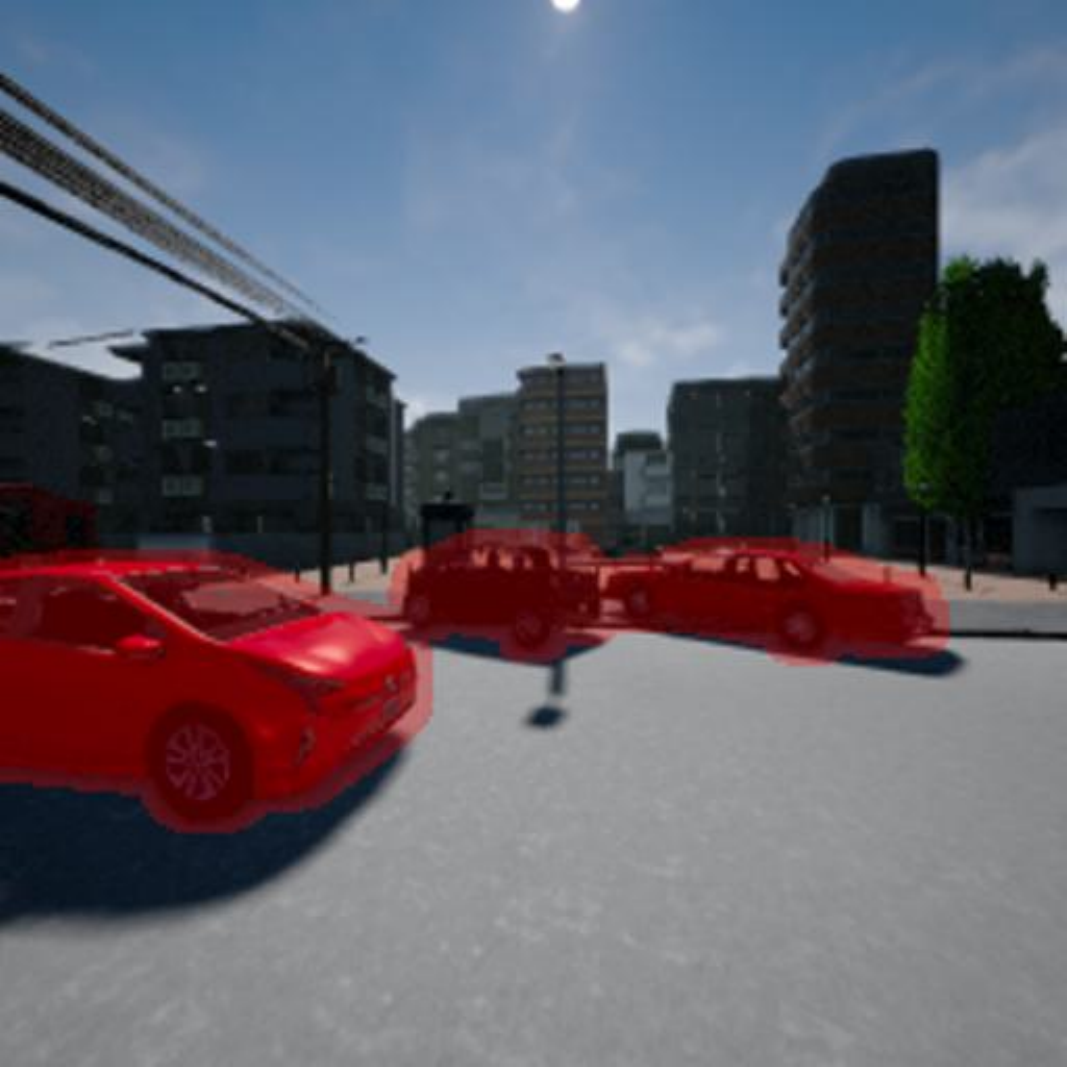}};
        \draw[draw=red] (-0.15-0.32-\offy2, -\myImgHsd-\offx2-0.32) rectangle ++(0.65, 0.65);
        
        \draw (-0.12+\myImgWsd,-\myImgHsd +0.8) node {\footnotesize{DeepFillV2}};
        \node[tight] (n3) at (-0.12+\myImgWsd, -\myImgHsd) {
        \includegraphics[width=0.155\columnwidth]{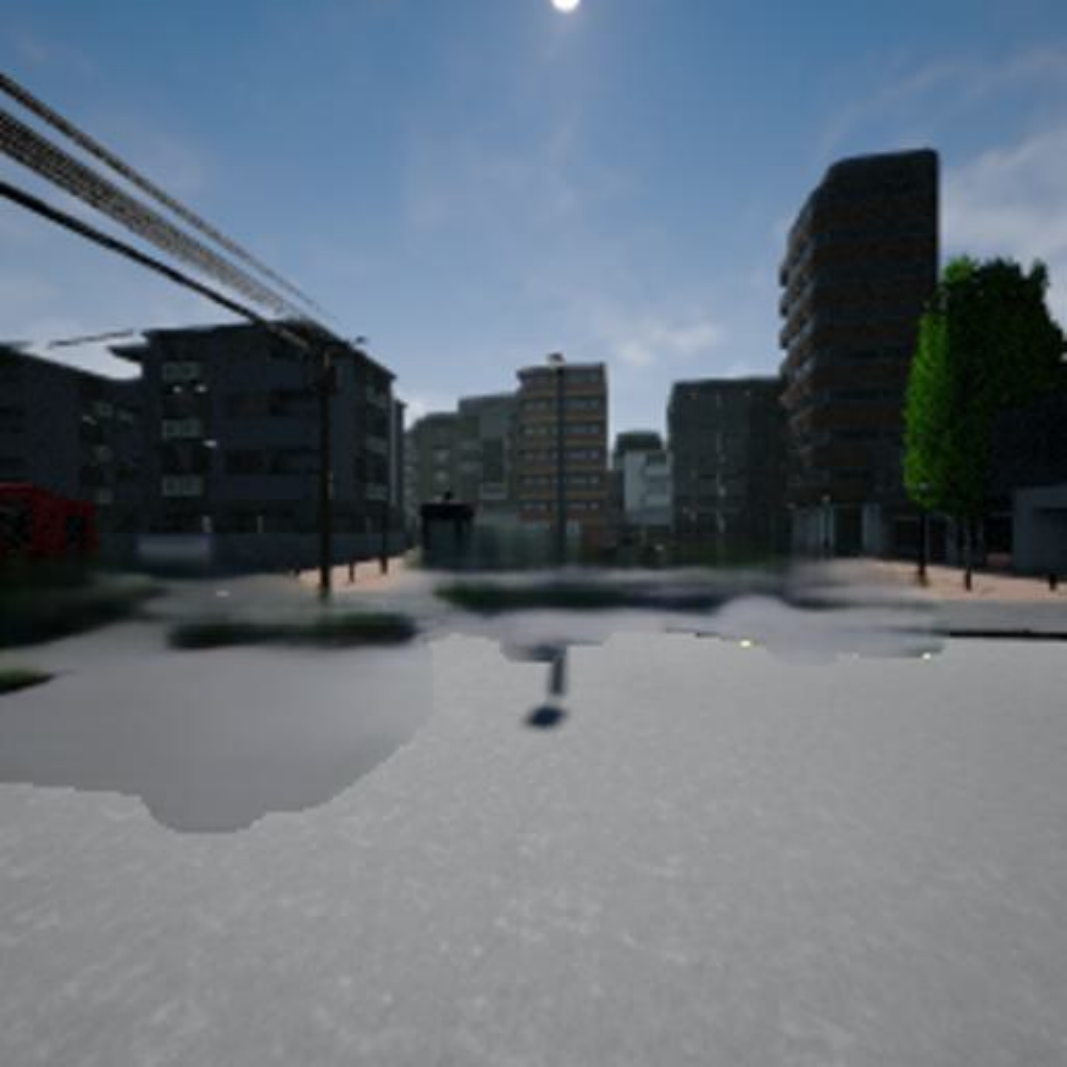}
        };
        \spy [magnification=\mag2, width=1.3cm, height=1.3cm] on (-0.12+\myImgWsd-\offy2, -\myImgHsd-\offx2) in node at (-0.12+\myImgWsd, -\myImgHsd);
        
        \draw (-0.09+2*\myImgWsd,-\myImgHsd +0.8) node {\footnotesize{PanoDR}};
        \node[tight] (n5) at (-0.09+2*\myImgWsd,-\myImgHsd) {
        \includegraphics[width=0.155\columnwidth]{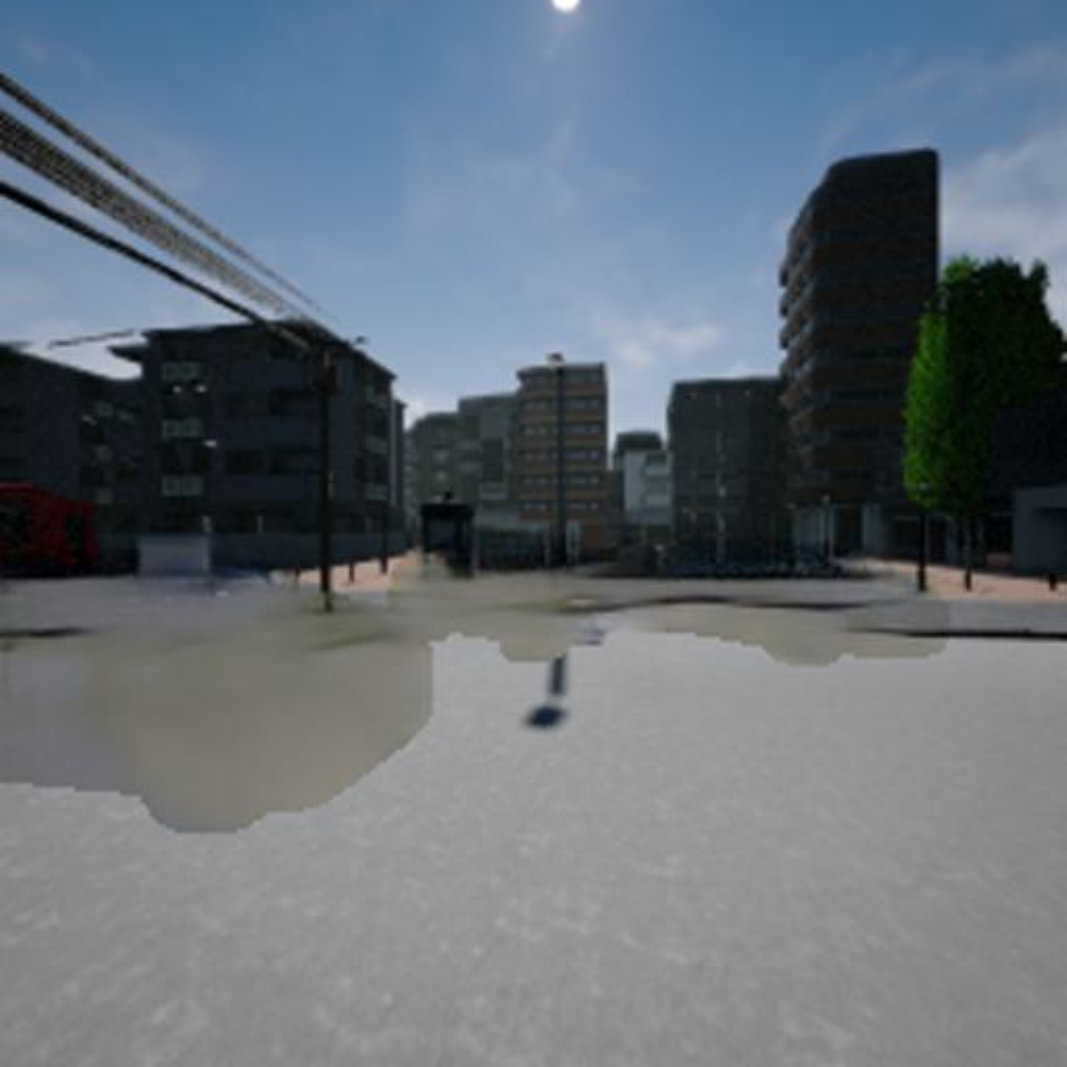}
        };
        \spy [magnification=\mag2, width=1.3cm, height=1.3cm] on (-0.09+2*\myImgWsd-\offy2, -\myImgHsd-\offx2) in node at (-0.09+2*\myImgWsd, -\myImgHsd);
        
        \draw (-0.06+3*\myImgWsd,-\myImgHsd +0.8) node {\footnotesize{E2FGVI}};
        \node[tight] (n7) at (-0.06+3*\myImgWsd,-\myImgHsd){
        \includegraphics[width=0.155\columnwidth]{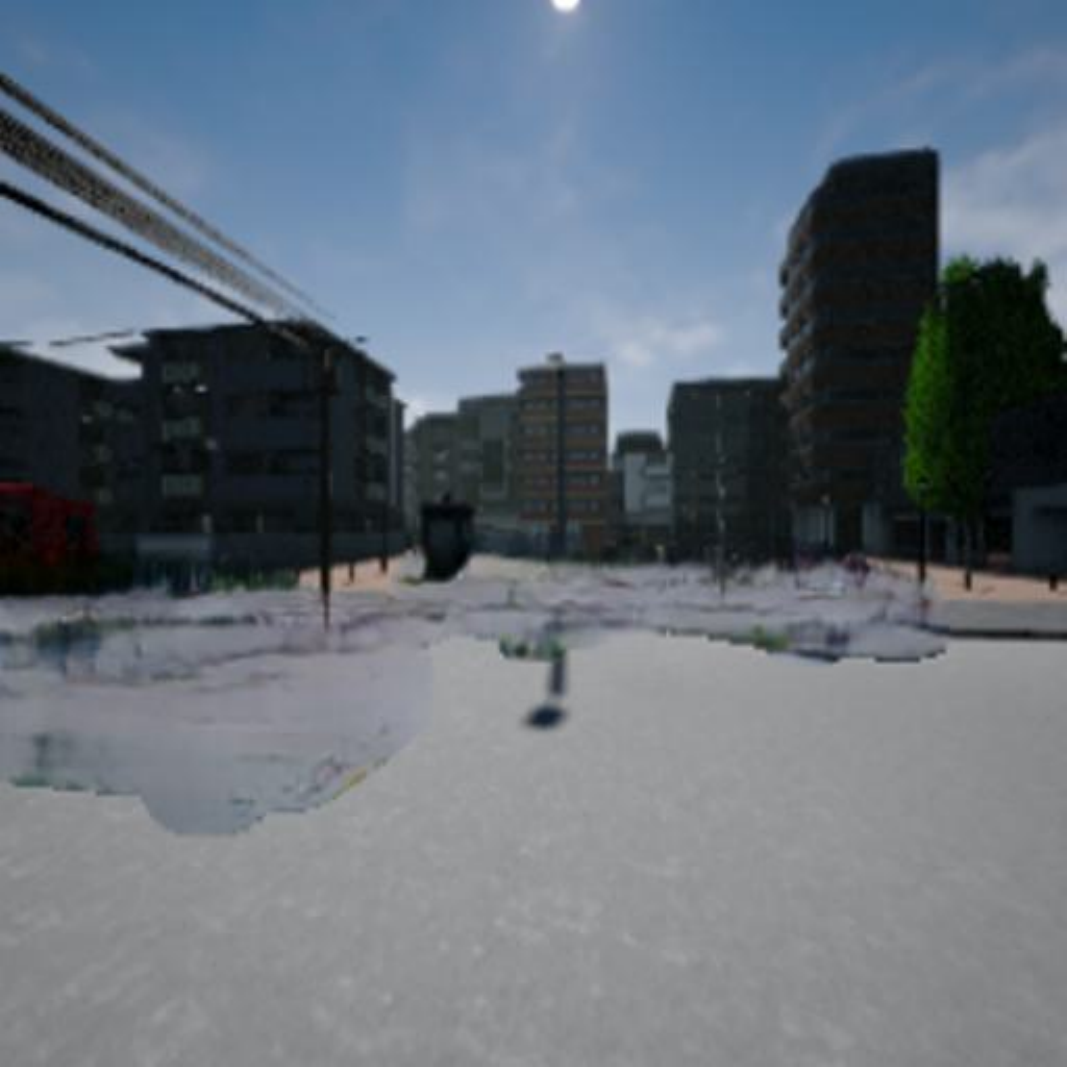}
        };
        \spy [magnification=\mag2, width=1.3cm, height=1.3cm] on (-0.06+3*\myImgWsd-\offy2, -\myImgHsd-\offx2) in node at (-0.06+3*\myImgWsd, -\myImgHsd);

        \draw (-0.03+4*\myImgWsd,-\myImgHsd +0.8) node {\footnotesize{DynaFill}};
        \node[tight] (n7) at (-0.03+4*\myImgWsd,-\myImgHsd){
        \includegraphics[width=0.155\columnwidth]{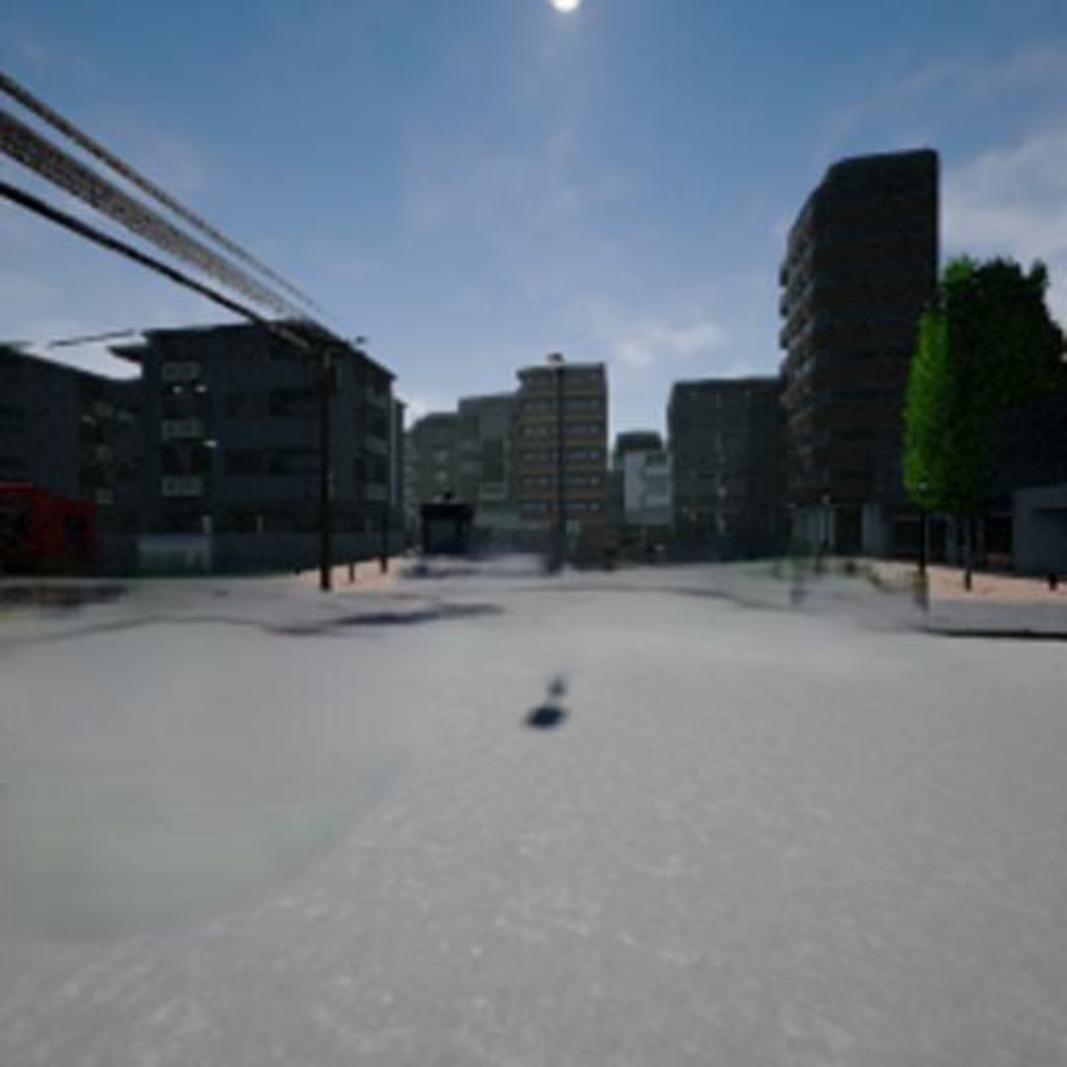}
        };
        \spy [magnification=\mag2, width=1.3cm, height=1.3cm] on (-0.03+4*\myImgWsd-\offy2, -\myImgHsd-\offx2) in node at (-0.03+4*\myImgWsd, -\myImgHsd);
        
        \draw (5*\myImgWsd,-\myImgHsd +0.8) node {\footnotesize{DeepDR}};
        \node[tight] (n9) at (5*\myImgWsd,-\myImgHsd) {\includegraphics[width=0.155\columnwidth]{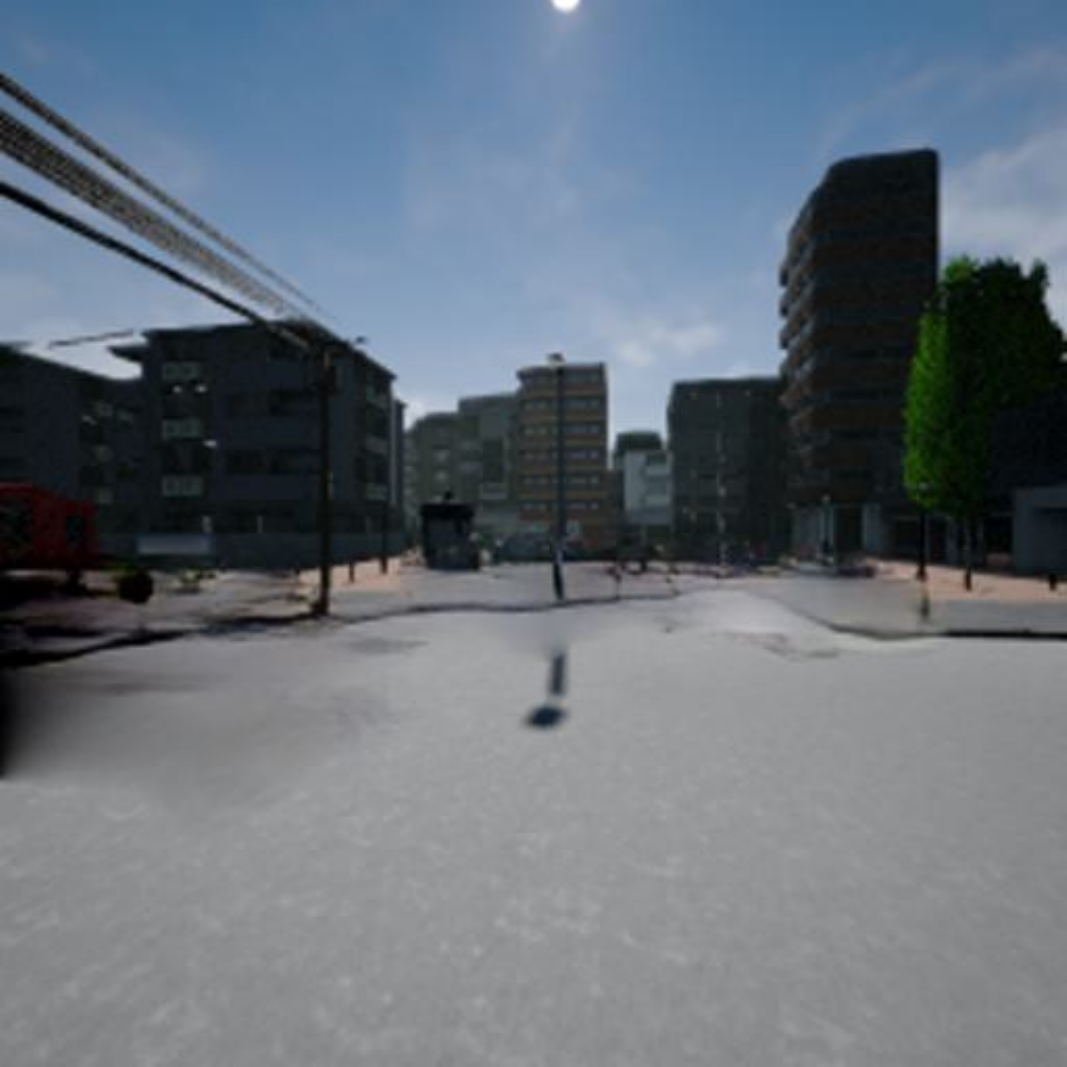}};
        \spy [magnification=\mag2, width=1.3cm, height=1.3cm] on (5*\myImgWsd-\offy2, -\myImgHsd-\offx2) in node at (5*\myImgWsd, -\myImgHsd);
    \end{tikzpicture}
    \caption{Failure cases. All current methods struggle with irregularly textured objects (\eg, the carpet, top) and highly ambiguous object borders (\eg the curb, bottom).}
    \label{fig:failure}
\vspace{-0.25cm}
\end{figure}
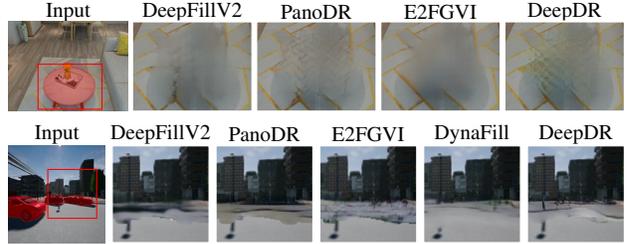

For large diminished areas in front of highly irregularly textured objects (\eg,  the carpet in~\cref{fig:failure}, top), DeepDR may generate structural artifacts, while other methods tend to over-smoothed backgrounds. Furthermore, all methods struggle with completing highly ambiguous object boundaries (see ~\cref{fig:failure}, bottom), which may cause color bleeding artifacts or implausible borders. Lastly, since InteriorNet does not have ground truth for object removal and our mask generation does not include shadows cast by objects of interest, models trained on this dataset do not learn to remove those shadows. Shadow borders are very ambiguous once the object casting them has been removed, in particular, since shaded and un-shaded regions usually belong to the same semantic class, which may lead to artifacts. We analyze this effect in the supplementary \cref{sec:shadow_suppl}.

\section{Conclusion}
We introduced DeepDR, the first approach to deep, structure-aware RGB-D inpainting with temporal coherence for DR. Our generative approach uses an RGB-D SPADE decoder to exploit structural priors, consistently conditioning color and depth outputs on them at feature level. To minimize temporal artifacts, we utilize a simple recurrent architecture with a ConvLSTM, which, compared to recent video inpainting, does not require future frame information or expensive optical flow computation at inference time. Quantitative results demonstrate that DeepDR surpasses state-of-the art inpainting methods in terms of feature-based metrics, while qualitative results show that our method is capable of generating content which is perceptually plausible, realistic in the context of the scene, and blends seamlessly with the surroundings in the image and depth domain of synthetic and real data. DeepDR works better because it effectively leverages information from multiple modalities, in particular, color, depth and structure. Therefore, it has a more comprehensive understanding of the scene, and can inpaint missing regions more plausibly in 3D. 

\section*{Acknowledgement}
The work was funded by the Austrian Research Promotion Agency (FFG) project  ``TRIP - Simulation-Based Training for AI-based Interior Planning'' (BRIDGE 883658) and the Austrian Science Fund (FWF) project ``enFaced 2.0 - Instant AR Tool for Maxillofacial Surgery'' (KLI 1044). We thank xCAD Solutions GmbH for their continuous support.

{
    \small
    \bibliographystyle{ieeenat_fullname}
    \bibliography{main}
}
\clearpage
\setcounter{page}{1}
\maketitlesupplementary

\section{Implementation details}

\subsection{Architecture details}
\label{sec:architecture_suppl}

The input of our network is a 3-channel RGB image and a 1-channel depth map. RGB inputs are normalized to $[-1,1]$ and depth inputs are scaled to $[0,1]$. Our generator uses a coarse-to-fine principle. First, we use two coarse networks identical to the ones of DeepFillV2~\cite{yu2019free} to produce coarsely inpainted RGB images $\Tilde{I}^t_o$ and depth maps $\Tilde{D}^t_o$. These outputs are used to coarsely fill the missing regions in the masked input, yielding $\Tilde{I}^t_m$ and $\Tilde{D}^t_m$. These intermediate images are then concatenated along the channel dimension with the previous in- and outputs $I^{t-1}$ and $D^{t-1}$, as well as $I^{t-1}_o$ and $D^{t-1}_o$, and are fed into two separate, architecturally identical, image and depth fine encoders. In our final implementation, we set $L=3$, thus, the fine encoders uses three down-sampling layers. After separate encoding, image and depth features are fused and completed by a common completion bottleneck using dilation, followed by a ConvLSTM layer identical to the original implementation~\cite{shi2015convolutional}. Afterward, they are fed into our structure-aware decoder with $L=3$ up blocks, whose architecture is described in \cref{sec:decoder}. Finally, the architecture of our RGB and depth discriminators is identical and follows the dense, spectral-normalized patch discriminator (SN-PatchGAN) introduced in DeepFillV2~\cite{yu2019free}.

Hereafter, we denote kernel size, dilation, stride size and channel number as K, D, S, and C, respectively. 

\textbf{Coarse generator:} K5S1C24 - K3S2C48 - K3S1C48 - K3S2C96 - K3S1C96 - K3D2S1C96 - K3D4S1C96 - K3D8S1C96 - K3D16S1C96 - K3S1C96 - K3S1C96 - up-sample(2) - K3S1C48 - K3S1C48 - up-sample(2) - K3S1C24 - K3S1C12 - K3S1C*

\textbf{Refinement encoder:} K5S1C64 - K3S2C64 - K3S1C128 - K3S2C128 - K3S1C256 - K3S2C256 - K3S1C512

\textbf{Bottleneck:} concat - K3S1C512 - K3D2S1C512 - K3D4S1C512 - K3D8S1C512 - K3D16S1C512 - K3S1C512 - K3S1C512 - ConvLSTM

\textbf{Decoder:} up block C256 - up block C128 - up block C64 - K3S1C32 - K3S1C*

In the output layers, the number of channels (C*) is three for image outputs and one for depth outputs. We use gated convolutions~\cite{yu2019free}, ReLU activation and instance normalization throughout convolution layers.  Image output layers use the tanh activation function, while depth output layers clamp the output to $[0,1]$.

\subsection{Details of training objectives}
\label{sec:training_suppl}

The SN-PatchGAN discriminators $D_I$ and $D_D$ are trained using Hinge loss~\cite{lim2017geometric}, which is widely adopted for inpainting tasks. For an incomplete image and depth map $I_m$ and $D_m$, and their ground truth counterparts $I$ and $D$, the adversarial discriminator losses are
\begin{align}
\mathcal{L}_{D_I} &=
\begin{aligned}[t]
       -&\mathbb{E}_{I\sim p_d} [ReLu(-\mathbbm{1} + D_I(I))] - \\
        &\mathbb{E}_{I_m\sim p_z} [ReLu(-\mathbbm{1} - D_I(G_I(I_m)))],   
\end{aligned} \\
\mathcal{L}_{D_D} &=
\begin{aligned}[t]
     -&\mathbb{E}_{D\sim p_d} [ReLu(-\mathbbm{1} + D_D(D))] - \\
      &\mathbb{E}_{D_m\sim p_z} [ReLu(-\mathbbm{1} - D_D(G_D(D_m)))],  
\end{aligned}
\end{align}
while the generator losses are defined as
\begin{align}
          \mathcal{L}_{adv,I}^G &= -\mathbb{E}_{I_m\sim p_z}[D_I(G_I(I_m))], \\ 
          \mathcal{L}_{adv,D}^G &= -\mathbb{E}_{D_m\sim p_z}[D_D(G_D(I_m))].  
\end{align}
Here, $p_d$ and $p_z$ are the distributions of real data and the latent space, respectively, and $\mathbb{E}_{I\sim p_d}$ denotes the expectation value of $I$ with respect to distribution $p_d$.

The $\ell_1$-reconstruction loss on a pixel level is computed between ground truth images and depth $I$ and $D$ and the corresponding generator outputs $I_o$ and $D_o$ as
\begin{align}
    \mathcal{L}_{rec,I} &= ||I - I_o||_1, \\
    \mathcal{L}_{rec,D} &= ||D - D_o||_1. \label{eq:rec_loss}
\end{align}
Perceptual and style loss operate in feature space, by computing the distance of $i$\textsubscript{th} level features $\phi_i$ of a pre-trained network, in our case, VGG-19~\cite{simonyan2014very}. Specifically, perceptual loss is defined as 
\begin{equation}
    \mathcal{L}_{per}= \sum_i \frac{||\phi_i(I)-\phi_i(I_o)||_1}{N_i},
\end{equation}
where $N_i$ is the number of elements in $\phi_i$, and style loss is given by
\begin{equation}
    \mathcal{L}_{sty} = ||G_i^\phi(I)-G_i^\phi(I_o)||_1,
\end{equation}
with $G_i^\phi$ being the Gram matrix constructed from activation $\phi_i$.

Our weight parameters are set empirically and according to literature~\cite{liu2018image,lai2018learning} as $\lambda_{rec}=10$, $\lambda_{per}=10$, $\lambda_{sty}=250$, $\lambda_{grad}=100$, $\lambda_{seg}=10$ and $\lambda_{t}=10$.

\subsection{Hardware and training strategy}
\label{sec:training_suppl_2}

We implemented our model in PyTorch~\cite{pytorch}. For training, we use an Nvidia Quadro RTX 8000 GPU, set the batch size to four and train for 1M iterations. We use an Nvidia GeForce GTX 1080 Ti for testing. 

As training input, we select a series of $T=5$ consecutive RGB and depth frames with their semantic segmentation from our training datasets, together with randomly sampled object masks in the case of InteriorNet, and masks covering dynamic scene objects (pedestrians, vehicles) in case of DynaFill. For testing on InteriorNet and ScanNet, we set $T=100$ and use a fixed set of random object masks. Since DynaFill sequences have a shorter, varying number of frames, we set $T$ according to the sequence length. We resize all inputs to a resolution of 256$\times$256 pixels during training and testing. Adam optimizer~\cite{kingma2014adam} with $\beta_1=0.0$ and $\beta_2=0.9$ is used for optimization, and we set the learning rate to $2*10^{-4}$ for all modules. After 500k iterations, we reduce the learning rate to $2*10^{-5}$. 

\paragraph{Data augmentation.} Since InteriorNet already contains varying lighting conditions and different scene views, we do not apply additional data augmentation. Instead, to limit redundancy in the dataset, we sub-sample every fourth frame from each sequence during training. For DynaFill, we use the same data augmentation as in the original paper (brightness, contrast, saturation and hue modulation, as well as random horizontal flipping). 

\paragraph{Recurrent network training.} Recurrent network training incurs some additional computational costs, which we minimize through several strategies: Firstly, MaskFlowNet~\cite{zhao2020maskflownet} for flow estimation is very lightweight with only 10.5 M params and 13.4 G MADs. Recent, more efficient flow estimation architectures like FastFlowNet~\cite{kong2021fastflownet} could further reduce the costs. Secondly, we perform sequence truncation by processing subsets of $T=5$, while preserving the ConvLSTM hidden state for each sequence, which, according to our informal experiments, provides a good trade-off between capturing temporal dependencies and computational cost during training. In summary, recurrent training increases memory consumption by approximately 2.2 GB compared to single-image training, which we deem justified by the obtained quality gain (see \cref{tab:ablation}). Importantly, the recurrent feedback loop has minimal impact on inference efficiency compared to other video inpainting methods (see \cref{tab:ablation_suppl}).

\paragraph{Obtaining semantic segmentations and depth.} We trained our models on synthetic datasets with accurate ground-truth segmentations and depth. Our experiments (\cref{tab:comparison_scan}, \cref{fig:dr:comparison_interiornet}, \cref{fig:dr:comparison_scannet}) demonstrate their strong generalization to real-world data. As image segmentation techniques such as SAM~\cite{kirillov2023segment} advance, obtaining high-quality segmentations from various image and video datasets will likely be feasible in the near future. Depth is usually available in our targeted mixed reality systems, as they require an understanding of their 3D surroundings. If not, recent monocular depth estimation models~\cite{Ranftl2022} can be used to obtain depth.

\section{Additional experiments}

\subsection{Analysis of RGB-D datasets for object removal}
\label{sec:datasets_suppl}

Datasets suitable for DR need to contain consecutive video frames of aligned RGB and depth. Our framework furthermore requires semantic segmentations: For supervising structural guidance during training, and for generating object masks for a convenient qualitative evaluation. The few works about fused RGB-D object removal~\cite{bevsic2020dynamic,pintore2022instant,dhamo2019peeking,fujii2020rgb} use Structured3D~\cite{zheng2020structured3d}, SceneNet RGBD~\cite{mccormac2017scenenet} or DynaFill~\cite{bevsic2020dynamic} (see~\cref{tab:dr:inpainting_datasets}). Structured3D does not fulfill the criteria of consecutive frames and we did not consider SceneNet RGBD due to its poor realism. Similarly, common depth completion benchmarks are not suitable, as shown in~\cref{tab:dr:depth_datasets}.

\begin{table}[htbp]
  \centering
  \caption{Datasets used in related RGB-D object removal works.}
  \resizebox{\linewidth}{!}{
    \begin{tabular}{lccc}
    \toprule
    & Segmentation & \makecell{Consecutive \\ frames} & Photorealistic\\
    \midrule
    Structured3D~\cite{zheng2020structured3d} & \cmark      & \xmark  & \cmark \\
    SceneNet RGBD~\cite{mccormac2017scenenet} & \cmark     & \cmark  & \xmark \\
    DynaFill~\cite{bevsic2020dynamic} & \cmark & \cmark & \cmark \\
    \bottomrule
    \end{tabular}%
    }
  \label{tab:dr:inpainting_datasets}%
\end{table}%

\begin{table}[htbp]
  \centering
  \caption{Common dense depth completion datasets and their suitability for DR.}
  \resizebox{\linewidth}{!}{
    \begin{tabular}{lccc}
    \toprule
    & Segmentation & \makecell{Consecutive \\ frames} & Available\\
    \midrule
    NYU-depth V2~\cite{silberman2012indoor} & subset & subset & \cmark\\
    Middlebury~\cite{hirschmuller2007evaluation,scharstein2014high} & \xmark & \xmark & \cmark \\
    Matterport3D~\cite{chang2017matterport3d} & \cmark    & \xmark & \cmark\\
    VOID~\cite{wong2020unsupervised} & \xmark & \cmark & \cmark\\
    DIODE~\cite{wong2020unsupervised} & \xmark & \xmark & \cmark\\
    SUNCG~\cite{song2017semantic} & \cmark & \cmark & \xmark\\
    \bottomrule
    \end{tabular}%
    }
  \label{tab:dr:depth_datasets}%
\end{table}%

\subsection{Comparison of different indoor depth completion methods}
\label{sec:depth_compl_suppl}

In total, we considered three methods designed for indoor and outdoor depth completion to fill missing depth regions in baselines that don't handle depth: InDepth~\cite{zhang2022indepth}, NLSPN~\cite{park2020non} and DM-LRN~\cite{senushkin2021decoder}. These networks receive previously completed RGB and masked depth as input and are designed to leverage both RGB and depth features, with different methods to fuse them. Their goal is to fill missing depth based on \emph{complete} RGB information, which is typically unavailable in DR. In~\cref{tab:depth}, we compare their performance on our datasets, reporting root mean squared error (RMSE) for depth completion using inpainted color images from our baseline methods. Evidently, InDepth works best for InteriorNet, while NLSPN performs best on ScanNet and DynaFill. We use these best-performing methods as baselines for the comparison of depth RMSE in \cref{tab:comparison_interior}, \cref{tab:comparison_dyna} and \cref{tab:comparison_scan}. Note that we did not consider works that complete depth from sparse measurements in different scenarios, \eg, LiDAR-based depth completion in outdoor scenarios.

\begin{table*}[htbp]
  \centering
  \caption{Root mean squared errors (RMSE) for different indoor depth completion methods based on color inpainting using our baseline methods. All measurements are given in meters.}
  \resizebox{\textwidth}{!}{
    \begin{tabular}{l|ccc|ccc|ccc}
    \toprule
     & \multicolumn{3}{c|}{InteriorNet~\cite{li2018interiornet}} & \multicolumn{3}{c|}{ScanNet~\cite{dai2017scannet}} & \multicolumn{3}{c}{DynaFill~\cite{bevsic2020dynamic}}\\
    \midrule
    Model & DeepFillV2 & PanoDR & E2FGVI & DeepFillV2 & PanoDR & E2FGVI & DeepFillV2 & PanoDR & E2FGVI\\
    \midrule
    NLSPN~\cite{park2020non} & 0.706 & 0.635 &  0.619 & \textbf{0.508} & \textbf{0.536} & \textbf{0.512} & \textbf{7.92} & \textbf{8.12} & \textbf{7.83} \\
    DM-LRN~\cite{senushkin2021decoder} & 1.034 & 1.223 &  1.366 & 0.781 & 0.789 & 0.852 & 11.81 & 11.92 & 11.80 \\
    InDepth~\cite{zhang2022indepth} & \textbf{0.572} & \textbf{0.564} & \textbf{0.563} & 0.643 & 0.659 & 0.629 & 11.84 & 12.97 & 12.33 \\
    \bottomrule
    \end{tabular}%
    }
  \label{tab:depth}%
\end{table*}

\subsection{Influence of shadow mask}
\label{sec:shadow_suppl}

To fully diminish objects from a scene as if they were not there in the first place, it is also necessary to remove the shadow they cast. While automatic shadow segmentation remains a topic for our future work, we are interested in the performance of DeepDR in the case of a combined object and shadow mask. Thus, we have manually added shadow masks to the automatically derived object masks from InteriorNet and ScanNet. \cref{fig:shadowseg} provides visual results in order to demonstrate the performance of DeepDR for complete object and shadow removal. For comparison, we also provide results without a shadow mask. Apparently, DeepDR is capable of reliably inpainting shadow masks and moreover, results with masked shadows often look better than without. The reason for that is that our model does not need to hallucinate the very ambiguous shadow borders, leading to a more realistic color with fewer artifacts. 

The same observation holds for the automatically derived object masks from ScanNet, which, due to inaccurate instance segmentation in the original dataset, sometimes do not cover the entire diminished object. In such cases, artifacts and flickering between consecutive frames can appear. 

\renewcommand{\myImgHs}{1.6}
\renewcommand{\myImgWs}{2.13}
\begin{figure}[htb]
    \centering
    \begin{tikzpicture}[spy using outlines={rectangle, width=2.00cm, height=1.6cm, white}, remember picture]
        \draw (\myImgWs / 2, \myImgHs / 1.5) node {\small{Automatically created mask}};
        \draw (2*\myImgWs + \myImgWs / 2, \myImgHs / 1.5) node {\small{Manually created mask}};
        
        \node[tight] (n1) at (0, 0) {\includegraphics[width=0.24\linewidth]{examples/qual_interiornet/00_in_035_img.pdf}};
        \spy [magnification=3.4] on (0-0.10, 0-0.32) in node at (0, 0);
        
        \node[tight] (n9) at (1*\myImgWs, 0) {\includegraphics[width=0.24\linewidth]{examples/qual_interiornet/00_ours_035_img.pdf}};
        \spy [magnification=3.4] on (1*\myImgWs-0.10, 0-0.32) in node at (1*\myImgWs, 0);
        
        \node[tight] (n1) at (2*\myImgWs, 0) {\includegraphics[width=0.24\linewidth]{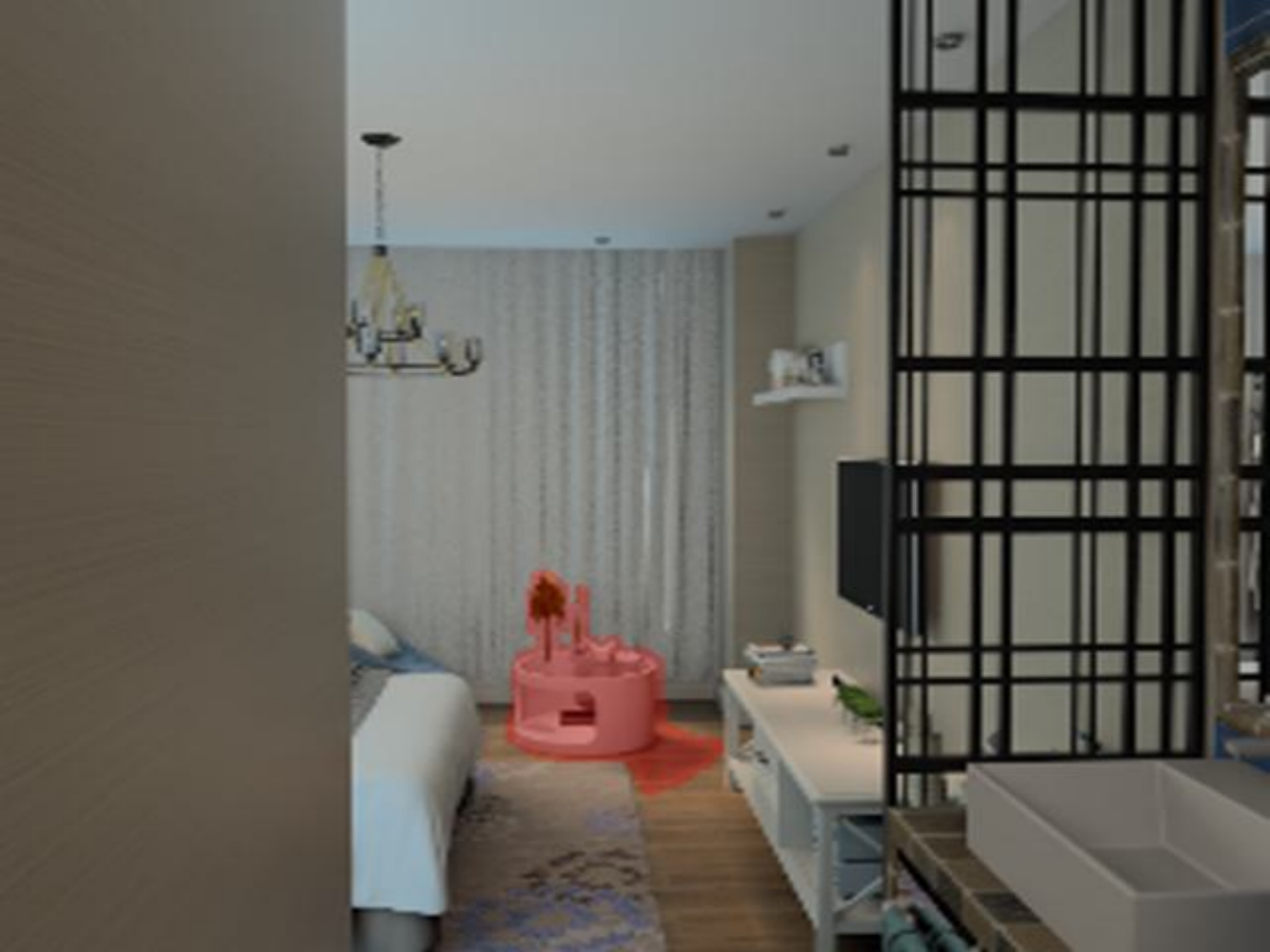}};
        \spy [magnification=3.4] on (2*\myImgWs-0.00, 0-0.32) in node at (2*\myImgWs, 0);
        
        \node[tight] (n9) at (3*\myImgWs, 0) {\includegraphics[width=0.24\linewidth]{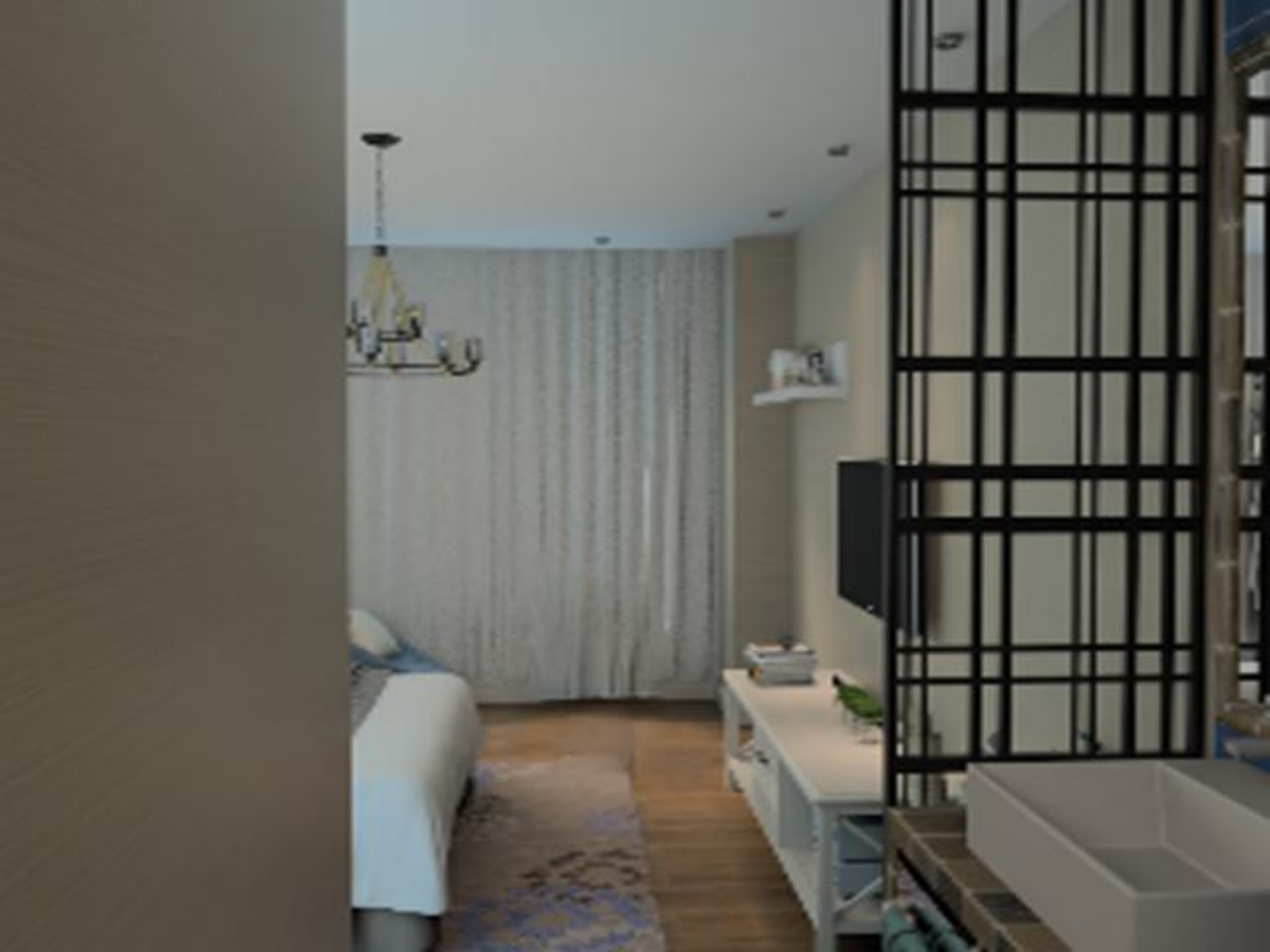}};
        \spy [magnification=3.4] on (3*\myImgWs-0.00, 0-0.32) in node at (3*\myImgWs, 0);
        
        \node[tight] (n1) at (0,-1*\myImgHs) {\includegraphics[width=00.24\linewidth]{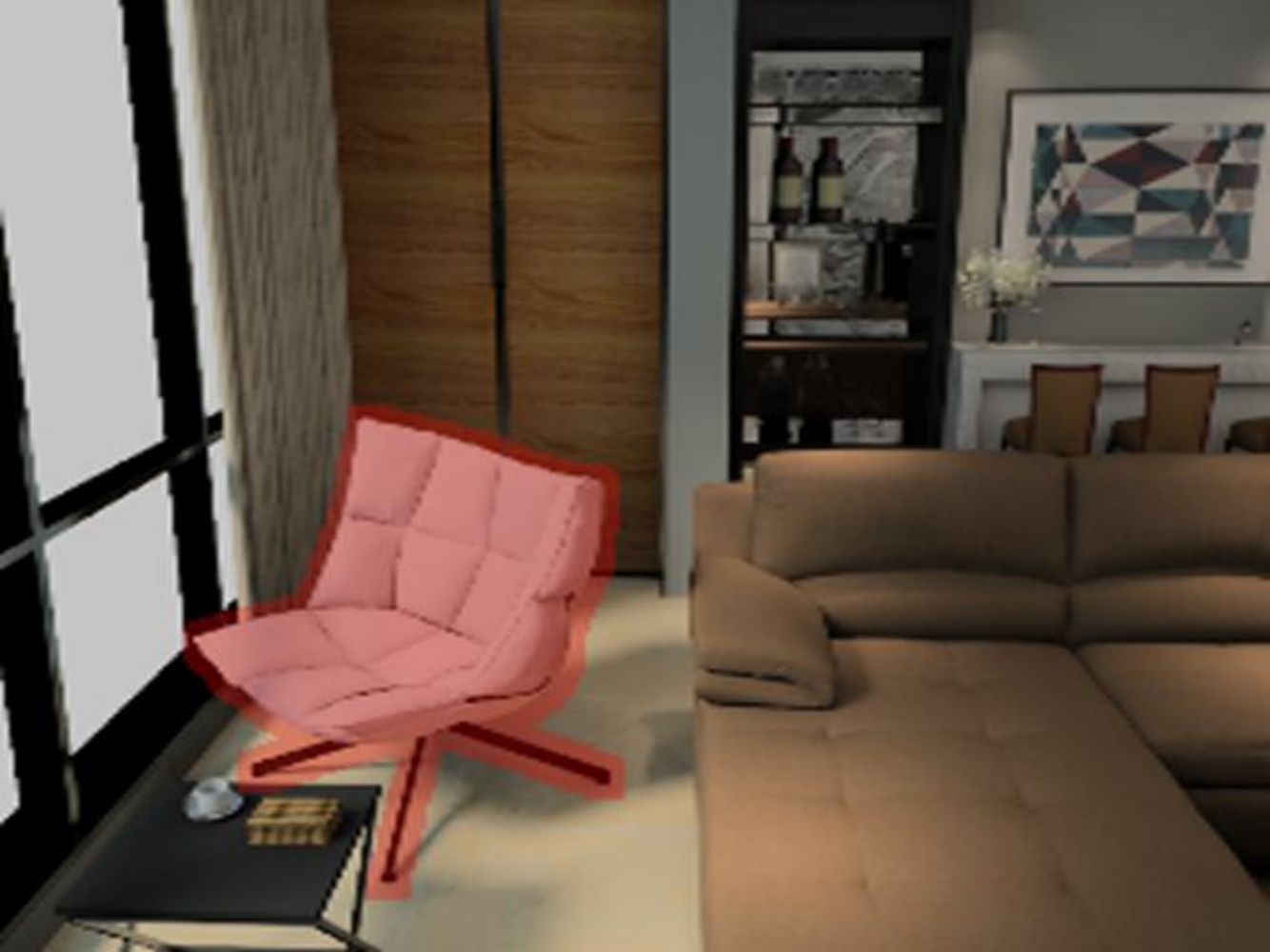}};
        \spy[magnification=1.8] on (0*\myImgWs-0.30,-0.25-1*\myImgHs) in node at (0*\myImgWs,-1*\myImgHs);
        
        \node[tight] (n9) at (1*\myImgWs,-1*\myImgHs) {\includegraphics[width=0.24\linewidth]{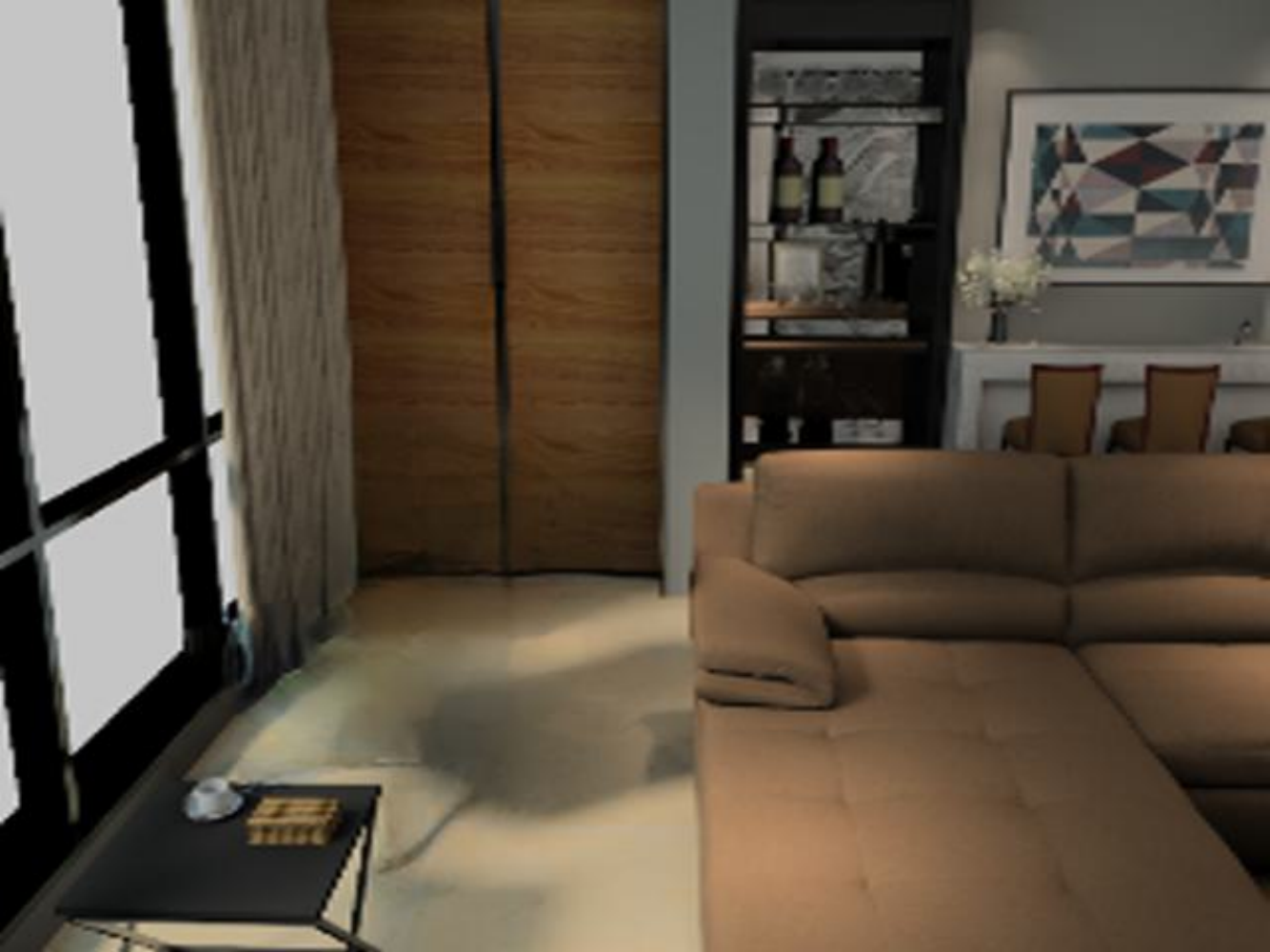}};
        \spy[magnification=1.8] on (1*\myImgWs-0.30,-0.25-1*\myImgHs) in node at (1*\myImgWs,-1*\myImgHs);
        
        \node[tight] (n1) at (2*\myImgWs, -1*\myImgHs) {\includegraphics[width=0.24\linewidth]{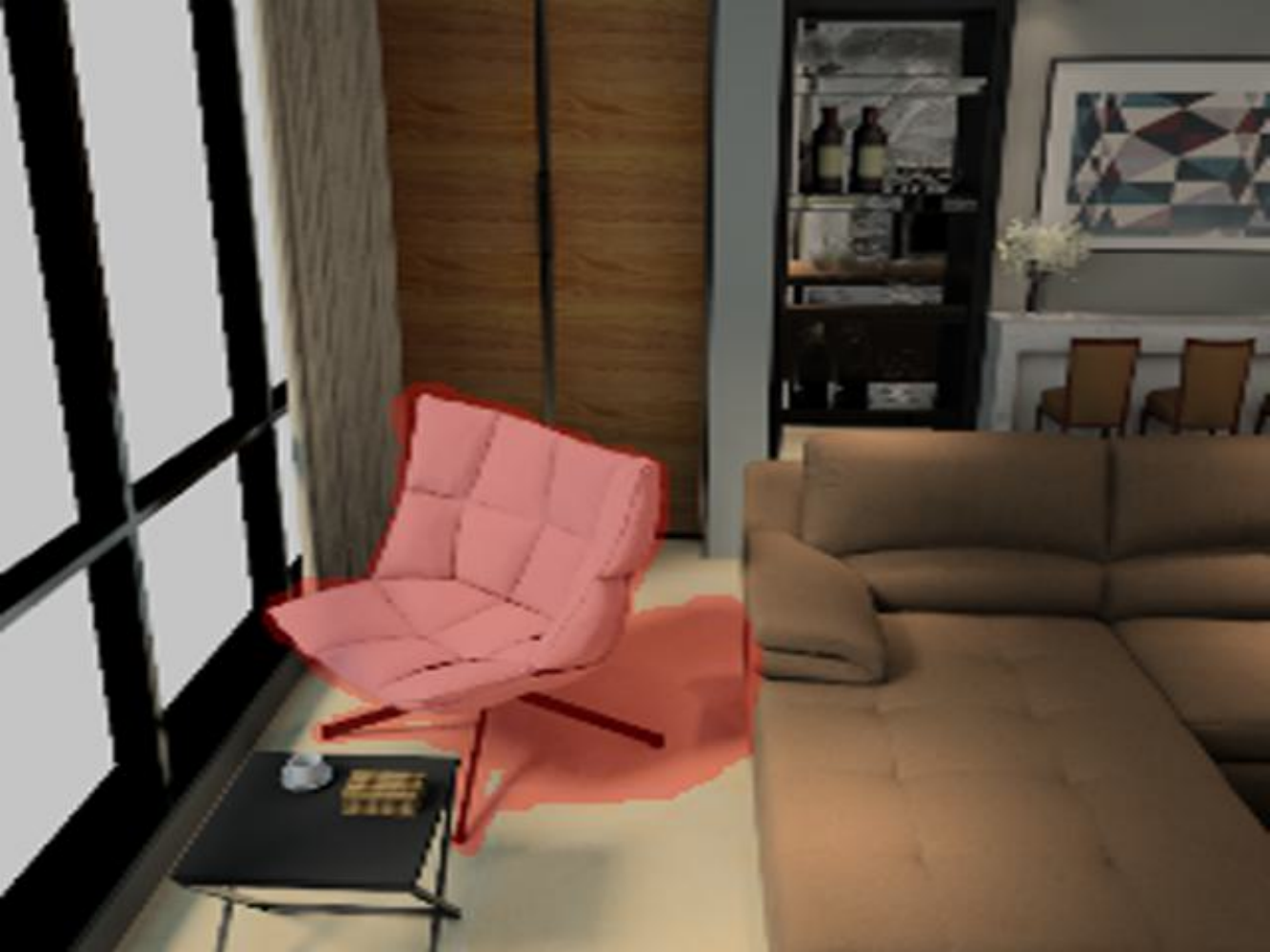}};
        \spy [magnification=1.8] on (2*\myImgWs-0.2, -1*\myImgHs-0.25) in node at (2*\myImgWs, -1*\myImgHs);
        
        \node[tight] (n9) at (3*\myImgWs, -1*\myImgHs) {\includegraphics[width=0.24\linewidth]{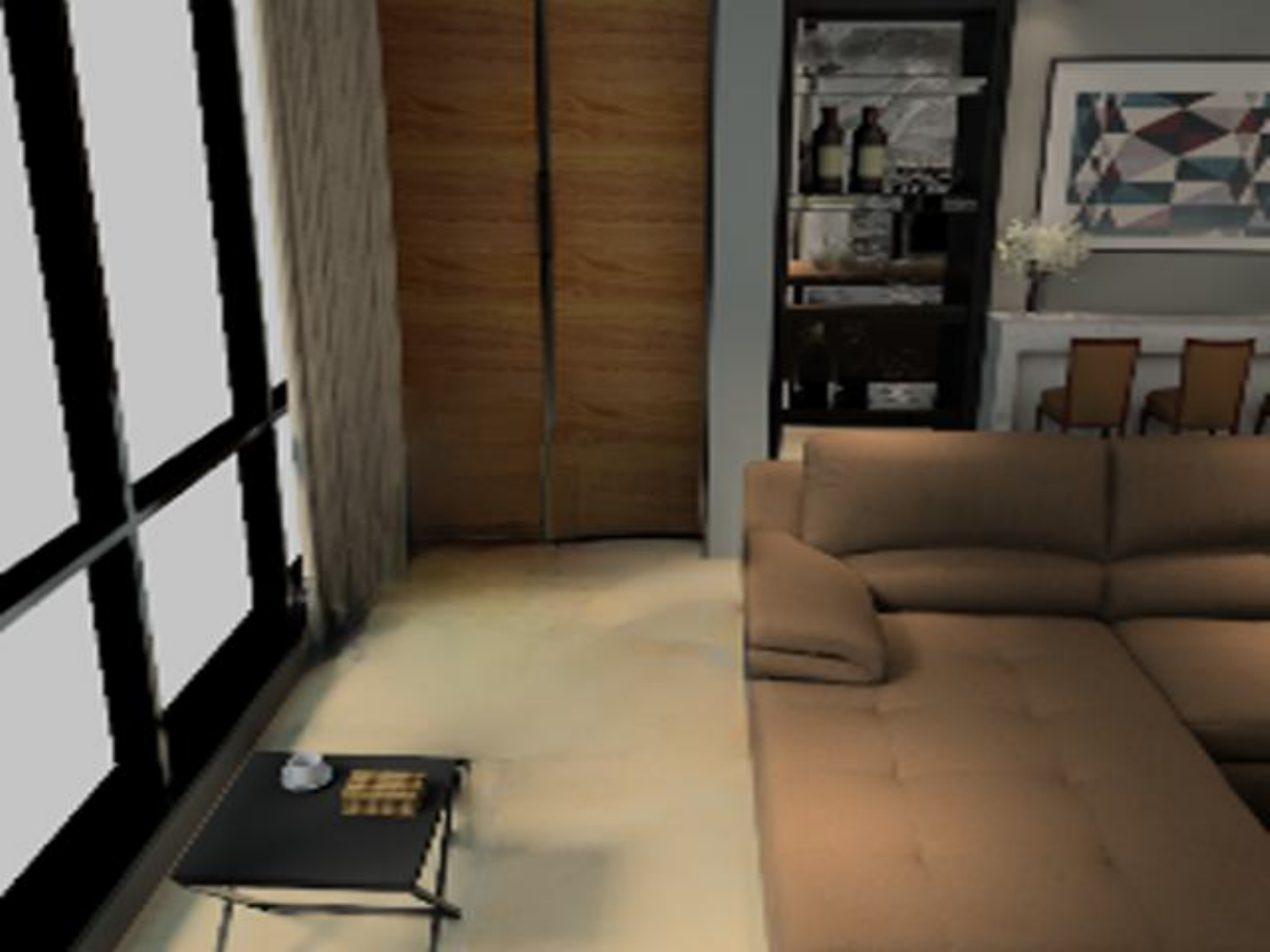}};
        \spy [magnification=1.8] on (3*\myImgWs-0.2, -1*\myImgHs-0.25) in node at (3*\myImgWs, -1*\myImgHs);
        
        \node[tight] (n1) at (0,-2*\myImgHs) {\includegraphics[width=00.24\linewidth]{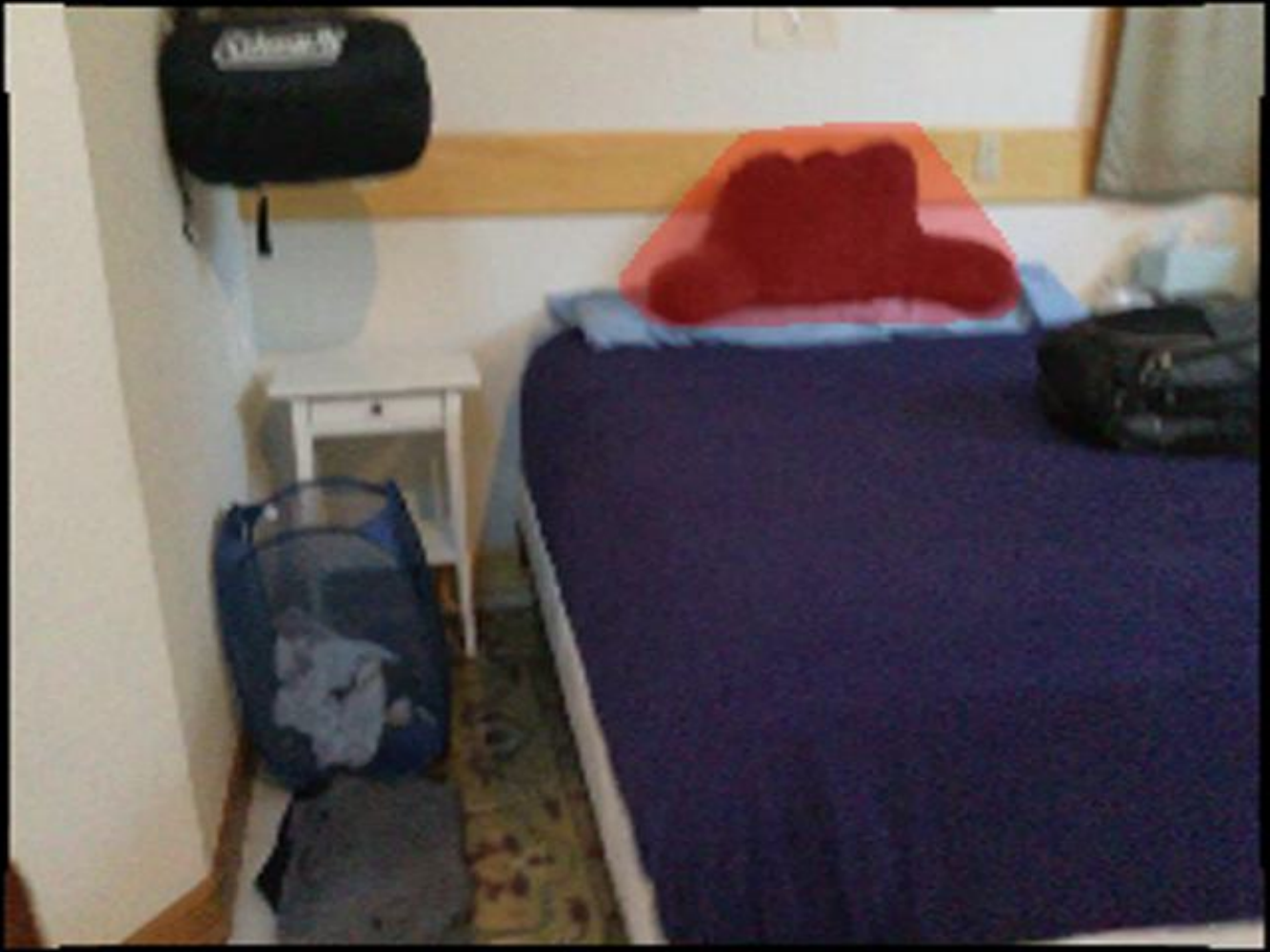}};
        \spy[magnification=2.3] on (0*\myImgWs+0.30, -2*\myImgHs+0.40) in node at (0*\myImgWs,-2*\myImgHs);
        
        \node[tight] (n9) at (1*\myImgWs,-2*\myImgHs) {\includegraphics[width=0.24\linewidth]{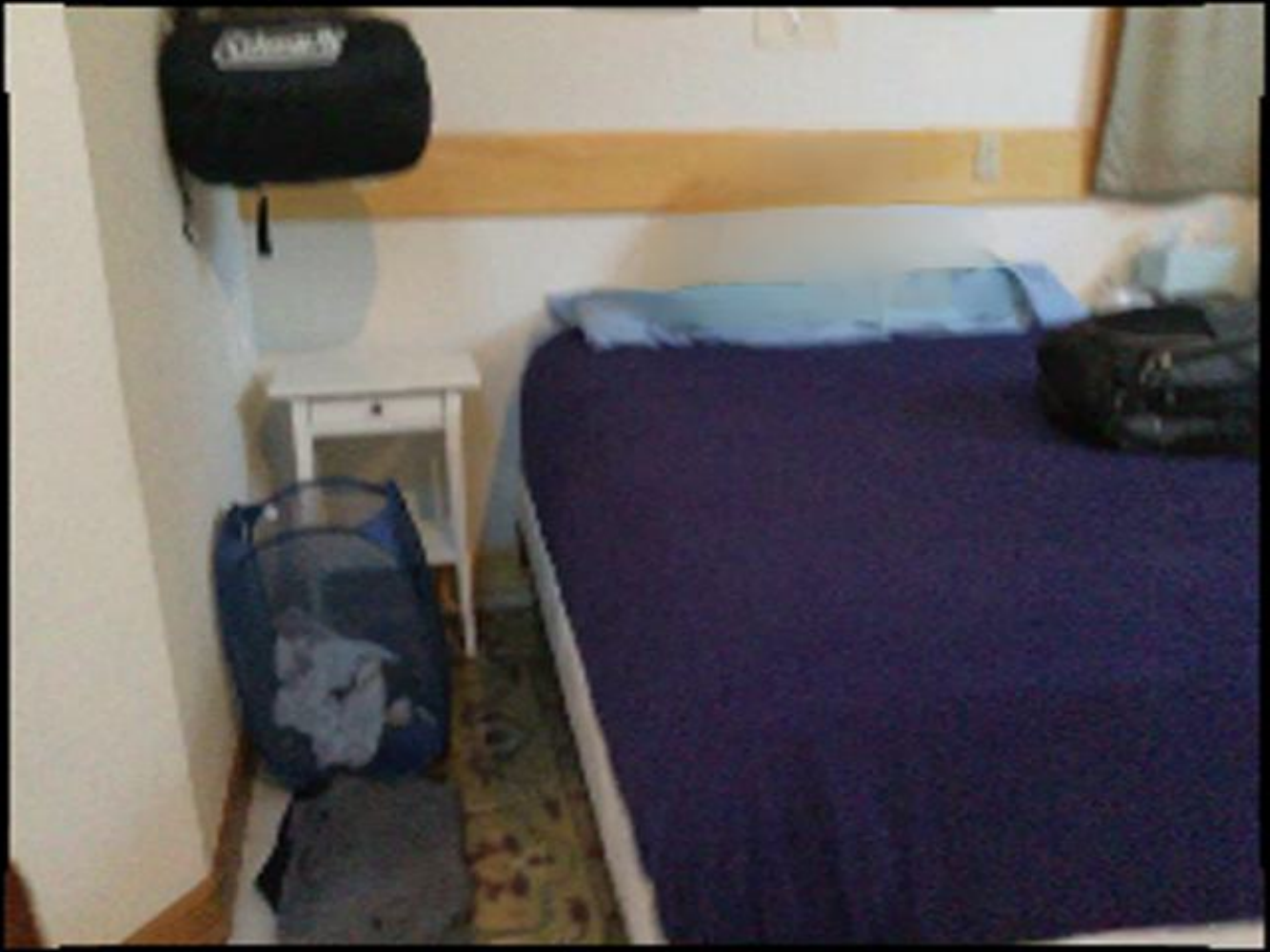}};
        \spy[magnification=2.3] on (1*\myImgWs+0.30, -2*\myImgHs+0.40) in node at (1*\myImgWs,-2*\myImgHs);
        
        \node[tight] (n1) at (2*\myImgWs, -2*\myImgHs) {\includegraphics[width=0.24\linewidth]{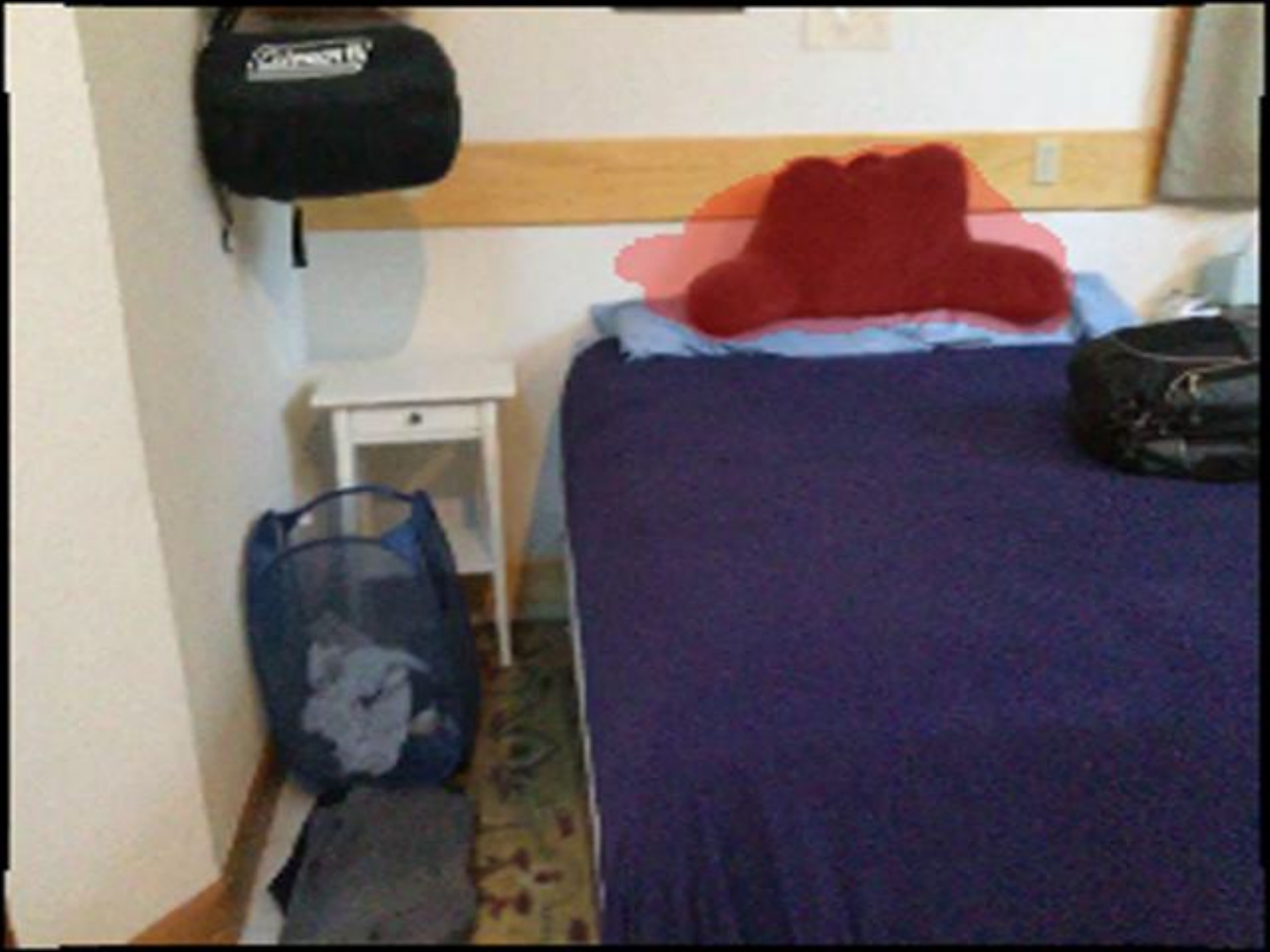}};
        \spy [magnification=2.3] on (2*\myImgWs+0.30, -2*\myImgHs+0.40) in node at (2*\myImgWs, -2*\myImgHs);
        
        \node[tight] (n9) at (3*\myImgWs, -2*\myImgHs) {\includegraphics[width=0.24\linewidth]{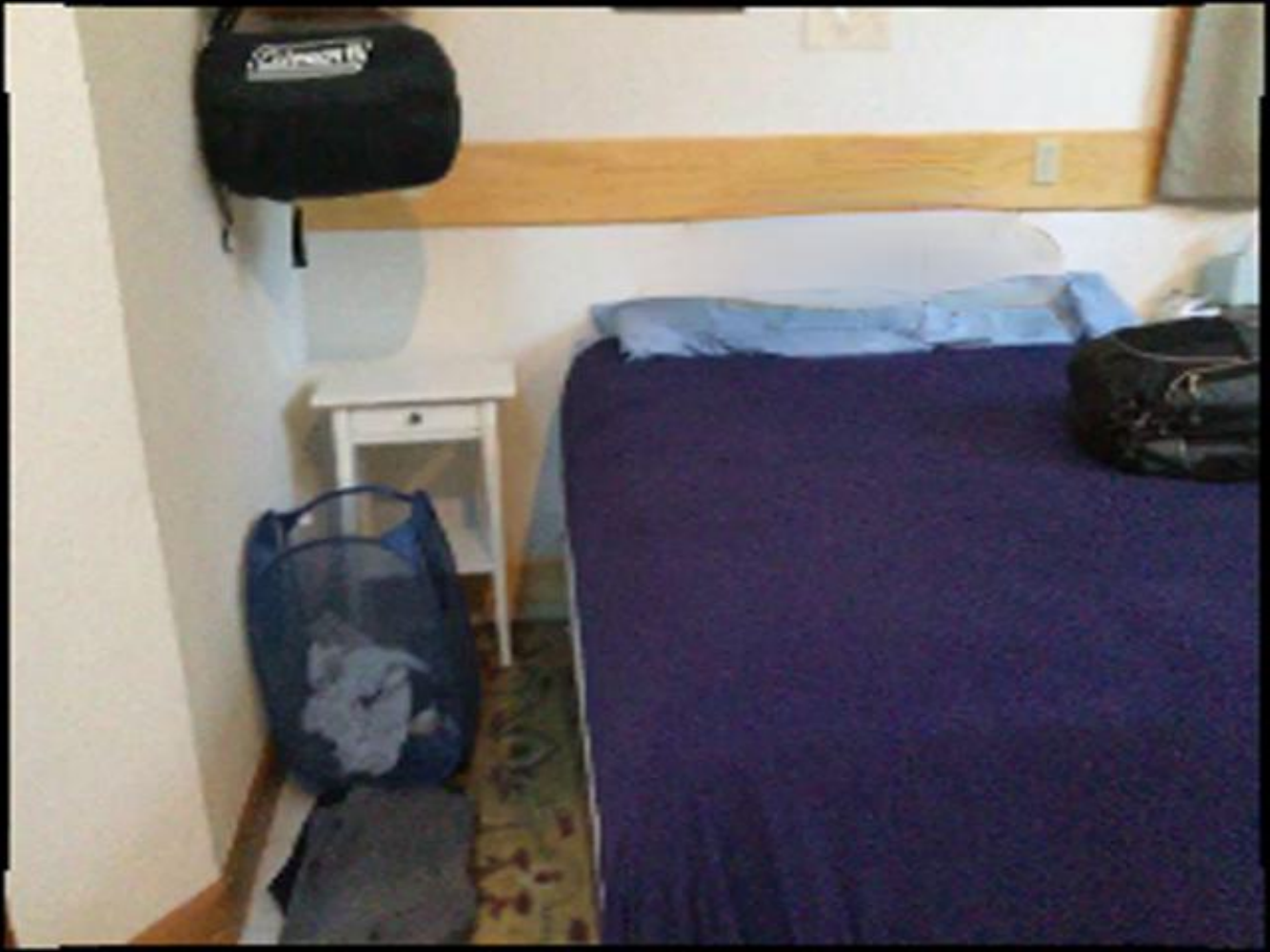}};
        \spy [magnification=2.3] on (3*\myImgWs+0.30, -2*\myImgHs+0.40) in node at (3*\myImgWs, -2*\myImgHs);
        
        \node[tight] (n1) at (0,-3*\myImgHs) {\includegraphics[width=00.24\linewidth]{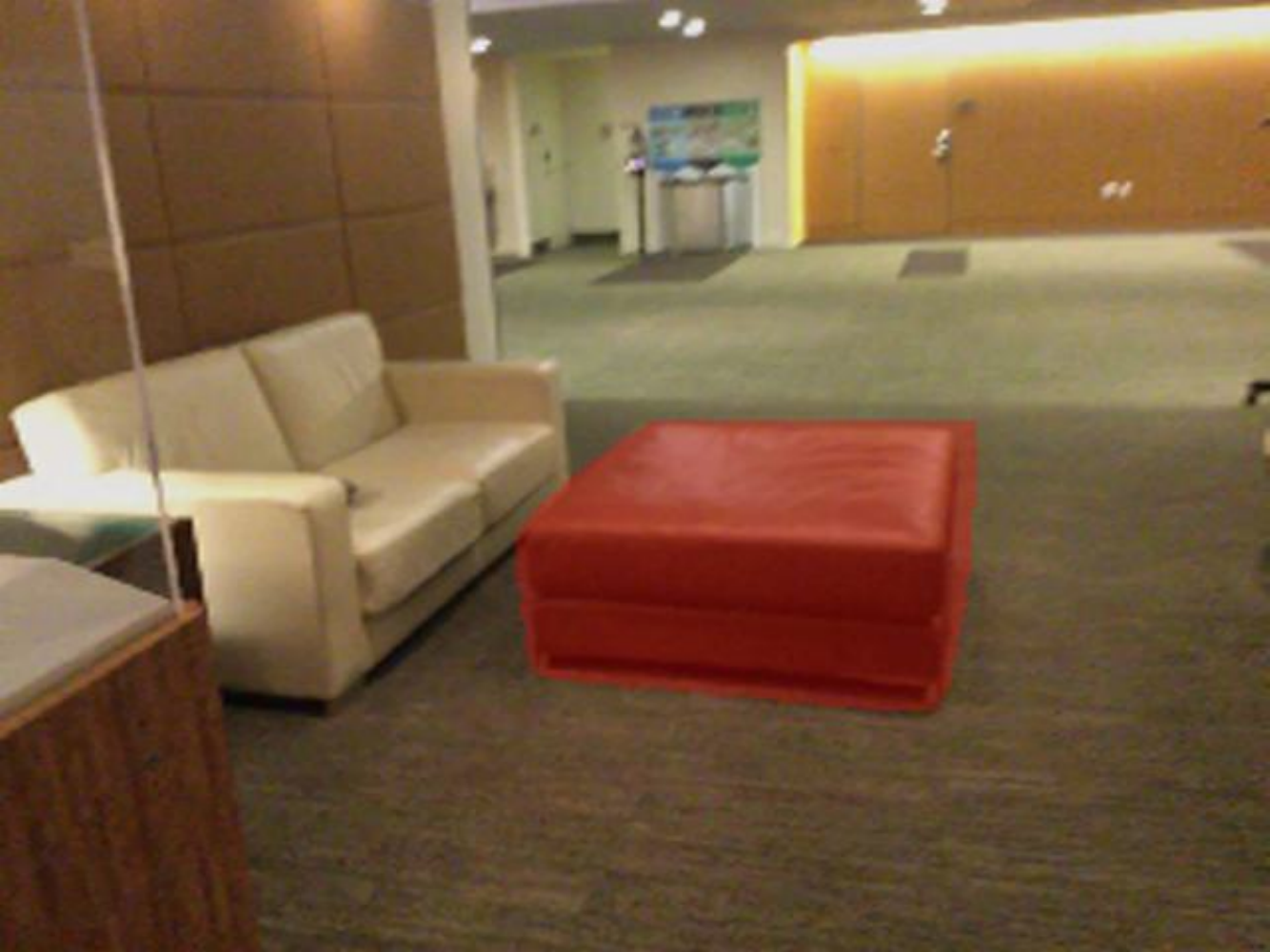}};
        \spy[magnification=2.0] on (0*\myImgWs+0.15, -3*\myImgHs-0.10) in node at (0*\myImgWs,-3*\myImgHs);
        
        \node[tight] (n9) at (1*\myImgWs,-3*\myImgHs) {\includegraphics[width=0.24\linewidth]{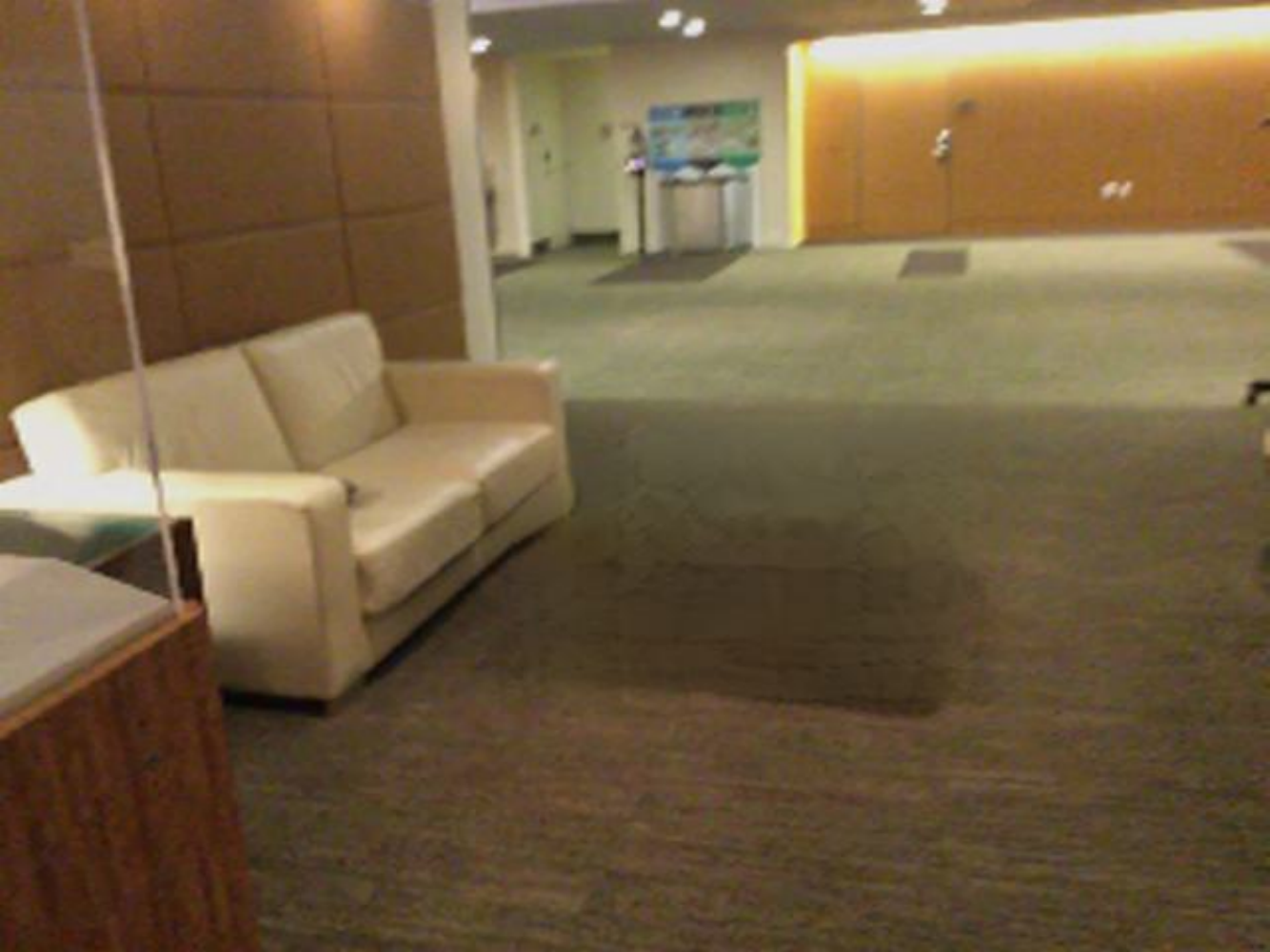}};
        \spy[magnification=2.0] on (1*\myImgWs+0.15, -3*\myImgHs-0.10) in node at (1*\myImgWs,-3*\myImgHs);
        
        \node[tight] (n1) at (2*\myImgWs, -3*\myImgHs) {\includegraphics[width=0.24\linewidth]{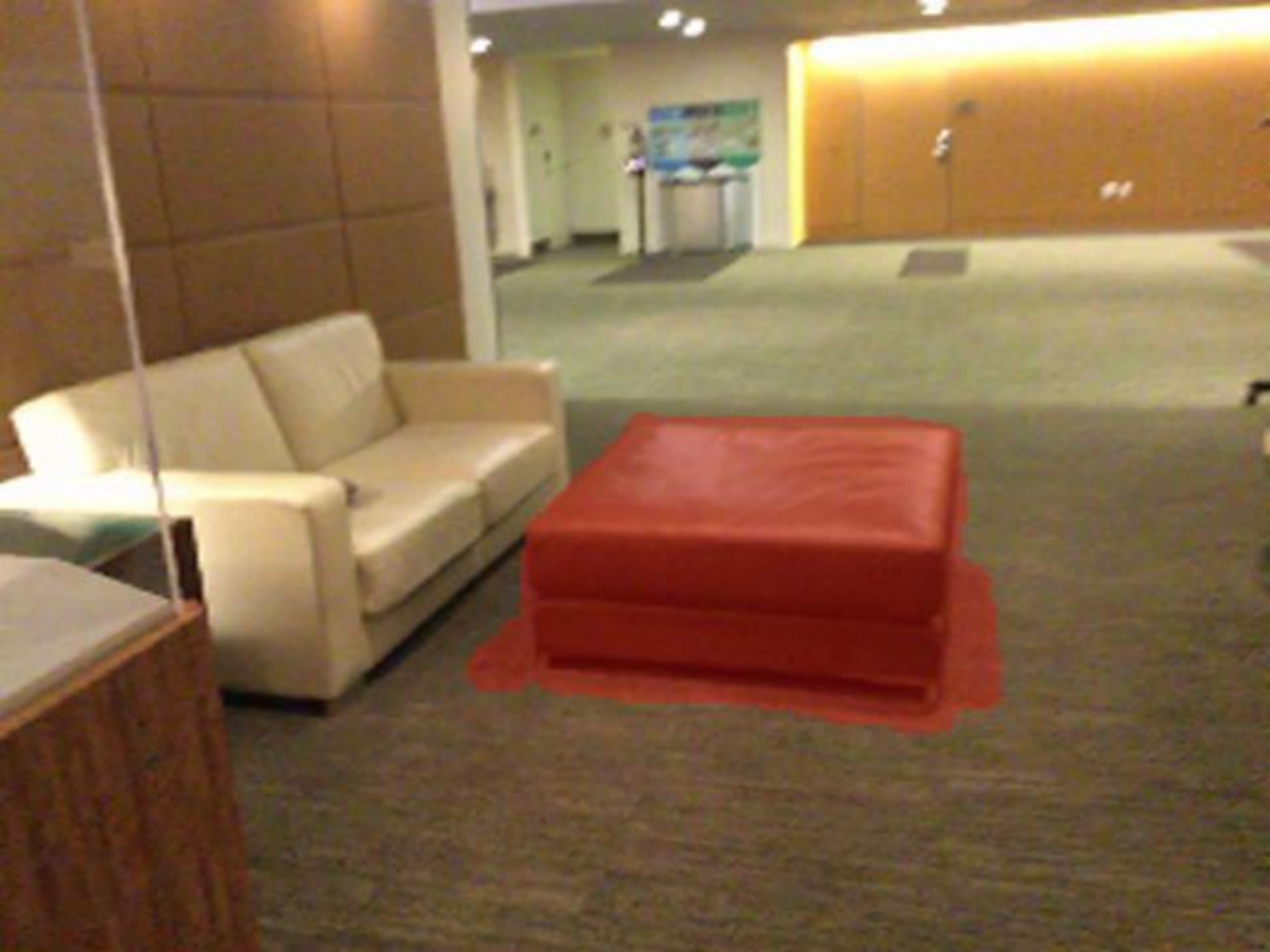}};
        \spy [magnification=2.0] on (2*\myImgWs+0.15, -3*\myImgHs-0.10) in node at (2*\myImgWs, -3*\myImgHs);
        
        \node[tight] (n9) at (3*\myImgWs, -3*\myImgHs) {\includegraphics[width=0.24\linewidth]{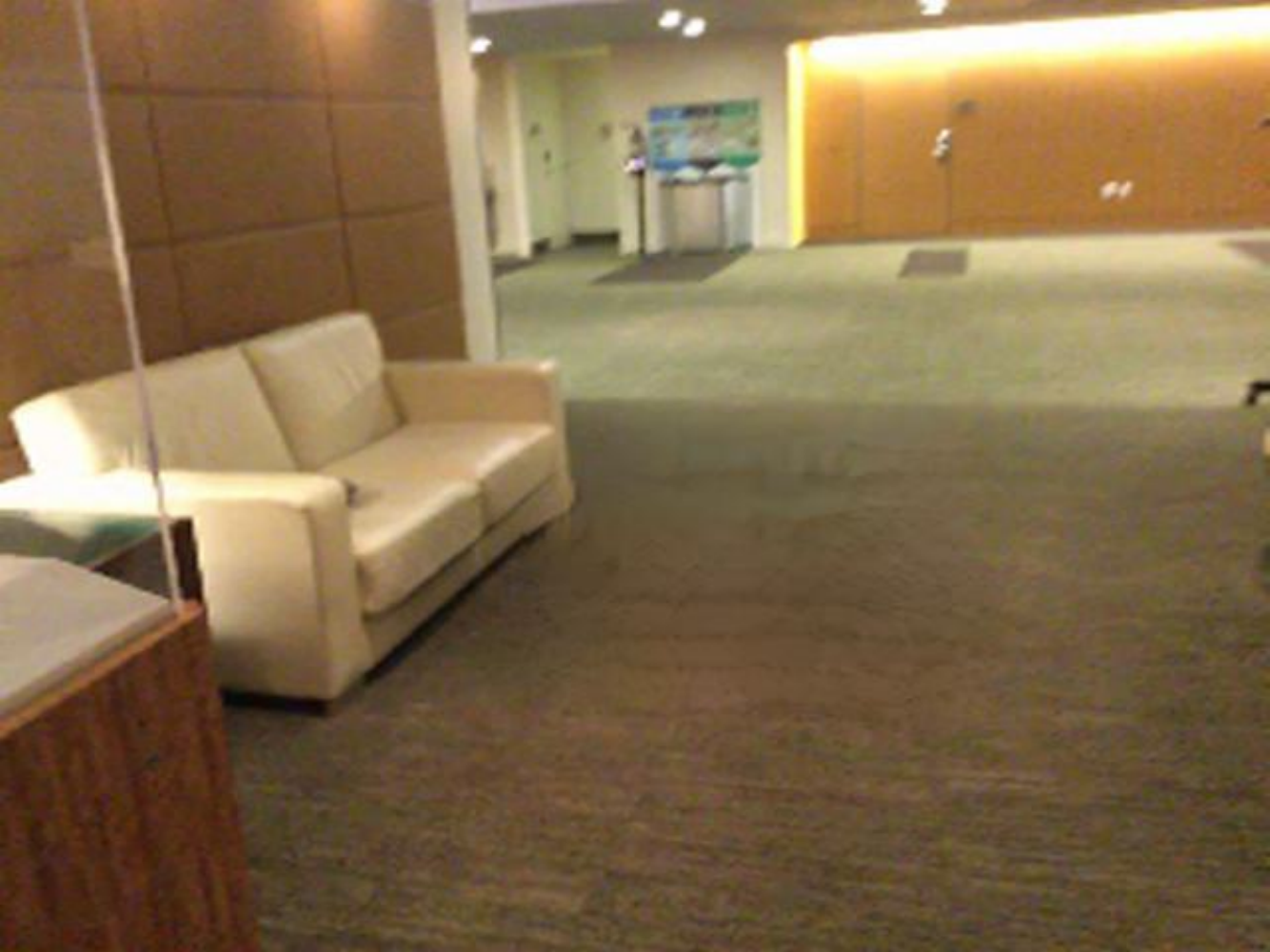}};
        \spy [magnification=2.0] on (3*\myImgWs+0.15, -3*\myImgHs-0.10) in node at (3*\myImgWs, -3*\myImgHs);
        
    \end{tikzpicture}
    \caption{Comparison between automatically created mask using instance segmentations and manually created masks. DeepDR is also able to inpaint shadows, if they are appropriately masked.}
    \label{fig:shadowseg}
\end{figure}

\subsection{Comparison with depth-from-RGB inpainting}
As already mentioned in the main paper, depth inpainting literature mostly focuses on depth completion in regions that are visible in the corresponding RGB image. InDepth, NLSPN and DM-LRN mentioned above are examples of such methods. While our task of inpainting hidden structures in diminished parts of the scene is fundamentally different from depth-from-RGB inpainting, it is compelling to draw comparisons between the two tasks. Therefore, we conducted an experiment applying DeepDR for depth-from-RGB inpainting on NYU-depth V2~\cite{Silberman:ECCV12} and ScanNet, which are common benchmarks in depth completion~\cite{senushkin2021decoder,park2019semantic,ghosh2020depth,wang2022rgb,atapour2017depthcomp}.  We directly compare to results reported in DM-LRN, which is the only work evaluating, but not trained on these datasets, allowing a fair comparison. We use the semi-dense sampling strategy reported in their paper to generate masks and feed complete RGB and masked depth to our framework. Results are shown in \cref{tab:depth_completion}. The results on ScanNet show that DeepDR is effective in using visible RGB information to fill missing depth, resulting in an RMSE of 0.262 m compared to an RMSE of 0.484 m for the joint RGB-D inpainting task in \cref{tab:comparison_scan}. Still, DeepDR performs slightly worse on the depth-from-RGB inpainting task than the baseline method, particularly on NYU-depth V2, which is challenging for DeepDR due to its lack of consecutive frames.

\begin{table}[htbp]
  \centering
  \footnotesize
  \caption{RMSE in meters for D-from-RGB inpainting.}
    \begin{tabular}{lcc}
    \toprule
    & NYU-d V2 & ScanNet\\
    \midrule
    DeepDR (ours) & 0.281 & 0.262 \\
    DM-LRN & 0.205 & 0.198 \\
    \bottomrule
    \end{tabular}%
  \label{tab:depth_completion}%
\end{table}%

\section{Additional results}

\subsection{Computational complexity of ablation models}
\label{sec:ablation_suppl}
We analyze the computational complexity in terms of inference time, multiply-adds (MADs) and number of total parameters of our ablation models in~\cref{tab:ablation}. Evidently, RGB-D SPADE is the major driver of computational complexity, leading to an almost doubled inference time, as well as a significantly higher number of MADs and parameters. Our separate encoding strategy proves to be very efficient, improving the performance of our final model while decreasing the overall parameter count.
 
\begin{table}[ht]
  \centering
  \caption{Efficiency of DeepDR on a Nvidia GeForce GTX 1080 Ti GPU in comparison to the ablation models.}
  \resizebox{\linewidth}{!}{
    \begin{tabular}{l|ccc}
    \toprule
     & \multicolumn{3}{c}{Efficiency} \\
    \cmidrule{2-4}
    Model & Time $\downarrow$  (ms) & MADs $\downarrow$ & Params $\downarrow$ \\
    \midrule
    no temporal & 4.17 & 163.3 G & 69.8 M\\
    no RGB-D SPADE & \textbf{2.41} & \textbf{125.8 G} & \textbf{65.7 M} \\
    joint encoder & 4.42 & 174.6 G & 71.1 M \\
    \midrule
    DeepDR (Full model) & 4.43 & 184.3 G & 69.9 M\\
    \bottomrule
    \end{tabular}
    }
  \label{tab:ablation_suppl}%
\end{table}

\subsection{More qualitative results for the DR use case}
\label{sec:qual_suppl}

\begin{figure*}[ht]
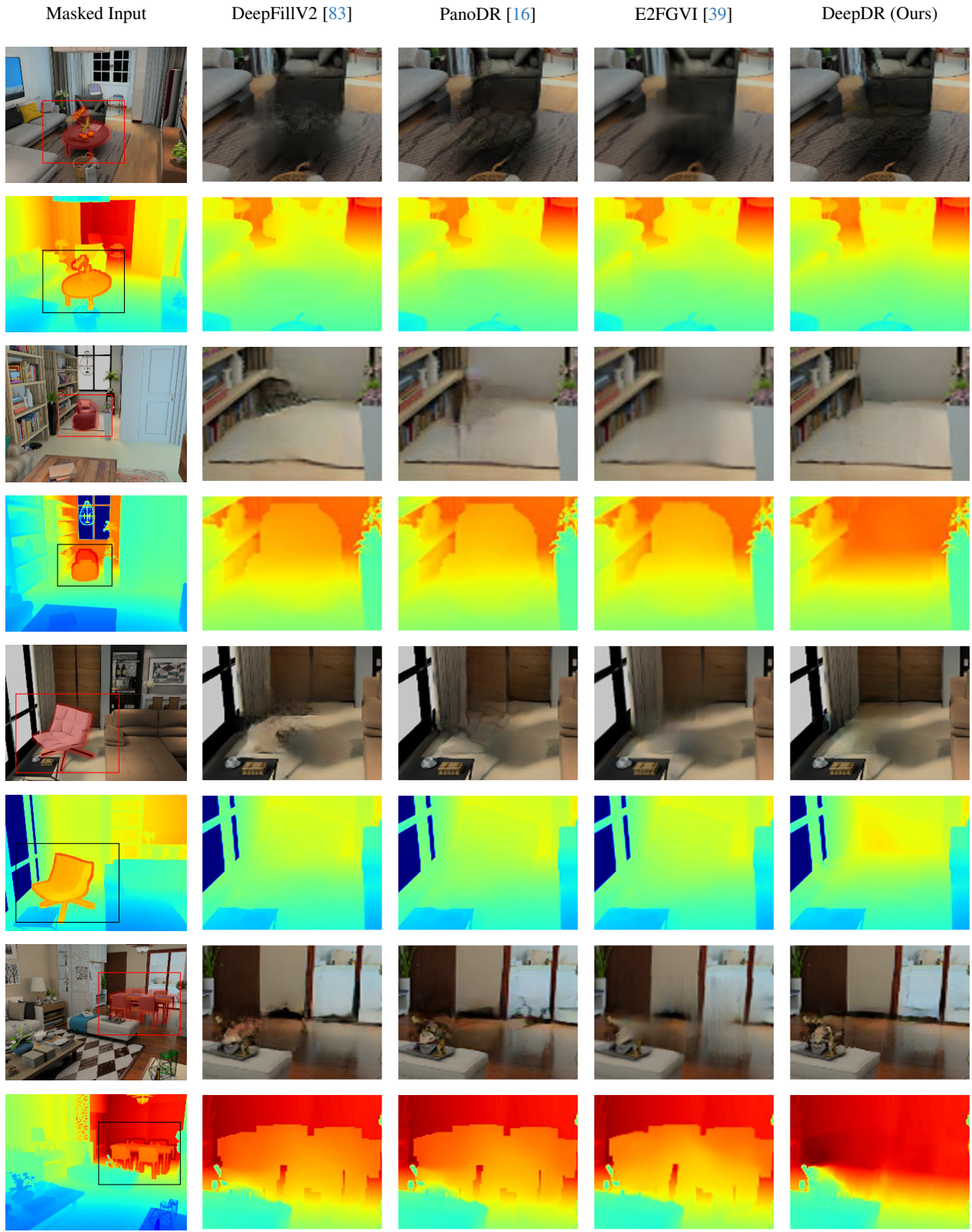

    \noindent
    \centering

    \caption[Qualitative comparison of color images and depth maps for diminishing objects from InteriorNet]{Qualitative comparison of color images and depth maps for diminishing objects from InteriorNet~\cite{li2018interiornet}.}
    \label{fig:dr:comparison_interiornet}
\end{figure*}

\begin{figure*}[ht]
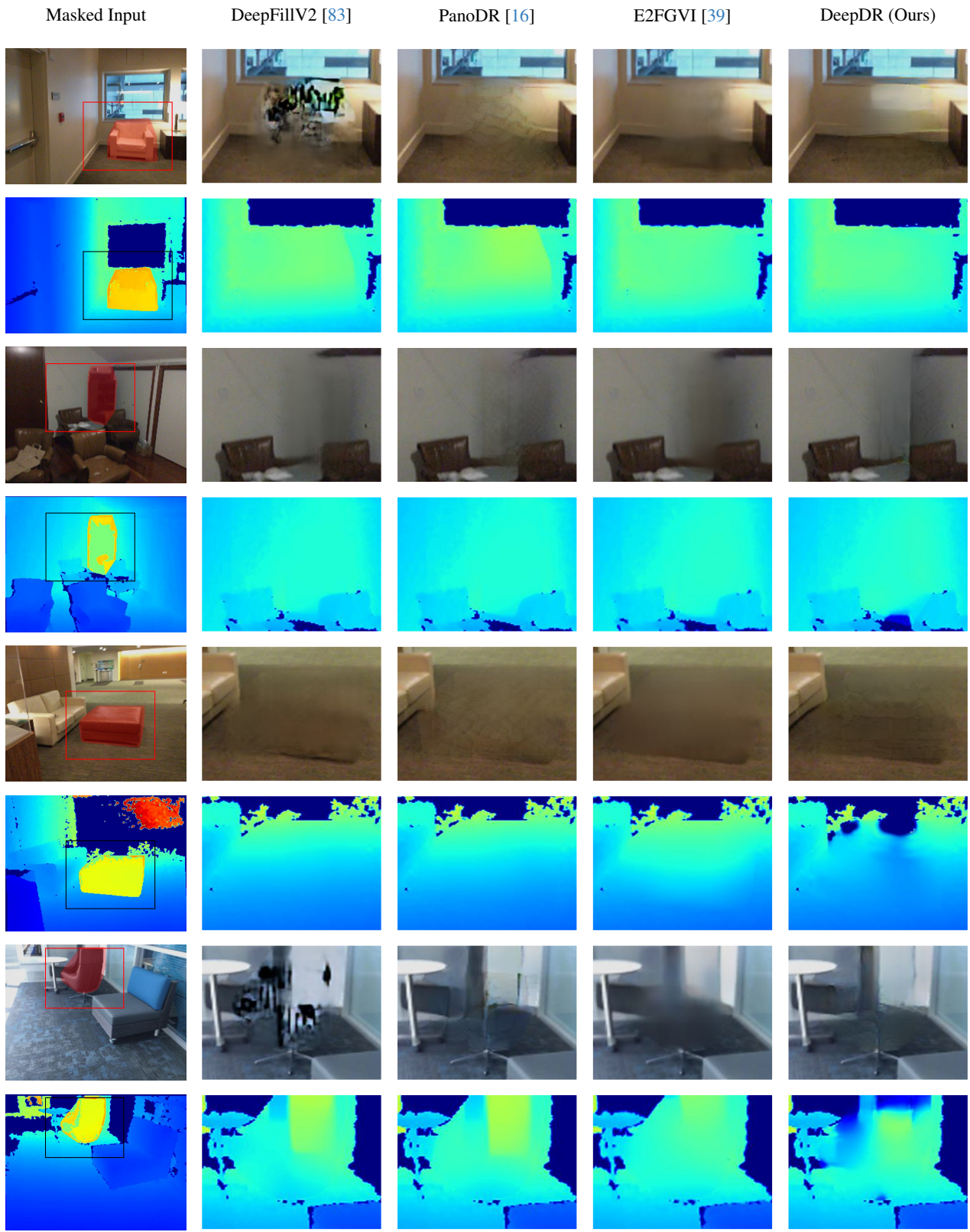

    \noindent
    \centering
    %
    \caption[Qualitative comparison of color images and depth maps for diminishing objects from ScanNet]{Qualitative comparison of color images and depth maps for diminishing objects from ScanNet~\cite{dai2017scannet}.}
    \label{fig:dr:comparison_scannet}
\end{figure*}

Supplementary to the qualitative results in \cref{fig:comparison_interior_scan} and \cref{fig:comparison_dyna}, we show more results of DeepDR in comparison to the baselines~\cite{yu2019free,gkitsas2021panodr,li2022towards,bevsic2020dynamic} for DR object removal on InteriorNet in \cref{fig:dr:comparison_interiornet}, DynaFill in \cref{fig:dr:comparison_dyna} and on ScanNet in \cref{fig:dr:comparison_scannet}. 

\subsection{Qualitative results for 3D scene editing}
\begin{figure*}[ht]
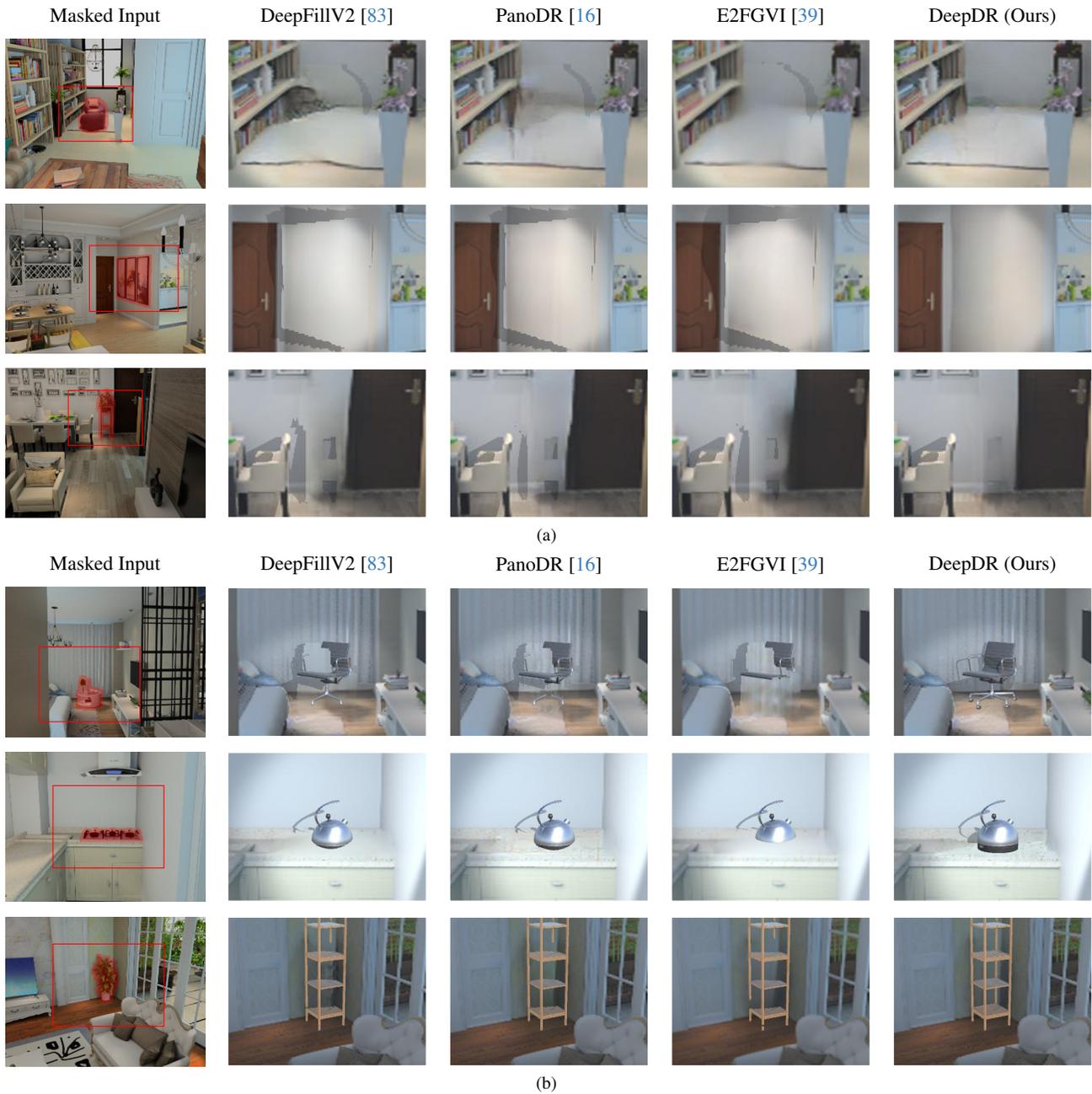

    \noindent
    \centering
    \subcaptionbox{}{
    %
    }
    
    \subcaptionbox{}{
    %
    }
    \caption[Qualitative comparison for 3D scene editing after diminishing objects via inpainting.]{Qualitative comparison for 3D scene editing after diminishing objects via inpainting. The scene is reconstructed in 3D, light sources are added (a) and furniture or accessory items are replaced (b).}
    \label{fig:dr:comparison_editing}
\end{figure*}

While the importance of coherent image and geometry inpainting may not be immediately obvious, it becomes clear when looking at applications in 3D scene editing, such as interior re-design. In our indoor scene scenario, a typical use case is re-decorating rooms. To demonstrate this use case, we reconstruct a textured 3D mesh from the inpainted RGB-D pairs in 3D using pose and augment it with additional virtual light sources (\cref{fig:dr:comparison_editing}a) and furniture or decoration objects (\cref{fig:dr:comparison_editing}b). 

As seen from these examples, incorrect depth inpainting leads to serious artifacts, such as ghost shadows or intersections and overlapping of the inpainted background with newly added objects. Since our method significantly outperforms related work in terms of depth inpainting, it does not cause such artifacts and is, therefore, best suited for 3D scene editing applications.

\subsection{Qualitative results using random object masks}
\label{sec:qual_test_suppl}

\begin{figure*}[ht]
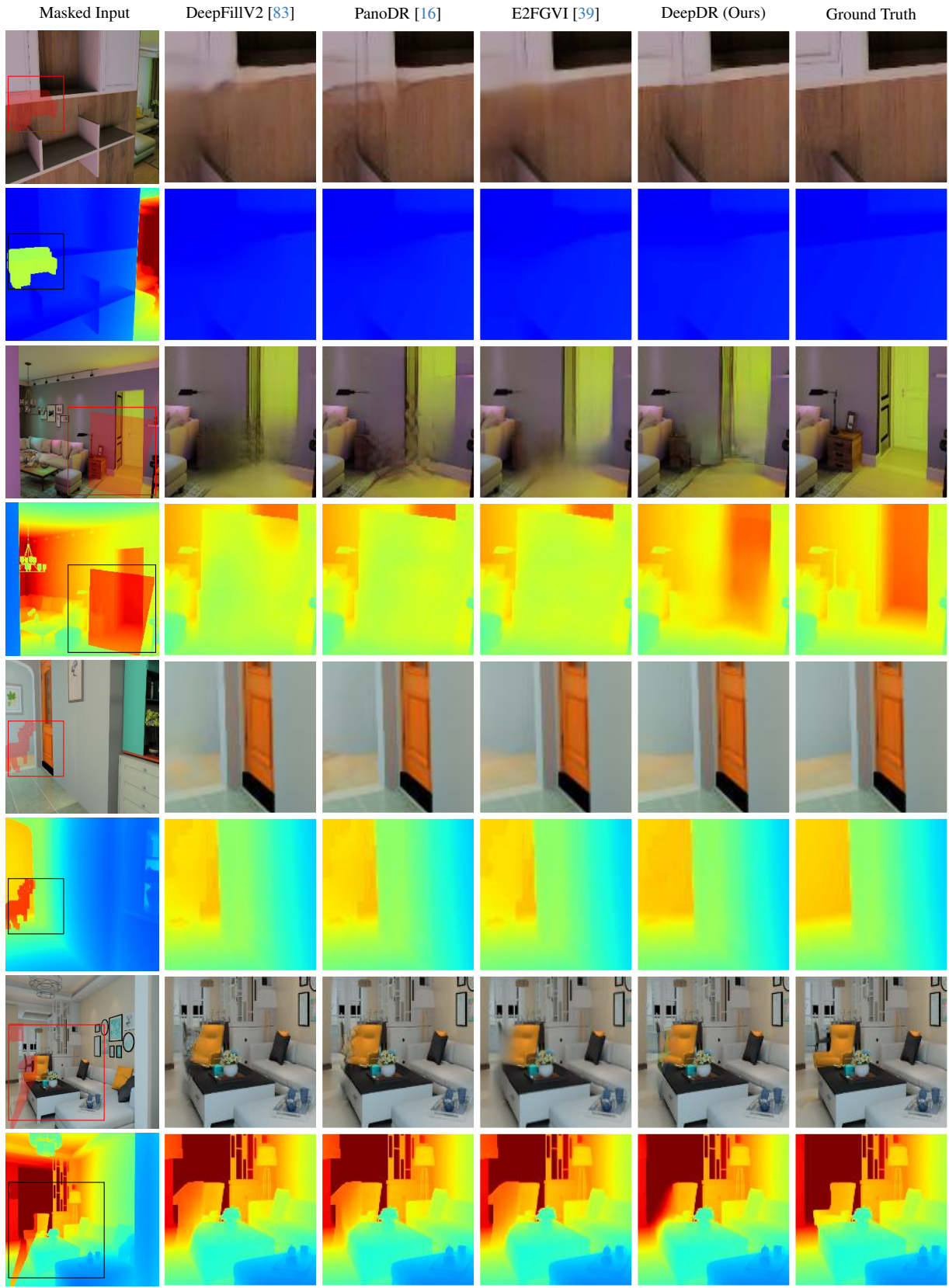

    \centering
    \noindent
    %
    \caption[Qualitative comparison on InteriorNet for inpainting random object masks.]{Qualitative comparison on InteriorNet~\cite{li2018interiornet} for inpainting random object masks.}
    \label{fig:dr:testset_interiornet}
\end{figure*}

As mentioned before, InteriorNet and ScanNet have no ground truth for the DR use case. Ideally, we would use training and testing pairs consisting of rooms before and after some items have been removed. Such data is very difficult to obtain in a real setting, but even synthetic data is costly to obtain, both in terms of computational and human resources as well as time. Therefore, for this datasets, we simulate the object removal task by overlaying random object masks over the scene and thus, the original image serves as ground truth. We use this strategy for both training and computing our quantitative results during testing. A qualitative comparison of inpainting using random object masks between our method and the baselines is given in~\cref{fig:dr:testset_interiornet} for InteriorNet, and~\cref{fig:dr:testset_scannet} for ScanNet.

Akin to the qualitative results for the DR use case, it is noticeable that DeepDR exceeds other methods in reconstructing sharp textures while preserving important structural properties of the scene. Furthermore, our method can reconstruct sharp depth edges, while the baselines fail to reconstruct the geometry of the scene, particularly for structures far away from the camera.

\subsection{Intermediate segmentation results}
\label{sec:seg_suppl}

\begin{figure*}
    \centering
    \includegraphics[width=\textwidth]{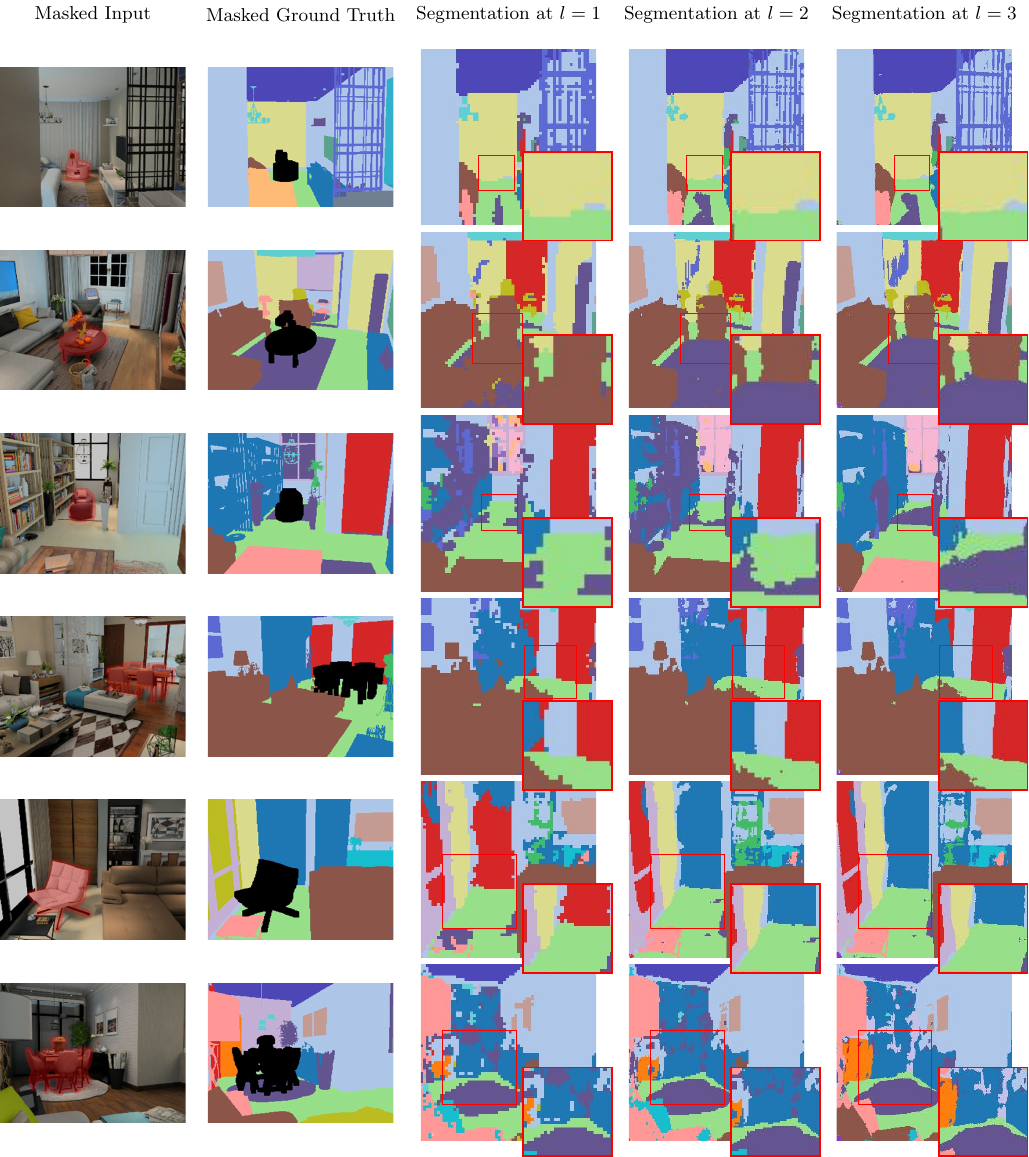}
    \caption{Analysis of the semantic segmentations produced within our up blocks on InteriorNet~\cite{li2018interiornet}.}
    \label{fig:sem_seg}
\end{figure*}

\begin{figure*}
    \centering
    \includegraphics[width=\textwidth]{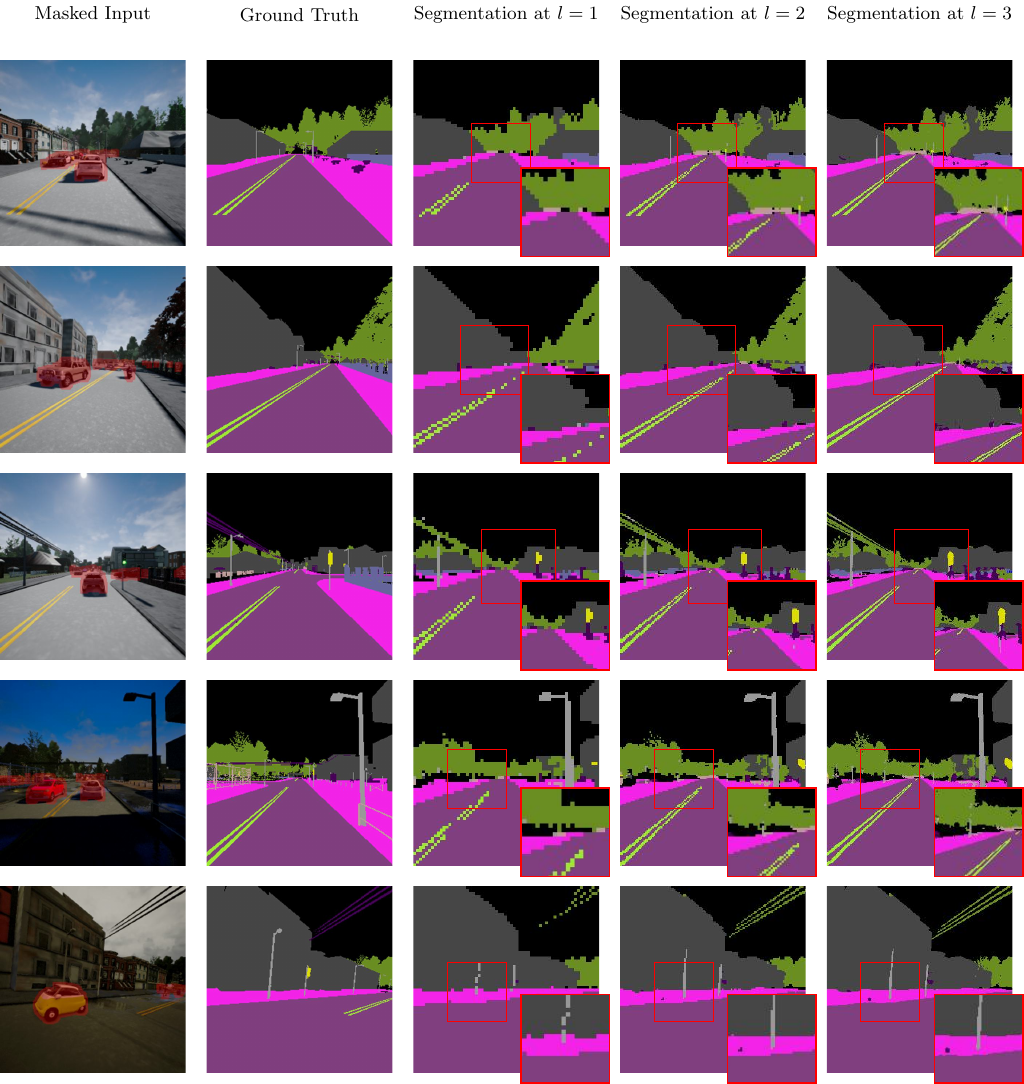}
    \caption{Analysis of the semantic segmentations produced within our up blocks on DynaFill~\cite{bevsic2020dynamic}.}
    \label{fig:sem_seg_dynafill}
\end{figure*}

\begin{figure*}
    \centering
    \includegraphics[width=\textwidth]{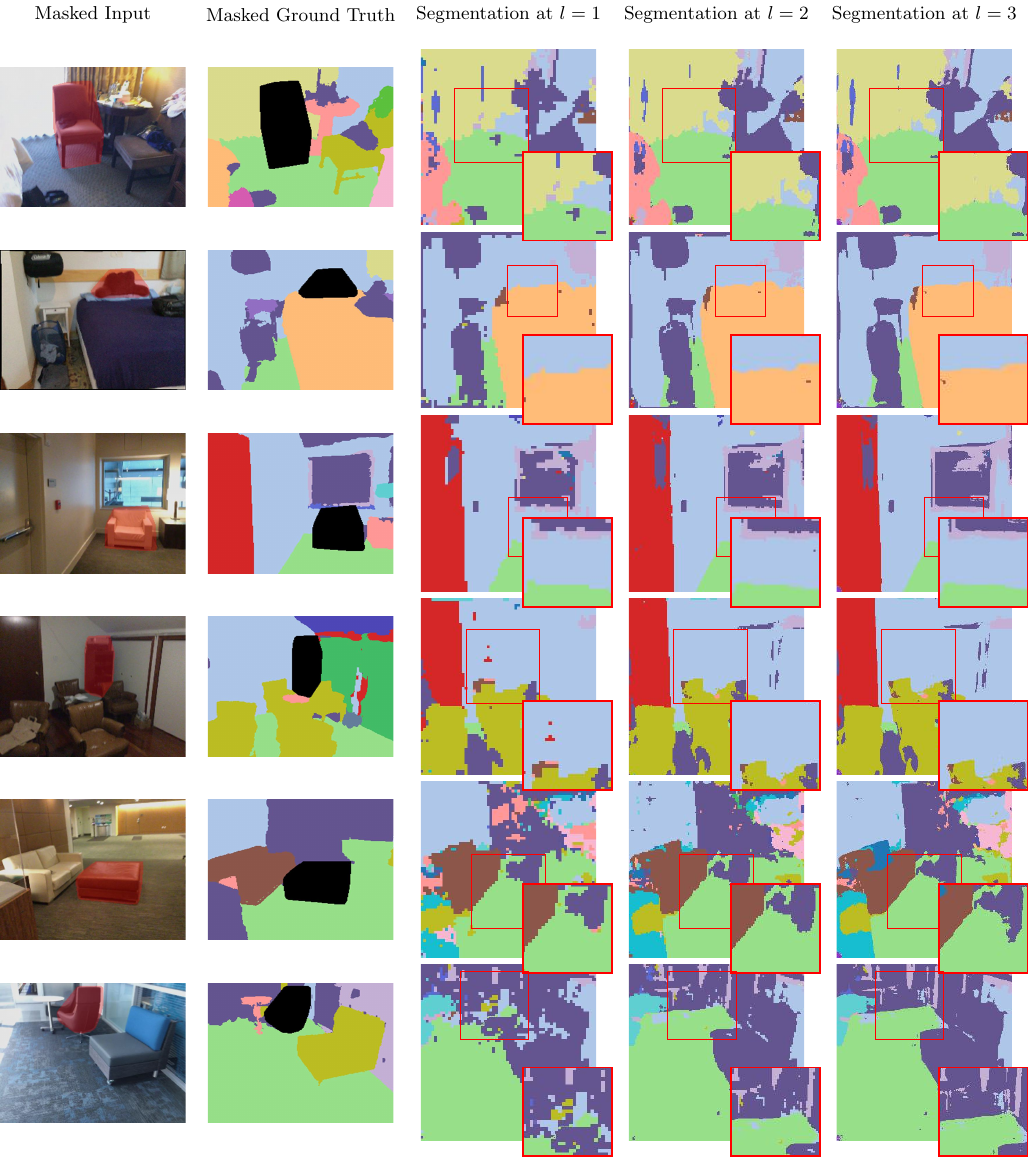}
    \caption{Analysis of the semantic segmentations produced within our up blocks on ScanNet~\cite{dai2017scannet}.}
    \label{fig:sem_seg_scannet}
\end{figure*}

Our up blocks produce intermediate semantic segmentations of the scene at feature scale using a pyramid pooling module~\cite{zhao2017pyramid}. These maps are used to modulate the activations during decoding to ensure sharp and coherent boundaries in RGB and depth outputs. In~\cref{fig:sem_seg},~\cref{fig:sem_seg_dynafill} and~\cref{fig:sem_seg_scannet}, we show these intermediate segmentations from each of the three up blocks in our final architecture. It is evident that the segmentation accuracy improves with higher feature dimensions. Notably, segmentation on DynaFill (\cref{fig:sem_seg_dynafill}) is more accurate, which we attribute to the lower variability and smaller number of semantic classes (12 vs. 40) in the dataset. Although our network does not produce perfect segmentations, it is able to accurately reconstruct clean object borders and plausible semantics, which leads to sharp edges and coherent textures in the resultant image and depth outputs. Still, in particular, on unseen, real data in ScanNet (\cref{fig:sem_seg_scannet}), some regions are incorrectly classified, which might decrease the effectiveness of RGBD SPADE. We aim to overcome this limitation by fine-tuning our models on real data, reducing the number of semantic classes by merging similar classes, and by exploring RGB-D semantic segmentation strategies~\cite{chen2020bi, xiong2020variational, cao2021shapeconv} to leverage depth information more effectively for intermediate semantic segmentation.

\gdef\myImgW{3.4}
\gdef\myImgH{2.6}

\gdef\myImgW{2.6}
\gdef\myImgH{2.6}

\begin{figure*}[ht]
    \noindent
    \centering
    %
    \caption[Qualitative comparison for diminishing objects from DynaFill.]{Qualitative comparison for diminishing objects from DynaFill~\cite{bevsic2020dynamic}.}
    \label{fig:dr:comparison_dyna}
\end{figure*}%

\gdef\myImgW{3.4}
\gdef\myImgH{2.6}

\gdef\myImgW{3.5}
\gdef\myImgH{2.6}

\gdef\myImgW{2.7}
\gdef\myImgH{2.7}

\gdef\myImgW{2.7}
\gdef\myImgH{2.7}

\begin{figure*}[ht]
    \centering
    \noindent
    %
    \caption{Qualitative comparison on ScanNet~\cite{dai2017scannet} for inpainting random object masks.}
    \label{fig:dr:testset_scannet}
\end{figure*}

\end{document}